%% file: iccv_for_review.tex
\documentclass[10pt,twocolumn,letterpaper]{article}

\usepackage{iccv}
\usepackage{times}
\usepackage{epsfig}
\usepackage{graphicx}
\usepackage{amsmath}
\usepackage{amssymb}

\usepackage[pagebackref=true,breaklinks=true,letterpaper=true,colorlinks,bookmarks=false]{hyperref}

\usepackage{url}
\usepackage{wrapfig}
\usepackage{xfrac}
\usepackage{xspace}
\usepackage{lipsum} 

\usepackage{changepage}  
\usepackage{amsmath,amssymb} 
\usepackage{color}
\usepackage{bm}
\usepackage{stackengine}
\usepackage{placeins}
\usepackage{makecell}
\usepackage[export]{adjustbox}
\usepackage[percent]{overpic}

\usepackage{tablefootnote}

\makeatletter
\renewcommand{\paragraph}{%
  \@startsection{paragraph}{4}%
  {\z@}{2ex \@plus 1ex \@minus .2ex}{-1em}%
  {\normalfont\normalsize\bfseries}%
}
\makeatother

\usepackage{array,multirow}
\usepackage{float}

\usepackage{enumitem}
 
\usepackage{microtype}
\usepackage{graphicx}
\usepackage{subfigure}
\usepackage{booktabs} 

\usepackage{hyperref}

\input{mymacros.tex}

\newcommand{\etalnospace}{{et al.}}

\iccvfinalcopy 

\def\iccvPaperID{2939} 

\ificcvfinal\fi
\begin{document}

\title{Generative Adversarial Networks for Extreme Learned Image Compression}

\author{
\begin{tabular}{ccccc}
        Eirikur Agustsson\thanks{The first three authors contributed equally.} &
        Michael Tschannen$^{*}$ &
        Fabian Mentzer$^{*}$ &
        Radu Timofte &
        Luc Van Gool\\
        {\scriptsize aeirikur@vision.ee.ethz.ch} &
        {\scriptsize mi.tschannen@gmail.com} &
        {\scriptsize mentzerf@vision.ee.ethz.ch} &
        {\scriptsize timofter@vision.ee.ethz.ch} &
        {\scriptsize vangool@vision.ee.ethz.ch} 
    \end{tabular}\\[3ex]
    ETH Z\"urich, Switzerland
}

\maketitle

\begin{abstract}
We present a learned image compression system based on GANs, operating at extremely low bitrates. 
Our proposed framework combines an encoder, decoder/generator and a multi-scale discriminator, which we train jointly for a generative learned compression objective.
The model synthesizes details it cannot afford to store, obtaining visually pleasing results at bitrates where previous methods fail and show strong artifacts.
Furthermore, if a semantic label map of the original image is available, our method can fully synthesize unimportant regions in the decoded image such as streets and trees from the label map, proportionally reducing the storage cost.
A user study confirms that for low bitrates, our approach is preferred to state-of-the-art methods, even when they use more than double the bits.
\end{abstract}
\begin{figure}[th!]
    \tabcolsep=1pt\relax
\tiny{
    \begin{tabular}{lcrlcrlcr}
        \multicolumn{3}{c}{%
            \includegraphics[width=.33\linewidth]{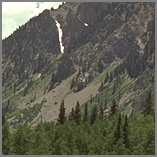}}&
        \multicolumn{3}{c}{%
            \includegraphics[width=.33\linewidth]{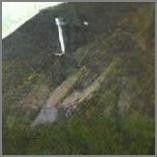}}&
        \multicolumn{3}{c}{%
            \includegraphics[width=.33\linewidth]{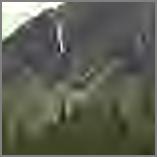}}\\[-1pt]

        \multicolumn{3}{l}{Original}&
        Ours & \multicolumn{2}{l}{1567 Bytes [B]}&
        JP2K & 3138B       & \emph{+100\% larger} \\[2pt]

        \multicolumn{3}{c}{%
            \includegraphics[width=.33\linewidth]{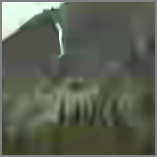}}&
        \multicolumn{3}{c}{%
            \includegraphics[width=.33\linewidth]{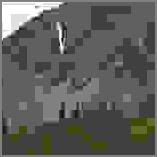}}&
        \multicolumn{3}{c}{%
            \includegraphics[width=.33\linewidth]{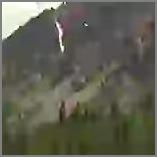}}\\[-1pt]

        BPG &   3573B       & \emph{+120\%} &
        JPEG &  13959B      & \emph{+790\%} &
        WebP &  9437B       & \emph{+502\%} \\[2pt]

        \multicolumn{9}{c}{%
            \includegraphics[width=\linewidth]{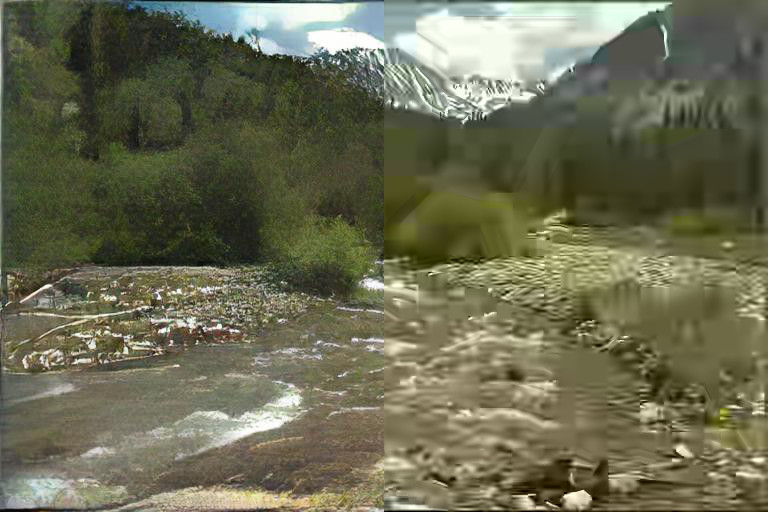}%
        } \\
        \multicolumn{6}{l}{\textbf{Ours} 1567 Bytes}&
        \multicolumn{3}{r}{\textbf{BPG} 3573 Bytes \enspace \emph{+128\%}} \\

    \end{tabular}}
\caption{\label{fig:visualexgckodak} Visual comparison of our result to that obtained by other codecs. Note that even when using more than twice the number of bytes, all other codecs are outperformed by our method visually.}\vspace{-0.2cm}
\end{figure}

\section{Introduction}
Image compression systems based on deep neural networks (DNNs), or deep compression systems for short, have become an active area of research recently.
These systems (e.g. \cite{theis2017lossy,balle2016end,rippel17a,balle2018variational,mentzer2018conditional}) are often competitive with modern engineered codecs such as WebP \cite{webpurl}, JPEG2000 \cite{jpeg2000taubman} and even BPG \cite{bpgurl} (the state-of-the-art engineered codec). Besides achieving competitive compression rates on natural images, they can be easily adapted to specific target domains such as stereo or medical images, and promise efficient processing and indexing directly from compressed representations \cite{torfason2018towards}. However, deep compression systems are typically optimized for traditional distortion metrics such as peak signal-to-noise ratio (PSNR) or multi-scale structural similarity (MS-SSIM) \cite{SSIM-MS}. For very low bitrates (below 0.1 bits per pixel (bpp)),
where preserving the full image content becomes impossible, 
these distortion metrics lose significance as they favor pixel-wise preservation of local (high-entropy) structure over preserving texture and global structure (see \cite{blau2018perception} and Sec.~\ref{sec:psnrbad}). 
To further advance deep image compression it is therefore of great importance to develop new training objectives beyond PSNR and MS-SSIM. 
A promising candidate towards this goal are adversarial losses \cite{goodfellow2014generative} which were shown recently to capture global semantic information and local texture, yielding powerful generators that produce visually appealing high-resolution images from semantic label maps \cite{isola2017image,wang2018highres}. 

In this paper, 
we propose a principled GAN framework for full-resolution image compression and use it to realize an extreme image compression system, targeting bitrates below 0.1bpp.
Furthermore, in contrast to prior work, we provide the first thorough user study of such a framework in the context of full-resolution image compression.

In our framework, we consider two modes of operation (corresponding to unconditional and conditional GANs \cite{goodfellow2014generative,mirza2014conditional}), namely
\begin{itemize}
\item \emph{generative compression (GC)}, preserving the overall image content while generating structure of different scales such as leaves of trees or windows in the facade of buildings, and
\item \emph{selective generative compression (SC)}, completely generating parts of the image from a semantic label map while preserving user-defined regions with a high degree of detail.
\end{itemize}

We emphasize that GC does not require semantic label maps (neither for training, nor for deployment). A typical use case for GC are bandwidth constrained scenarios, where one wants to preserve the full image as well as possible, while falling back to synthesized content instead of blocky/blurry blobs for regions for which not sufficient bits are available to store the original pixels. 
SC could be applied in a video call scenario where one wants to fully preserve people in the video stream, but a visually pleasing synthesized background serves the purpose as well as the true background. In the GC operation mode the image is transformed into a bitstream and encoded using arithmetic coding. SC requires a semantic/instance label map of the original image which can be obtained using off-the-shelf semantic/instance segmentation networks, e.g., PSPNet \cite{zhao2017pspnet} and Mask R-CNN \cite{he2017mask}, and which is stored as a vector graphic. This amounts to a small, image dimension-independent overhead in terms of coding cost. However, the size of the compressed image is reduced proportionally to the area which is generated from the semantic label map, typically leading to a significant overall reduction in storage cost.

For GC, a comprehensive user study shows that our compression system yields visually considerably more appealing results than BPG \cite{bpgurl} (the current state-of-the-art engineered compression algorithm) and the recently proposed autoencoder-based deep compression (AEDC) system \cite{mentzer2018conditional}. In particular, our GC models trained for compression of general natural images are preferred to BPG when BPG uses up to 95\% and 124\% more bits than those produced by our models on the Kodak \cite{kodakurl} and RAISE1K \cite{dang2015raise} data set, respectively. When constraining the target domain to the street scene images of the Cityscapes data set \cite{2016cityscapes}, the reconstructions of our GC models are preferred to BPG even when the latter uses up to 181\% more bits. To the best of our knowledge, these are the first results showing that a deep compression method outperforms BPG on the Kodak data set in a user study---and by large margins.

In the SC operation mode, our system seamlessly combines preserved image content with synthesized content, even for regions that cross multiple object boundaries, while faithfully preserving the image semantics. By partially generating image content we achieve bitrate reductions of over 50\% without notably degrading image quality. 

In summary, our main contributions are as follows.

\begin{itemize}
\item We provide a principled GAN framework for full-resolution image compression and use it to build an extreme image compression system.
\item We are the first to thoroughly explore such a framework in the context of full-resolution image compression. 
\item We set new state-of-the-art in visual quality based on a user study, with dramatic bitrate savings.
\end{itemize}

\section{Related work}

Deep image compression has recently emerged as an active area of research. The most popular DNN architectures for this task are to date auto-encoders \cite{theis2017lossy,balle2016end,agustsson2017soft,li2017learning,torfason2018towards,minnen2018joint,balle2018variational} and recurrent neural networks (RNNs) \cite{toderici2015variable,toderici2016full}. These DNNs transform the input image into a bit-stream, which is in turn losslessly compressed using entropy coding methods such as Huffman coding or arithmetic coding. To reduce coding rates, many deep compression systems rely on context models to capture the distribution of the bit stream \cite{balle2016end,toderici2016full,li2017learning,rippel17a,mentzer2018conditional}. 
Common loss functions to measure the distortion between the original and decompressed images are the mean-squared error (MSE) \cite{theis2017lossy,balle2016end,agustsson2017soft,li2017learning,balle2018variational,torfason2018towards}, or perceptual metrics such as MS-SSIM \cite{toderici2016full,rippel17a,balle2018variational,mentzer2018conditional}. Some authors rely on advanced techniques including multi-scale decompositions \cite{rippel17a}, progressive encoding/decoding strategies \cite{toderici2015variable,toderici2016full}, and generalized divisive normalization (GDN) layers \cite{balle2016end,balle2016code}.

Generative adversarial networks (GANs) \cite{goodfellow2014generative} have emerged as a popular technique for learning generative models for intractable distributions in an unsupervised manner. 
Despite stability issues \cite{salimans2016improved,arjovsky2017towards,arjovsky2017wasserstein,mao2017least}, they were shown to be capable of generating more realistic and sharper images than prior approaches and to scale to resolutions of $1024 \times 1024$px \cite{zhang2017stackgan,karras2017progressive} for some data sets. 
Another direction that has shown great progress are conditional GANs \cite{goodfellow2014generative,mirza2014conditional}, obtaining impressive results for image-to-image translation \cite{isola2017image,wang2018highres,zhu2017unpaired,liu2017unsupervised} on various data sets (e.g. maps to satellite images), reaching resolutions as high as $1024\times2048$px \cite{wang2018highres}.

The work of \cite{rippel17a} trains and evaluates a deep compression system optimized for the classical MS-SSIM~\cite{SSIM-MS} metric.
Furthermore, they supplement their method with an adversarial training scheme to reduce compression artifacts.
However, it is impossible to assess the benefit of their adversarial scheme since there is no ablation study showing its effect.
In contrast, we provide a thorough study of the  benefit of our GAN formulation, compared to optimizing for classical losses such as MSE and MS-SSIM.
Additionally, their approach is very different: First, their GAN loss is non-standard,  operating on pairs of real/fake images classifying ``which one is the real one'', whereas ours has a principled interpretation in terms of divergences between probability distributions (as in \cite{goodfellow2014generative,nowozin2016f}). Second, their training uses various heuristics to balance the training, such as reweighting losses based on gradient magnitudes and alternating the training of the generator and discriminator based on manually defined thresholds on the losses. 

Santurkar \etal \cite{santurkar2017generative} use a GAN framework to learn a generative model over thumbnail images, which is then used as a decoder for thumbnail image compression. Other works use adversarial training for compression artifact removal (for engineered codecs) \cite{galteri2017deep} and single image super-resolution \cite{ledig2017photo}.
Finally, related to our SC mode, spatially allocating bitrate based on saliency of image content has a long history in the context of engineered compression algorithms, see, e.g., \cite{stella2009image,guo2010novel,gupta2013visual}.

\section{Background}

\paragraph{Generative Adversarial Networks:}
Given a data set $\Xc$, GANs can learn to approximate its (unknown) distribution $p_{\x}$ through a generator $G(\z)$ that tries to map samples $\z$ from a fixed prior distribution $p_{\z}$ to the data distribution  $p_{\x}$. The generator $G$ is trained in parallel with a discriminator $D$ by searching (using SGD) for a saddle point of a min-max objective $\min_G \Lc_{\text{GAN}}$ with
\begin{align}
\Lc_{\text{GAN}}:=\max_D  & \; \mathbb{E}[f(D({\x}))] + \mathbb{E}[g(D(G(\z)))] \label{eq:LGAN},
\end{align}
where $G$ and $D$ are DNNs and $f$ and $g$ are scalar functions.
Nowozin \etal \cite{nowozin2016f} show that for suitable choices of $f$ and $g$ solving $\min_G \Lc_{\text{GAN}}$ allows to minimize general $f$-divergences between the distribution of $G(z)$ and $p_x$.
We adapt  
Least-Squares GAN \cite{mao2017least} in this paper, where
$f(y)=(y-1)^2$ and $g(y)=y^2$ (which corresponds to the Pearson $\chi^2$ divergence).

\paragraph{Conditional Generative Adversarial Networks:}\label{sec:condgan}
For conditional GANs (cGANs) \cite{goodfellow2014generative,mirza2014conditional}, 
each data point $\x$ is associated with additional information $\s$, where $(\x,\s)$ have an unknown joint distribution $p_{\x,\s}$. 
We now assume that $\s$ is given and that we want to use the GAN to model the conditional distribution $p_{\x|\s}$. In this case, both the generator $G(\z,\s)$ and discriminator $D(\z,\s)$ have access to the side information $\s$, leading to the divergence
\begin{align*}
\Lc_{\text{cGAN}}:= \max_D  & \; \mathbb{E}[f(D({\x,\s}))] + \mathbb{E}[g(D(G(\z,\s),\s))]. 
\end{align*}

\paragraph{Deep Image Compression:}\label{sec:compression}
To compress an image $\x\in\Xc$, we follow the formulation of \cite{agustsson2017soft,mentzer2018conditional} where one learns an encoder $E$, a decoder $G$, and a finite quantizer $q$. The encoder $E$ maps the image to a latent feature map $\w$, whose values are then quantized to $L$ levels $\mathcal C = 
\{c_1,\ldots,c_L\}\subset\R$ to obtain a representation $\hat{\w}=q(E(\x))$ that can be encoded to a bitstream. The decoder then tries to recover the image by forming a reconstruction $\hat{\x}=G(\hat{\w})$. To be able to backpropagate through the non-differentiable $q$, one can use a differentiable relaxation of $q$, as in \cite{mentzer2018conditional}.

The average number of bits needed to encode $\hat{\w}$ is measured by the entropy $H(\hat{\w})$, which can be modeled with a prior \cite{agustsson2017soft} or a conditional probability model \cite{mentzer2018conditional}. The so called ``rate-distortion'' trade-off between reconstruction quality and bitrate to be optimized is then
\begin{equation}
\E[ d(\x,\hat{\x})] + \beta H(\hat{\w}). \label{eq:rd_tradeoff}
\end{equation}
where $d$ is a loss that measures how perceptually similar $\hat{\x}$ is to $\x$. Given a differentiable estimator of the entropy $H(\hat{\w})$, the weight $\beta$ controls the bitrate of the model.
However, since the number of dimensions $\text{dim}(\hat{\w})$ and the number of levels $L$ are finite, the entropy is bounded by (see, e.g.,~\cite{cover2012elements})
\begin{equation} \label{eq:entub}
H(\hat{\w}) \leq \text{dim}(\hat{\w}) \log_2(L).
\end{equation} 
It is therefore also valid to set $\beta=0$ and control the maximum bitrate through the bound \eqref{eq:entub} (i.e., adjusting $L$ and/or $\text{dim}(\hat{\w})$ through the architecture of $E$). While potentially leading to suboptimal bitrates, this avoids to model the entropy explicitly as a loss term.

\section{GANs for extreme image compression}
\subsection{Generative Compression}
\begin{figure}[h!]
    \centering
    \includegraphics[width=0.8\linewidth]{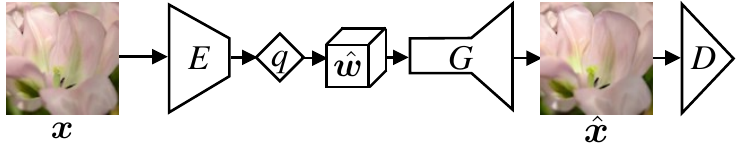}
    \caption{Architecture of our GC network.}
    \label{fig:gc_arch}
\end{figure}
The proposed GAN framework for extreme image compression can be viewed as a combination of (conditional) GANs and learned compression as introduced in the previous section. 
See Fig.~\ref{fig:gc_arch} for an overview of the architecture. With an encoder $E$ and quantizer $q$, we encode the image $\x$ to a compressed representation $\hat{\w}=q(E(\x))$. This representation is optionally concatenated with noise $\v$ drawn from a fixed prior $p_{\v}$, to form the latent vector $\z$.
The decoder/generator $G$ then tries to generate an image $\hat{\x}=G(\z)$ that is consistent with the image distribution $p_{\x}$ while also recovering the specific encoded image $\x$ to a certain degree.
Using $\z=[\hat{\w},\v]$, this can be expressed by our saddle-point objective for (unconditional) generative compression,
\begin{align}
    \min_{E,G} \max_D  \quad 
        &\mathbb{E}[f(D({\x}))] + \mathbb{E}[g(D(G({\z}))] \nonumber \\
        & + \lambda\E[ d(\x,G({\z}))] + \beta H(\hat{\w}), \label{eq:GANpression}
\end{align}
where $\lambda > 0$ balances the distortion term against the GAN loss and entropy terms.\footnote{In this formulation, we need to encode a real image to sample from $p_{\hat{\w}}$. However, this is not a limitation, as our goal is compressing real images, not generating completely new ones.}

Since the last two terms of \eqref{eq:GANpression} do not depend on the discriminator $D$, they do not affect its optimization directly.
This means that the discriminator still computes the same $f$-divergence $\Lc_{\text{GAN}}$ as in \eqref{eq:LGAN}, so we can write \eqref{eq:GANpression} as 
\begin{align}
\min_{E,G} & \quad  \Lc_{\text{GAN}}+ \lambda \E[d(\x,G(\z))] + \beta H(\hat{\w})\label{eq:GANpression2}.
\end{align}
We note that equation \eqref{eq:GANpression2} has completely different dynamics than a normal GAN, because the latent space $\z$ contains $\hat{\w}$, which stores information about a real image $\x$. 

The bitrate limitation on $H(\hat{\w})$ is a crucial element. 
If we allow $\hat{\w}$ to contain arbitrarily many bits (using $\beta=0$ and $L$, $\mathrm{dim}(\hat{\w})$ large enough), $E$ and $G$ could learn to near-losslessly recover $\x$ from $G(\z)=G(q(E(\x)))$, such that the distortion term would vanish. 
In this case, the divergence between $p_{\x}$ and $p_{G(\z)}$ would also vanish and the GAN loss would have no effect.
On the other hand, if $H(\hat{\w}) \to 0$ (using $\beta=\infty$ or $\text{dim}(\hat{\w})=0$), $\hat{\w}$ becomes deterministic. In this setting, $\z$ is random and independent of $\x$ (through the $\v$ component) and the objective reduces to a standard GAN plus the distortion term, which then acts as a regularizer.

By constraining the entropy of $\hat{\w}$, $E$ and $G$ will never be able to make $d$ fully vanish. In this case, $E,G$ need to balance the GAN objective $\Lc_{\text{GAN}}$ and the distortion term $\lambda \E[d(\x,G({\z}))]$, which leads to $G(\z)$ on one hand looking ``realistic'', and on the other hand preserving the original image. For example, if there is a tree for which $E$ cannot afford to store the exact texture (and make $d$ small) $G$ can synthesize it to satisfy $\Lc_{\text{GAN}}$, instead of showing a blurry green blob. Thereby, the distortion term stabilizes GAN training and tends to prevent mode collapse (as mode collapse would lead to a very large distortion value).  We refer to this setting as \textit{generative compression} (GC).

As for the GANs described in Sec.~\ref{sec:condgan}, we can easily extend GC to a conditional case. We consider the setting where the additional information $\s$ for an image $\x$ is a semantic label map of the scene, but with a twist: Instead of feeding $\s$ to $E,G$ and $D$, we \emph{only give it to the discriminator} $D$  during training.
We refer to this setting as ``GC $(D^+)$''.
We emphasize that \emph{no semantics are needed} to encode or decode images with the trained models, in neither GC nor GC $(D^+)$ (since $E,G$ do not depend on $\s$).

Finally, we note that Eq.~\ref{eq:GANpression2} is similar to classical rate-distortion theory,  
where $H(\hat \w)$ is the rate/entropy term.
Regarding the interaction between the GAN loss and the MSE loss, we observe that the MSE loss stabilizes the training as it penalizes collapse of the GAN.

\subsection{Selective Generative Compression}
For GC and GC $(D^+)$, $E,G$ automatically navigate the trade-off between generation and preservation over the entire image, without any guidance.
We also consider a different setting, \textit{selective generative compression} (SC). Here, the network is guided in terms of what should be generated and what should be preserved. An overview of the network structure is given in Fig.~\ref{fig:model_sc} in Appendix~\ref{sec:net_arch}.

For simplicity, we consider a binary setting, where we construct a single-channel binary heatmap $\m$ of the same spatial dimensions as $\hat \w$. Regions of zeros correspond to regions that should be fully synthesized, regions of ones should be preserved.
However, since our task is compression, we constrain the fully synthesized regions to have the same semantics $\s$ as the original image $\x$. We assume the semantics $\s$ are separately stored, and feed them through a feature extractor $F$ before feeding them to the generator $G$. 
To guide the network with the semantics, we mask the (pixel-wise) distortion $d$, such that it is only computed over the region to be preserved. Additionally, we zero out the compressed representation $\hat{\w}$ in the regions that should be synthesized. Provided that the heatmap $\m$ is also stored, we then only encode the entries of $\hat{\w}$ corresponding to the preserved regions, greatly reducing the bitrate needed to store it. 
At bitrates where $\hat{\w}$ is much larger on average than the storage cost for $\s$ and $\m$, this approach can result in large bitrate savings.

We consider two different training modes: Random instance (RI) which randomly selects 25\% of the instances in the semantic label map and preserves these, and random box (RB) which picks an image location uniformly at random and preserves a box of random dimensions. 
While the RI mode is appropriate for most use cases, RB can create more challenging situations for the generator as it needs to integrate the preserved box seamlessly into generated content.

\subsection{PSNR and MS-SSIM as quality measures}\label{sec:psnrbad}

Our model targets realistic reconstructions where texture and sometimes even more abstract image content is synthesized. Common distortion measures such as PSNR and MS-SSIM cannot measure the ``realistic-ness'', as they penalize changes in local structure rather than assessing preservation of the global image content. This fact was \emph{mathematically} proven recently by \cite{blau2018perception}, showing the existence of a fundamental perception-distortion tradeoff, i.e., low distortion is at odds with high perceptual quality in the context of lossy reconstruction tasks. Intuitively, measuring PSNR between synthesized and real texture patches essentially quantifies the variance of the texture rather than the perceptual quality of the synthesized texture. This becomes apparent by comparing reconstructions produced by our GC model with those obtained by the MSE baseline and BPG in Fig.~\ref{fig:visualexgccityscapes}. While our reconstructions clearly look realistic, they have 4.2dB larger MSE than those of BPG. We therefore rely on human opinions collected in a thorough user study to evaluate our GC models.

\section{Experiments}

\subsection{Architecture, Losses, and Hyperparameters}
The architecture for our encoder $E$ and generator $G$ is based on the global generator network proposed in \cite{wang2018highres}, which in turn is based on the architecture of \cite{Johnson2016Perceptual}. We present details in Appendix~\ref{sec:net_arch}.

For the entropy term $\beta H(\hat{\w})$, we adopt the simplified approach described in Sec.~\ref{sec:compression}, where we set $\beta=0$, use $L=5$ centers $\mathcal{C}=\{-2,1,0,1,2\}$, and control the bitrate through the upper bound $H(\hat{\w}) \leq \text{dim}(\hat{\w}) \log_2(L)$.
For example, for GC, with $C=2$ bottleneck channels, we obtain $0.0181$bpp.\footnote{
    $\sfrac{H(\hat{\w})}{WH} \leq 
    \sfrac{ \frac{WH}{16 \cdot 16}\cdot C \cdot \log_2(L)}{WH} = 0.0181 \text{bpp}$, where $W,H$ are the dimensions of the image and 16 is the downsampling factor to the feature map, see Appendix~\ref{sec:net_arch}.}
We note that this is an upper bound; the actual entropy of $H(\hat{\w})$ is generally smaller, since the learned distribution will neither be uniform nor i.i.d, which would be required for the bound to hold with equality. 
We use an arithmetic encoder to encode the channels of $\hat\w$ to a bit-stream, storing frequencies for each channel separately (similar to \cite{agustsson2017soft}). In our experiments, this leads to $8.8\%$ smaller bitrates compared to the upper bound. We leave the exploration of context models to potentially further reduce the bitrate for future work.

For the distortion term $d$ we adopt MSE with $\lambda=10$.
Furthermore, we adopt the feature matching and VGG perceptual losses, $\Lc_{\text{FM}}$ and $\Lc_{\text{VGG}}$, as proposed in \cite{wang2018highres} with the same weights, which improved the quality for images synthesized from semantic label maps. These losses can be viewed as a part of $d(\x,\hat{\x})$. However, we do not mask them in SC, since they also help to stabilize the GAN in this operation mode (as in \cite{wang2018highres}). We refer to Appendix~\ref{sec:training} for training details.

\begin{figure}[t!]
\centering
    {\setlength{\tabcolsep}{1pt}
        \scriptsize
    \begin{tabular}{cl}
        \includegraphics[width=0.9\linewidth, valign=c]{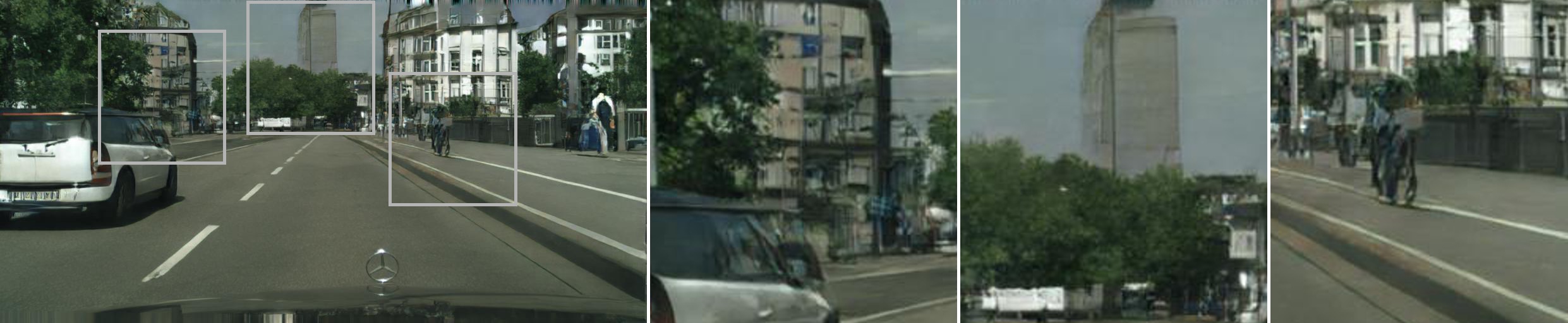}
        & \makecell[l]{Ours\\0.035\\ 21.8dB}\\[5.8ex]
        \includegraphics[width=0.9\linewidth, valign=c]{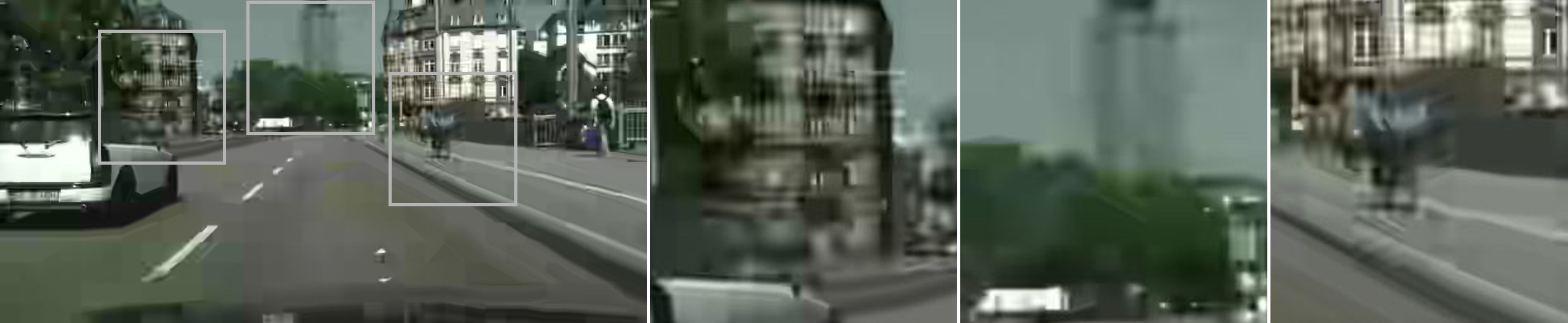}
        & \makecell[l]{BPG\\0.039\\ 26.0dB}\\[5.8ex]
        \includegraphics[width=0.9\linewidth, valign=c]{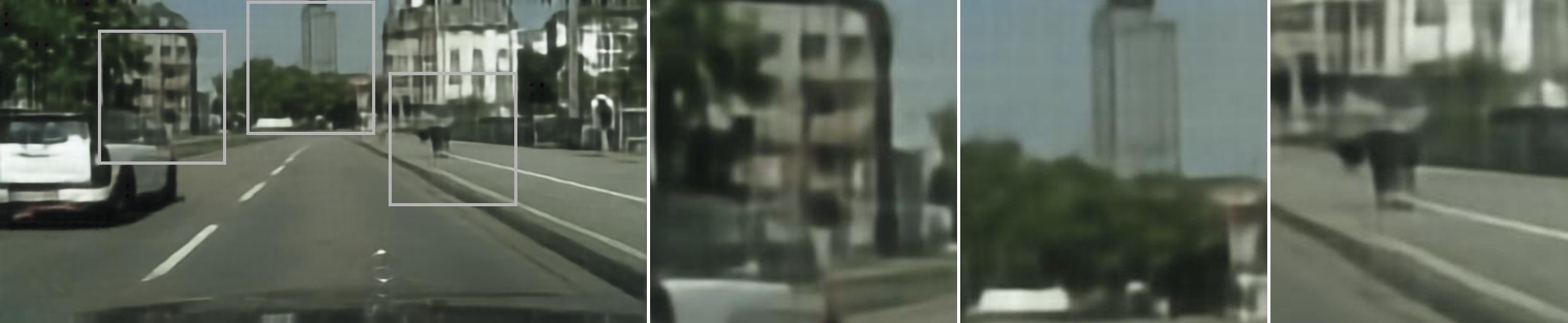}
        & \makecell[l]{MSE bl\\0.035\\ 24.0dB}\\
    \end{tabular}} 
\caption{\label{fig:visualexgccityscapes} Visual example of images produced by our GC network with $C=4$ bottleneck channels along with the corresponding results for BPG, and a baseline model with the same architecture ($C=4$) but trained for MSE only (MSE bl.), on Cityscapes. We show bitrate in bpp and PSNR in dB. The reconstruction of our GC network is sharper and has more realistic texture than BPG and MSE bl., even though the latter two have higher PSNR. In particular, the MSE bl.\ produces blurry reconstructions even though it was trained on the Cityscapes data set, demonstrating that domain-specific training alone is not enough to obtain sharp reconstructions at low bitrates.}
\end{figure}

\subsection{Evaluation} \label{sec:eval}

\paragraph{Data sets:} We train GC models (without semantic label maps) for compression of diverse natural images using 188k images from the \emph{Open Images} data set \cite{OpenImages} and evaluate them on the widely used Kodak image compression data set \cite{kodakurl} as well as 20 randomly selected images from the \emph{RAISE1K} data set \cite{dang2015raise}. To investigate the benefits of having a somewhat constrained application domain and semantic information at training time, we also train GC models with semantic label maps on the \emph{Cityscapes} data set \cite{2016cityscapes}, using 20 randomly selected images from the validation set for evaluation. 
To evaluate the proposed SC method (which requires semantic label maps for training and deployment) we again rely on the Cityscapes data set. Cityscapes was previously used to generate images form semantic label maps using GANs \cite{isola2017image,zhu2017unpaired}.
 
\paragraph{Baselines:} We compare our method to the HEVC-based image compression algorithm BPG~\cite{bpgurl} (in the 4:2:2 chroma format) and to the AEDC network from~\cite{mentzer2018conditional}. BPG is the current state-of-the-art engineered image compression codec and outperforms other recent codecs such as JPEG2000 and WebP on different data sets in terms of PSNR (see, e.g. \cite{balle2018variational}). 
We train the AEDC network (with bottleneck depth  $C=4$) for MS-SSIM  on Cityscapes exactly following the procedure in~\cite{mentzer2018conditional} except that we
use early stopping to prevent overfitting (note that Cityscapes is much smaller than the ImageNet data set used in~\cite{mentzer2018conditional}). The so-obtained model has a bitrate of 0.07 bpp and gets a slightly better MS-SSIM than BPG at the same bpp on the validation set. To investigate the effect of the GAN term in our total loss, we train a baseline model with an MSE loss only (with the same architecture as GC and the same training parameters, see Sec. \ref{sec:training} in the Appendix), referred to as ``MSE baseline''.

\paragraph{User study:} 
Given that classical distortion metrics like PSNR or MS-SSIM are not suited for the task we study here (Section~\ref{sec:psnrbad}), we quantitatively evaluate the perceptual quality of our GC models in comparison with BPG and AEDC (for Cityscapes) with a user study on Amazon Mechanical Turk (AMT).\footnote{\url{https://www.mturk.com/}} We consider two GC models with $C=4,8$ bottleneck channels trained on Open Images, three GC $(D^+)$ models with $C=2,4,8$ trained on Cityscapes, and BPG at rates ranging from $0.045$ to $0.12$ bpp. Questionnaires are composed by combining the reconstructions produced by the selected GC model for all testing images with the corresponding reconstruction produced by the competing baseline model side-by-side (presenting the reconstructions in random order). The original image is shown along with the reconstructions, and the pairwise comparisons are interleaved with 3 probing comparisons of an additional uncompressed image from the respective testing set with an obviously JPEG-compressed version of that image. 20 randomly selected unique users are asked to indicate their preference for each pair of reconstructions in the questionnaire, resulting in a total of 480 ratings per pairing of methods for Kodak, and 400 ratings for RAISE1K and Cityscapes. For each pairing of methods, we report the mean preference score as well as the standard error (SE) of the per-user mean preference percentages. Only users correctly identifying the original image in all probing comparisons are taken into account for the mean preference percentage computation.
To facilitate comparisons for future works, we will release all images used in the user studies.

\paragraph{Semantic quality of SC models:} The issues with PSNR and MS-SSIM described in Sec.~\ref{sec:psnrbad} become even more severe for SC models as a large fraction of the image content is generated from a semantic label map. Following image translation works \cite{isola2017image,wang2018highres}, we therefore measure the capacity of our SC models to preserve the image semantics in the synthesized regions and plausibly blend them with the preserved regions---the objective SC models are actually trained for. Specifically, we use PSPNet \cite{2016pspnet} and compute the mean intersection-over-union (IoU) between the label map obtained for the decompressed validation images and the ground truth label map. For reference we also report this metric for baselines that do not use semantic label maps for training and/or deployment.

\section{Results} \label{sec:results}

\subsection{Generative compression}  \label{sec:gcresults}

\begin{figure}[t]
    \centering
    {\setlength{\tabcolsep}{1pt}
        \scriptsize
    \begin{tabular}{cc}
        \includegraphics[width=0.49\linewidth]{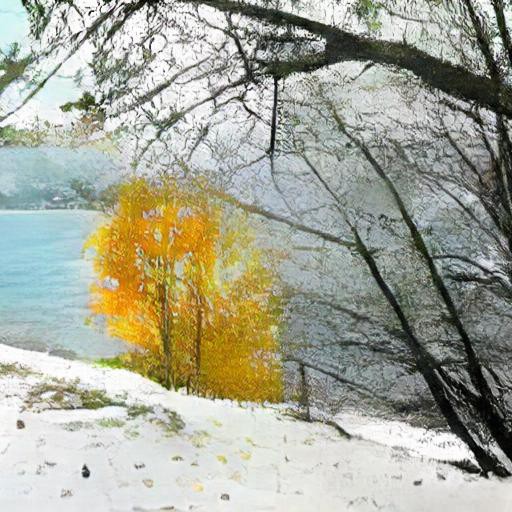}&
        \includegraphics[width=0.49\linewidth]{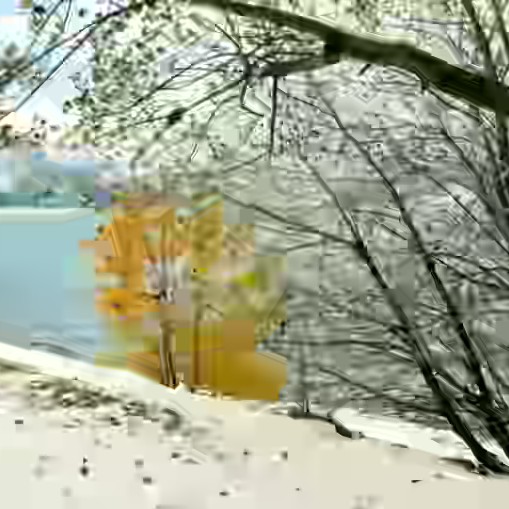}\\[-2pt]
        \includegraphics[width=0.49\linewidth]{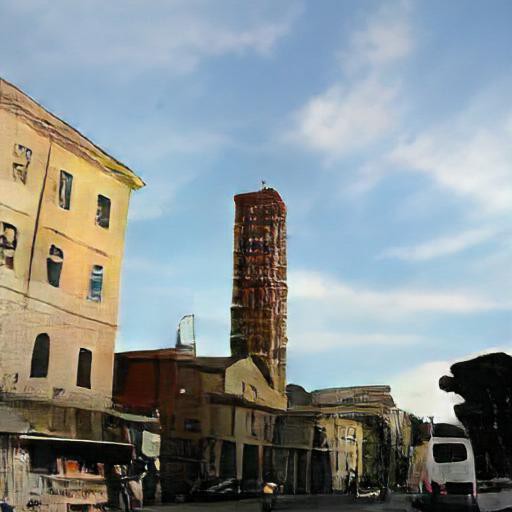} &
        \includegraphics[width=0.49\linewidth]{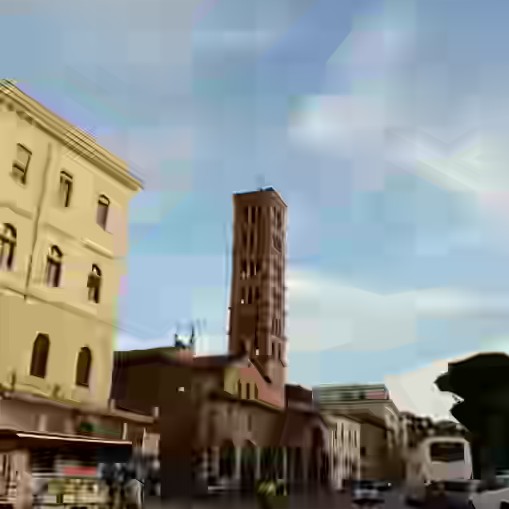}\\[-2pt]
        Ours $0.0341$bpp & BPG $0.102$bpp 
    \end{tabular}} 
    \vspace{0.1cm}
    \caption{\label{fig:visualexgc_raise1k} Visual example of an image from RAISE1k produced by our GC network with $C=4$ compared to BPG.}
    \vspace{-0.2cm}
\end{figure}

\begin{figure}[t!]
\centering
        \footnotesize
         \includegraphics[width=\linewidth]{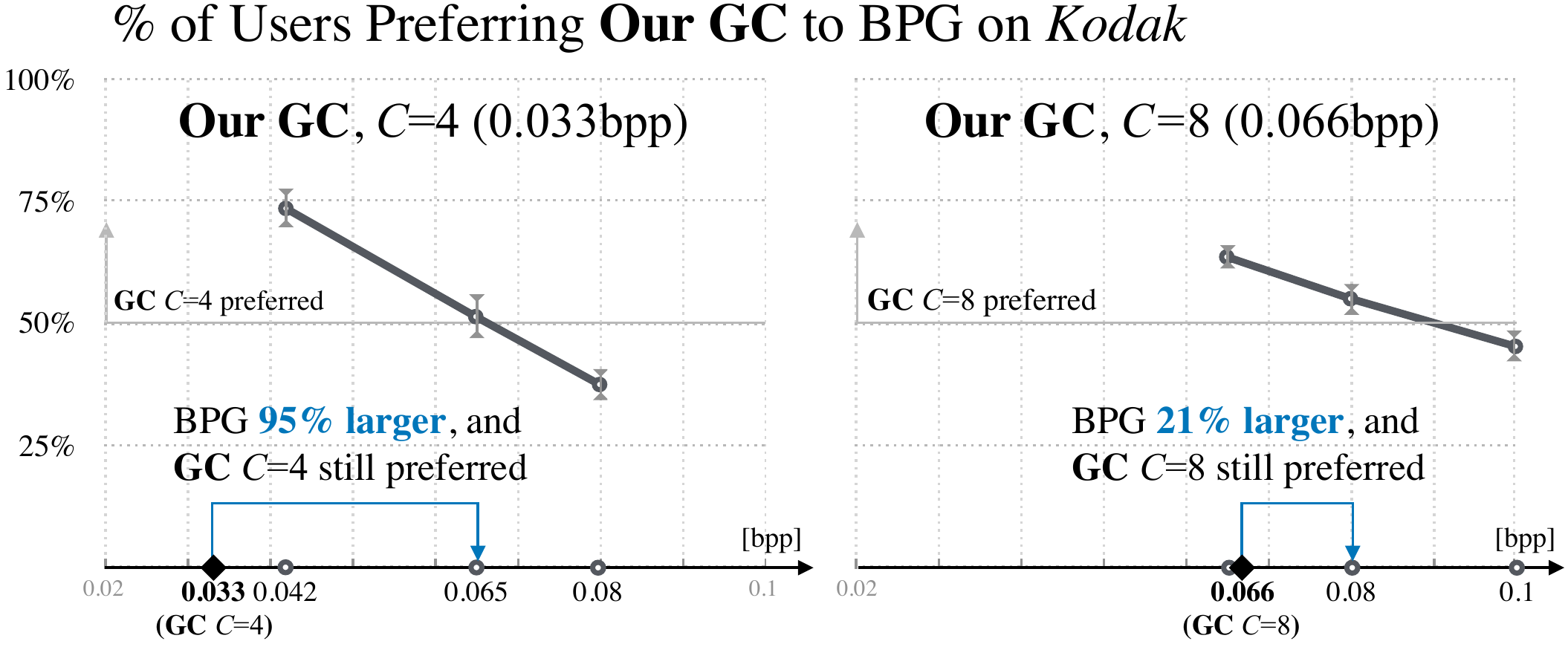}\vspace{0.1cm}
         \includegraphics[width=\linewidth]{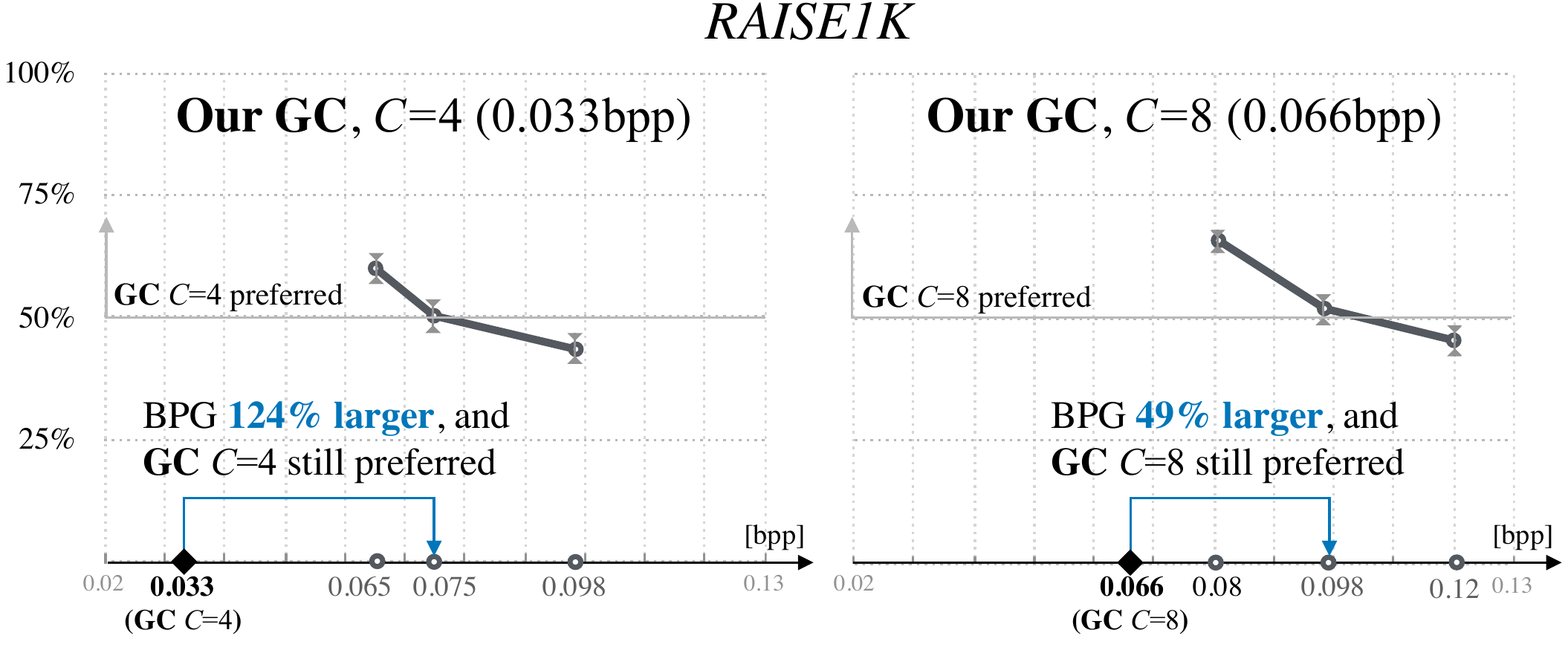}\vspace{0.1cm}
        \includegraphics[width=\linewidth]{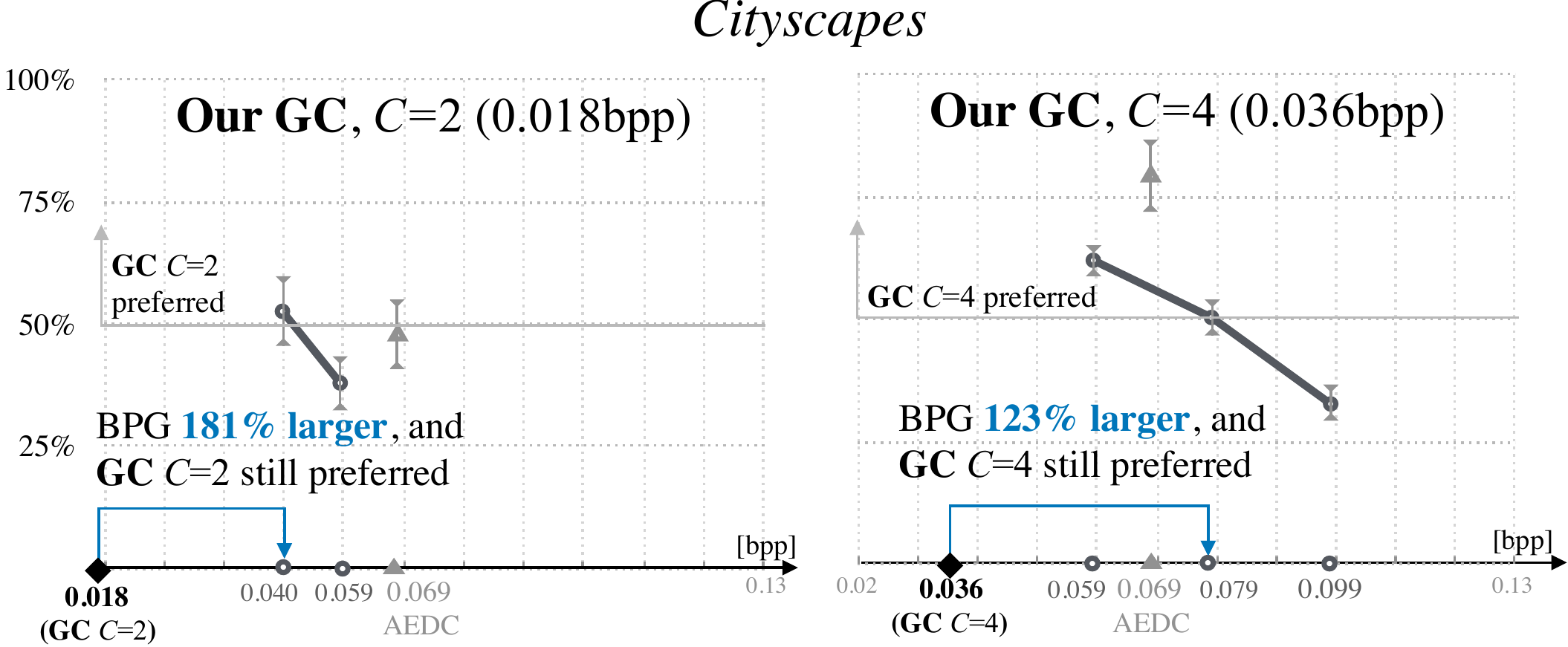}
\caption{\label{fig:pref_all}User study results evaluating our GC models on Kodak, RAISE1K and Cityscapes.
Each plot corresponds to one of our models. The bitrate of that model is highlighted on the x-axis with a black diamond. The thick gray line shows the percentage of users preferring our model to BPG at that bitrate (bpp). The blue arrow points from our model to the highest-bitrate BPG operating point where more than 50\% of users prefer ours, visualizing how many more bits BPG uses at that point.
For Kodak and RAISE1K, we use GC models trained on Open Images, without any semantic label maps. For Cityscapes, we used GC $(D^+)$ (using semantic label maps only for $D$ and only during training), and we additionally compared to the AEDC baseline (MS-SSIM optimized).\vspace{0.2cm}
}
\end{figure}

Fig.~\ref{fig:pref_all} shows the mean preference percentage obtained by our GC models compared to BPG at different rates, on the Kodak and the RAISE1K data set. In addition, we report the mean preference percentage for GC models compared to BPG and AEDC on Cityscapes. Example validation images for side-by-side comparison of our method with BPG for images from the Kodak, RAISE1K, and Cityscapes data set can be found in Figs.~\ref{fig:visualexgckodak},~\ref{fig:visualexgc_raise1k}, and~\ref{fig:visualexgccityscapes}, respectively. 
Furthermore, we perform extensive visual comparisons of all our methods and the baselines, presented in Appendix~\ref{sec:visuals}.

Our GC models with $C=4$ are preferred to BPG even when images produced by BPG use 95\% and 124\% more bits than those produced by our models for Kodak and RAISE1K, respectively. Notably this is achieved even though there is a distribution shift between the training and testing set (recall that these GC models are trained on the Open Images data set). 
The gains of domain-specificity and semantic label maps (for training) becomes apparent from the results on Cityscapes: Our GC models with $C=2$ are preferred to BPG even when the latter uses 181\% more bits. For $C=4$ the gains on Cityscapes are comparable to those obtained for GC on RAISE1K. For all three data sets, BPG requires between 21\% and 49\% more bits than our GC models with $C=8$.

\begin{figure}[h!]
    \centering
        \scriptsize
{\setlength{\tabcolsep}{1pt}
        \begin{tabular}{ccc}
            \includegraphics[width=0.2475\linewidth]{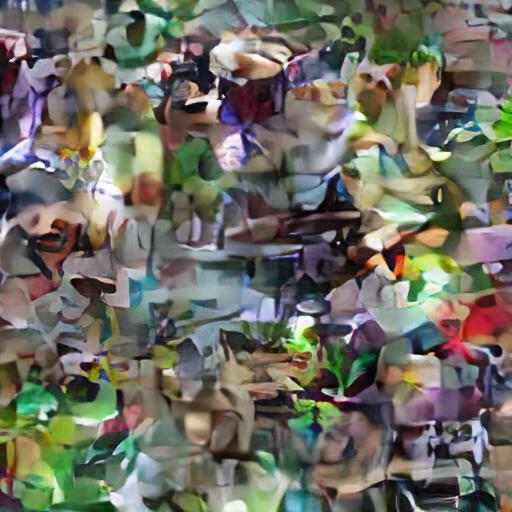} &
            \includegraphics[width=0.2475\linewidth]{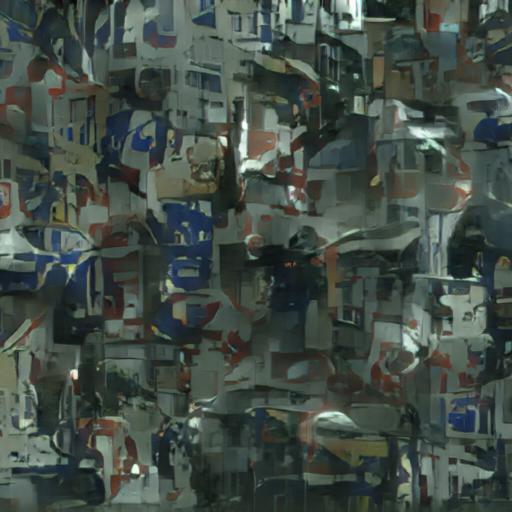}  &
            \includegraphics[width=0.495\linewidth]{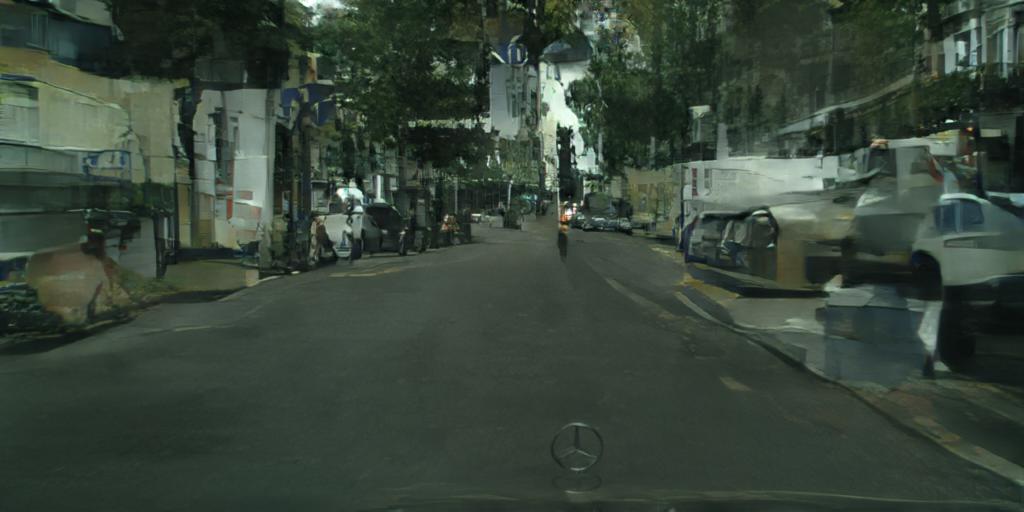} \\                  
            $\mathcal{U}$ (Open Images) & $\mathcal{U}$ (Cityscapes) & WGAN-GP (Cityscapes) 
        \end{tabular}}
        \caption{\label{fig:explore_samples} Sampling codes $\hat \w$ uniformly ($\mathcal{U}$, left), and generating them with a WGAN-GP (right).}
\end{figure}

\paragraph{Discussion:}
The GC models produce images with much finer detail than BPG, which suffers from smoothed patches and blocking artifacts. In particular, the GC models convincingly reconstruct texture in natural objects such as trees, water, and sky, and is most challenged with scenes involving humans. AEDC and the MSE baseline both produce blurry images.

We see that the gains of our models are maximal at extreme bitrates, with BPG needing 95--181\% more bits for the $C=2,4$ models on the three data sets. For $C=8$ gains are smaller but still very large (BPG needing 21--49\% more bits). This is expected, since as the bitrate increases the classical compression measures (PSNR/MS-SSIM) become more meaningful---and our system does not employ the full complexity of current state-of-the-art systems:

We give an overview of relevant recent learned compression methods and their differences to our GC method and BPG in Table~\ref{tab:soa_comparison} in Appendix~\ref{app:compsota}, where we see that BPG is still visually competitive with the current state-of-the-art. 

Given the dramatic bitrate savings we achieve according to the user study (BPG needing 21--181\% more bits), and the competitiveness of BPG to the most recent state-of-the-art \cite{minnen2018joint}, we conclude that our proposed system presents a \textbf{significant step forward} for visually pleasing compression at extreme bitrates.

\paragraph{Sampling the compressed representations:} In Fig.~\ref{fig:explore_samples} we explore the representation learned by our GC models (with $C=4$), by sampling the (discrete) latent space of $\hat{\w}$.
When we sample uniformly, and decode with our GC model into images, we obtain a ``soup of image patches'' which reflects the domain the models were trained on (e.g. street sign and building patches on Cityscapes).
Note that we should not expect these outputs to look like normal images, since nothing forces the encoder output $\hat{\w}$ to be uniformly distributed over the discrete latent space.

However, given the low dimensionality of $\hat{\w}$ ($32\times64\times4$ for $512\times1024$px Cityscape images), it would be interesting to try to learn the true distribution.
To this end, we perform a simple experiment and train an improved Wasserstein GAN (WGAN-GP) \cite{gulrajani2017improved} on $\hat{\w}$ extracted from Cityscapes, using default parameters and a ResNet architecture (only adjusting the architecture to output $32\times64\times4$ tensors instead of $64\times64\times3$ RGB images). By feeding our GC model with samples from the WGAN-GP generator, we easily obtain a powerful generative model, which generates sharp $1024\times512$px images \emph{from scratch}. We think this could be a promising direction for building high-resolution generative models.  
In Figs.~\ref{fig:sample_bn_openimages}--\ref{fig:sample_bn_gan} in the Appendix, we show more samples, and samples obtained by feeding the MSE baseline with uniform and learned code samples. The latter yields noisier ``patch soups'' and much blurrier image samples than our GC network.

\begin{figure}[t]
    \centering
    \includegraphics[width=0.9\linewidth]{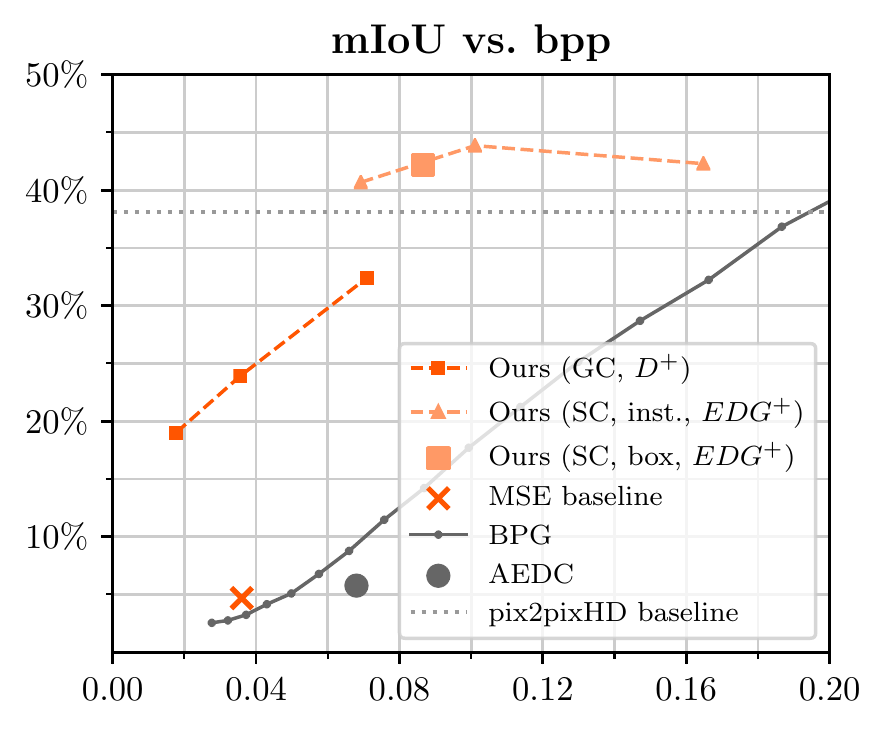} 
    \caption{\label{fig:miouplot} Mean IoU as a function of bpp on the Cityscapes validation set for our GC and SC networks, and for the MSE baseline. We show both SC modes: RI (inst.), RB (box). 
$D^+$ annotates models where instance semantic label maps are fed to the discriminator (only during training); $EDG^+$ indicates that semantic label maps are used both for training and deployment. The pix2pixHD baseline~\cite{wang2018highres} was trained from scratch for 50 epochs, using the same downsampled $1024\times512$px training images as for our method. } \vspace{-0.2cm}
\end{figure}

\begin{figure*}[t!]
\vspace{-0.1cm}
{\scriptsize
\centering
    {\setlength{\tabcolsep}{1pt}
    \input{fig_full_classes.tex}}
    }
    \vspace{0.2cm}
\caption{\label{fig:visualexsc}Synthesizing different classes using our SC network with $C=8$. 
In each image except for \emph{no synthesis}, we additionally synthesize the classes \emph{vegetation, sky, sidewalk, ego vehicle, wall}. The heatmaps in the lower left corners show the synthesized parts in gray.
We show the bpp of each image 
as well as the relative savings due to the selective generation.}
\vspace{-0.3cm}
\end{figure*}
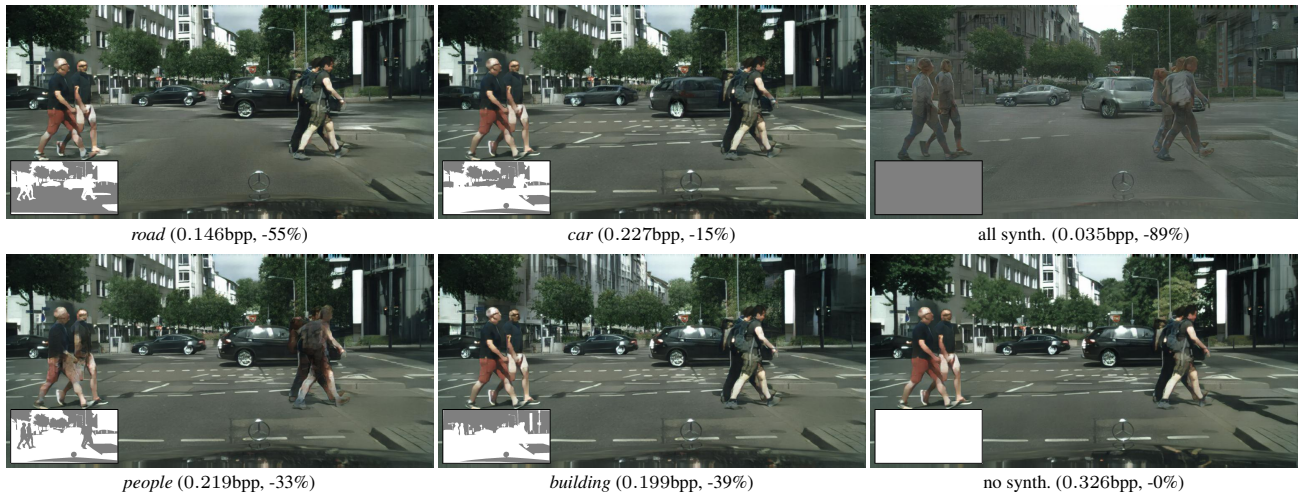

\subsection{Selective generative compression} 

Fig. \ref{fig:miouplot} shows the mean IoU on the Cityscapes validation set as a function of bpp for SC networks with $C=2, 4, 8$, along with the values obtained for the baselines. Additionally, we plot mean IoU for GC with semantic label maps fed to the discriminator ($D^+$), and the MSE baseline. 

In Fig.~\ref{fig:visualexsc} we present example Cityscapes validation images produced by the SC network trained in the RI mode with $C=8$, where different semantic classes are preserved. 
More visual results for the SC networks trained on Cityscapes can be found in Appendix~\ref{sec:sccity}, including results obtained for the RB operation mode and by using semantic label maps estimated from the input image via PSPNet \cite{zhao2017pspnet}.

\paragraph{Discussion:} The quantitative evaluation of the semantic preservation capacity (Fig. \ref{fig:miouplot}) reveals that the SC networks preserve the semantics somewhat better than pix2pixHD, indicating that the SC networks faithfully generate texture from the label maps and plausibly combine generated with preserved image content. The mIoU of BPG, AEDC, and the MSE baseline is considerably lower than that obtained by our SC and GC models, which can arguably be attributed to blurring and blocking artifacts. However, it is not surprising as these baseline methods do not use label maps during training and prediction. 

In the SC operation mode, our networks manage to seamlessly merge preserved and generated image content both when preserving object instances and boxes crossing object boundaries (see Appendix~\ref{sec:sccity}). Further, our networks lead to reductions in bpp of 50\% and more compared to the same networks without synthesis, while leaving the visual quality essentially unimpaired, when objects with repetitive structure are synthesized (such as trees, streets, and sky). In some cases, the visual quality is even better than that of BPG at the same bitrate. The visual quality of more complex synthesized objects (e.g. buildings, people) is worse. However, this is a limitation of current GAN technology rather than our approach. As the visual quality of GANs improves further, SC networks will as well. Notably, the SC networks can generate entire images from the semantic label map only.

Finally, the semantic label map, which requires 0.036 bpp on avg.\ for downscaled $1024{\times}512$px Cityscapes images, represents a relatively large overhead compared to the storage cost of the preserved image parts. This cost vanishes as the image size increases, since the semantic mask can be stored as an image dimension-independent vector graphic.

\vspace{-1ex}
\section{Conclusion}
We proposed a GAN-based framework for learned generative compression, and presented the first thorough study of such a framework for full-resolution image compression.
Our results show that for low bitrates, such generative compression (GC) can give dramatic bitrate savings compared to previous state-of-the-art methods optimized for classical objectives such as MS-SSIM and MSE, when evaluated in terms of visual quality in a user study.
Furthermore, we demonstrated that constraining the application domain to street scene images leads to additional storage savings, and explored (for SC) selectively combining fully synthesized image contents with preserved ones when semantic label maps are available.

Interesting directions for future work are to develop a mechanism for controlling spatial allocation of bits for GC (e.g.,\ to achieve better preservation of faces; possibly using semantic label maps), and to combine SC with saliency information to determine what regions to preserve. 

{\footnotesize
\noindent\textbf{Acknowledgments:} This work was supported by the ETH Zurich General Fund, and an Nvidia GPU hardware grant.
}

\bibliography{egbib}
\clearpage
\pagebreak

\appendix

\begin{table*}[t!]
        \centering
        \begin{tabular}{l|llll}
            & BPG                &   Rippel \etalnospace~(2017)             & Minnen \etalnospace~(2018)           & Ours (GC) \\ 
            \hline
Learned                    &No                  & Yes                                & Yes                           & Yes  \\
Arithmetic encoding        &Adaptive            & Adaptive                           & Adaptive                      & Static \\
Context model              &CABAC               & Autoregressive                     & Autoregressive                & None   \\
Visualized bitrates [bpp]\tablefootnote{This refers to the bitrates of decoded images the authors have made available.}\
                           & all\tablefootnote{Code available, image can be compressed from extreme bpps ($<0.1$bpp) to lossless.} & 0.08--                                 & 0.12--                       & 0.033--0.066 \\
GAN                        & No                 & Non-standard                       & No                            & f-div. based \\
S.o.t.a. in MS-SSIM        & No\                & No \                               & Yes                           & No     \\
S.o.t.a. in PSNR           & No                 & No                                 & Yes                           & No     \\
            Savings to BPG in PSNR            &        &                                    & 8.41\%                        &        \\
            Savings to BPG in User Study      &        &                                    &                               & 17.2--48.7\%  \\
        \end{tabular}
        \vspace{0.2cm}
        \caption{Overview of differences between \cite{minnen2018joint} (s.o.t.a.\ in MS-SSIM and PSNR), to BPG (previous s.o.t.a.\ in PSNR) and \cite{rippel17a} (s.o.t.a. in MS-SSIM in 2017, also used GANs).
        }
        \label{tab:soa_comparison}
\end{table*}

\section{Comparison with State-of-the-art} \label{app:compsota}
We give an overview of relevant recent learned compression methods and their differences to our GC method and BPG in Table~\ref{tab:soa_comparison}. 
\cite{rippel17a} were state-of-the-art in MS-SSIM in 2017, while work \cite{minnen2018joint} is the current state-of-the-art in image compression in terms of classical metrics (PSNR and MS-SSIM) when measured on the Kodak data set \cite{kodakurl}. Notably, all methods except ours (BPG, Rippel \etalnospace, and Minnen \etalnospace) employ adaptive arithmetic coding using context models for improved compression performance.
Such models could also be implemented for our system, and have led to additional savings of 10\% in \cite{mentzer2018conditional}. Since Rippel \etalnospace\ and Minnen \etalnospace\ have only released a selection of their decoded images (for 3 and 4, respectively, out of the 24 Kodak images), and at significantly higher bitrates, a comparison with a user study is not meaningful.
Instead, we try to qualitatively put our results into context with theirs.

\begin{figure}[h!]
    \centering
    {\setlength{\tabcolsep}{5pt}
    \begin{tabular}{ccc}
        \includegraphics[width=0.3\linewidth]{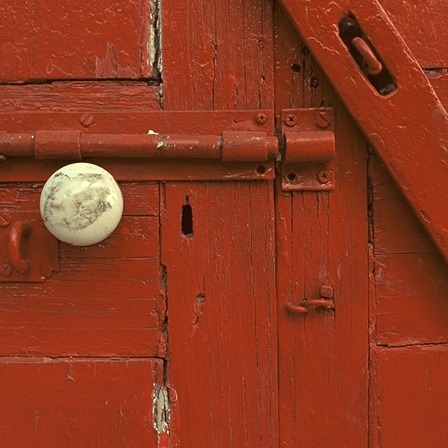} &
        \includegraphics[width=0.3\linewidth]{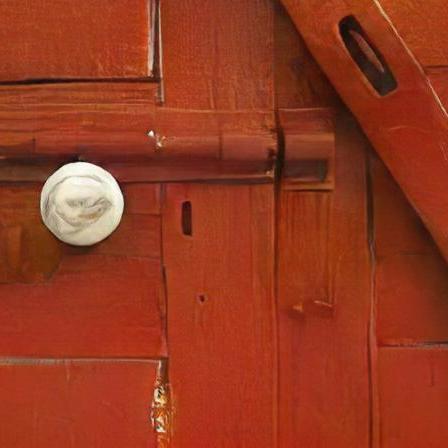}&
        \includegraphics[width=0.3\linewidth]{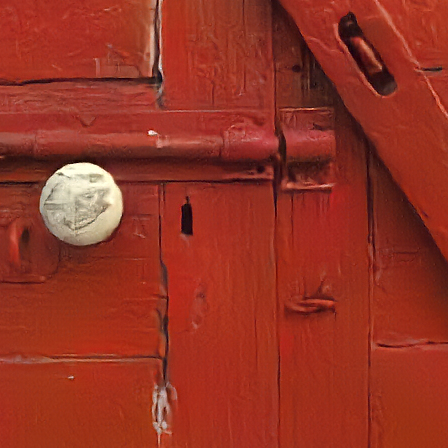} \\
        Original & Ours, $0.0304$bpp & Rippel \etalnospace, $0.0828$bpp (\emph{+172\%}) \\
    \end{tabular}}
    \vspace{0.2cm}
    \caption{\label{fig:rippel1}Our model loses more texture but has less artifacts on the knob. Overall, it looks comparable to the output of \cite{rippel17a}, using significantly fewer bits.}
\end{figure}

\begin{figure}[h!]
    \centering
    {\setlength{\tabcolsep}{5pt}
    \begin{tabular}{ccc}
        \includegraphics[width=0.3\linewidth]{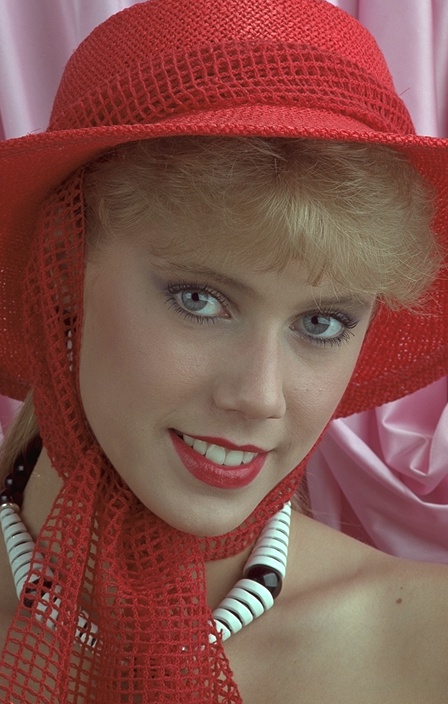} &
        \includegraphics[width=0.3\linewidth]{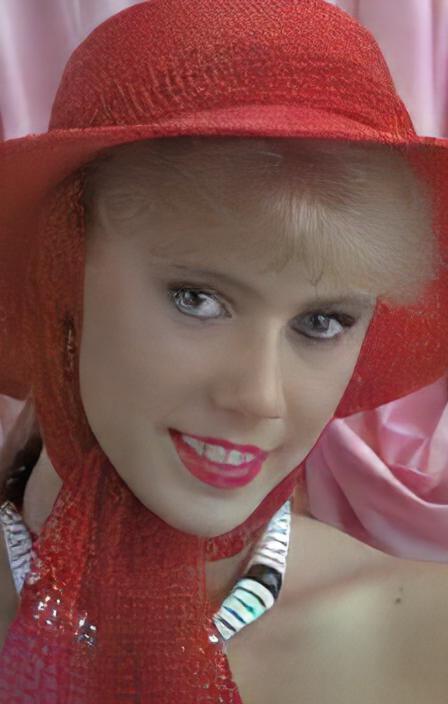}&
        \includegraphics[width=0.3\linewidth]{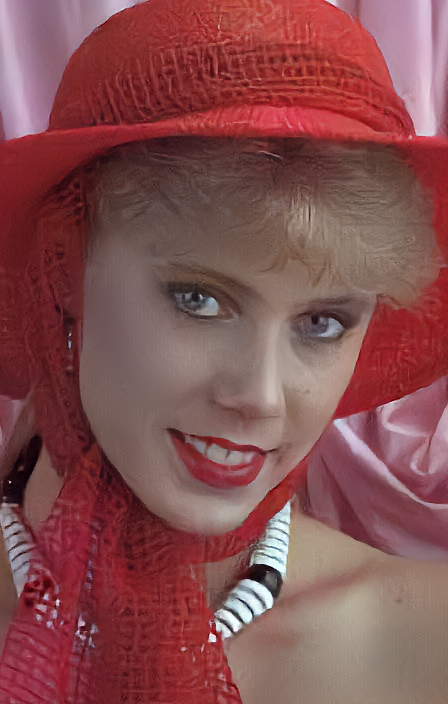} \\
        Original & Ours, $0.0651$bpp & Rippel \etalnospace, $0.0840$bpp (\emph{+29\%}) \\
    \end{tabular}}
    \vspace{0.2cm}
    \caption{\label{fig:rippel2}Notice that compared to \cite{rippel17a}, our model produces smoother lines at the jaw and a smoother hat, but provides a worse reconstruction of the eye.}
\end{figure}

\begin{figure}[hb!]
    \centering
    {\setlength{\tabcolsep}{5pt}
    \begin{tabular}{ccc}
        \includegraphics[width=0.3\linewidth]{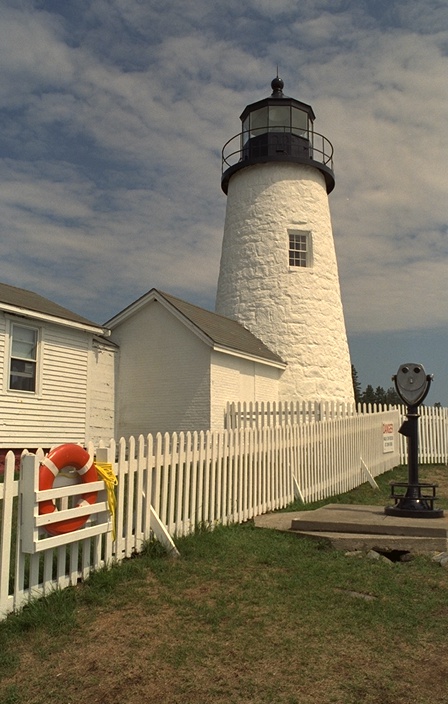} &
        \includegraphics[width=0.3\linewidth]{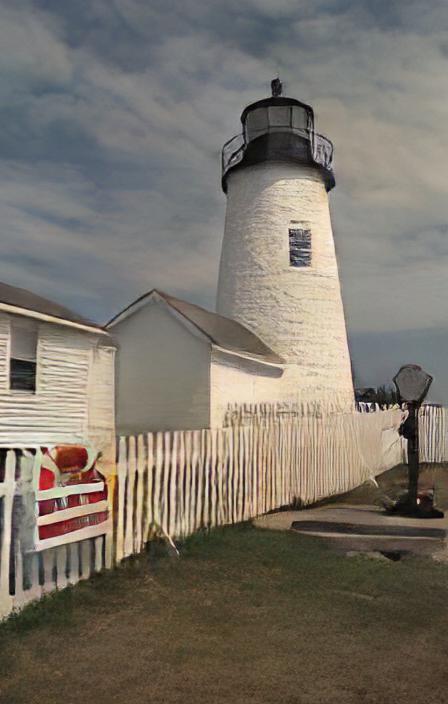}&
        \includegraphics[width=0.3\linewidth]{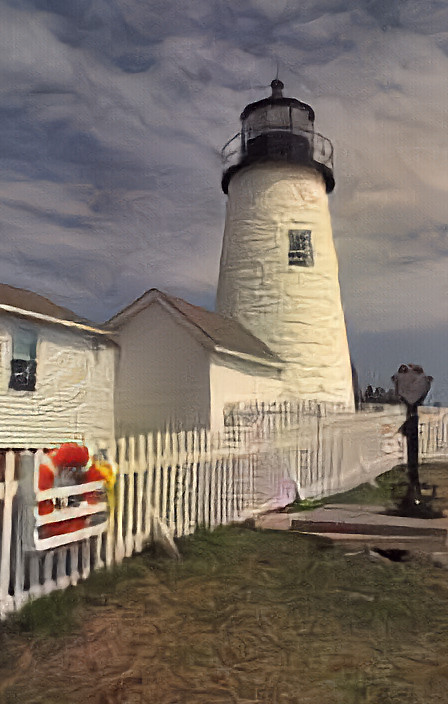} \\
        Original & Ours, $0.0668$bpp & Rippel \etalnospace, $0.0928$bpp (\emph{+39\%}) \\
    \end{tabular}}
    \vspace{0.2cm}
    \caption{\label{fig:rippel3}Notice that our model produces much better sky and grass textures than \cite{rippel17a}, and also preserves the texture of the light tower more faithfully.}
\end{figure}

In Figs.~\ref{fig:rippel1}--\ref{fig:rippel3} in Sec.~\ref{sec:comp_rippel}, we compare qualitatively to \cite{rippel17a}.
We can observe that even though Rippel \etal \cite{rippel17a} use 29--179\% more bits, our models produce images of comparable or better quality.

\begin{figure}[h!]
\vspace{-0.2cm}
    \centering
    {\setlength{\tabcolsep}{5pt}
    \begin{tabular}{ll}
        \includegraphics[width=0.38\linewidth]{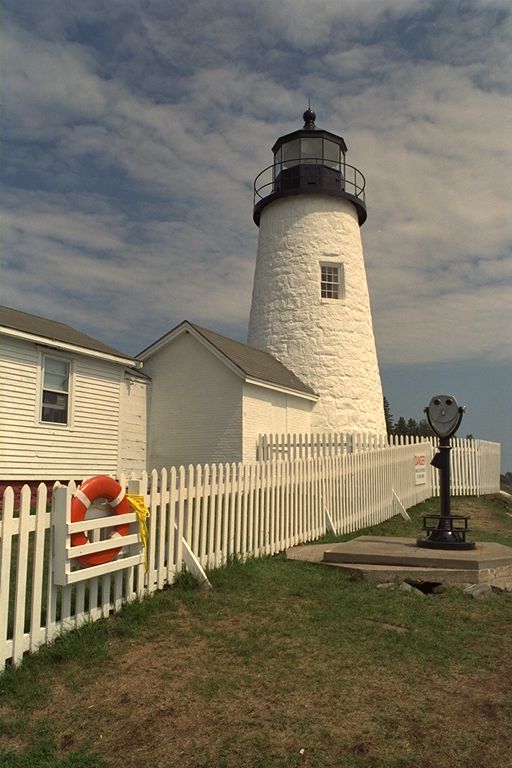} &
        \includegraphics[width=0.38\linewidth]{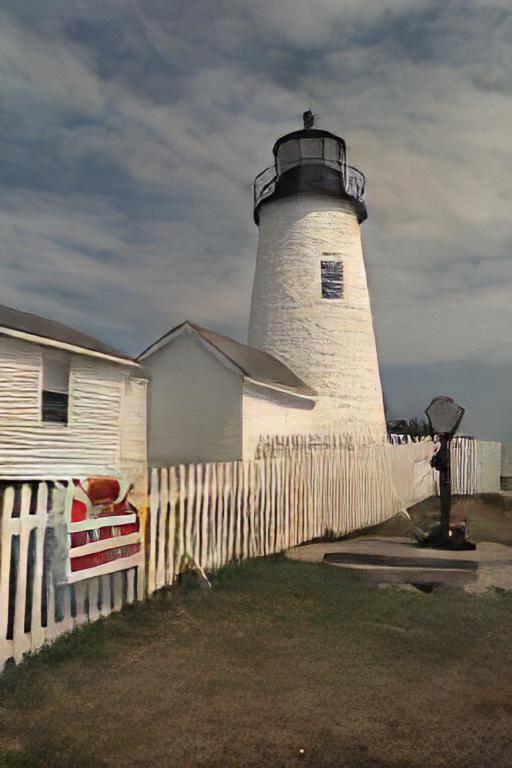}\\
        Original & Ours, $0.0668$bpp        \\[0.2cm] 
        \includegraphics[width=0.38\linewidth]{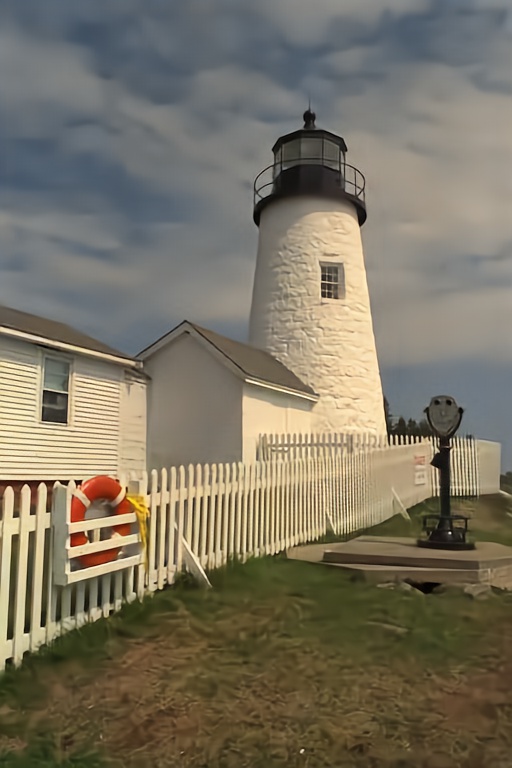} &
        \includegraphics[width=0.38\linewidth]{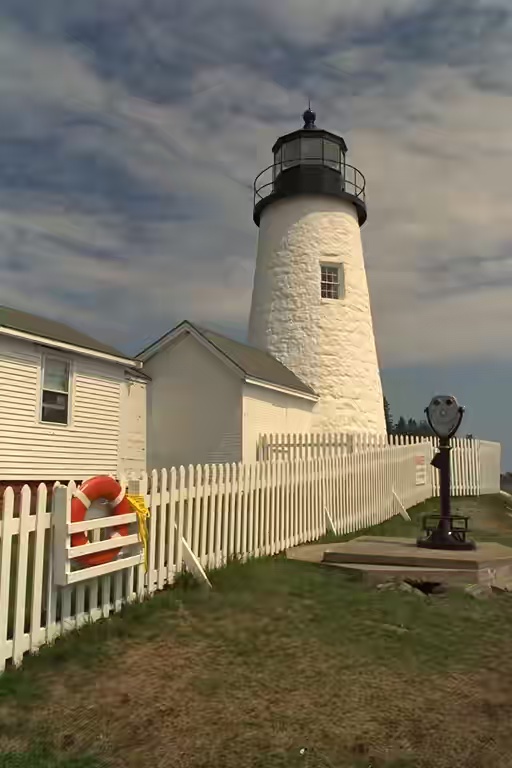} \\
        Minnen \etalnospace, $0.221$bpp \emph{230\% larger} & BPG, $0.227$bpp \\
    \end{tabular}}
    \vspace{0.2cm}
    \caption{\label{fig:minnen1}Notice that our model yields sharper grass and sky, but a worse reconstruction of the fence and the lighthouse compared to \cite{minnen2018joint}.  Compared to BPG, Minnen \etalnospace\ produces blurrier grass, sky and lighthouse but BPG suffers from ringing artifacts on the roof of the second building and the top of the lighthouse.}
    \vspace{-1cm}
\end{figure}

\begin{figure}[hb!]
    \centering
    {\setlength{\tabcolsep}{1pt}
    \begin{tabular}{ll}
        \includegraphics[width=0.5\linewidth]{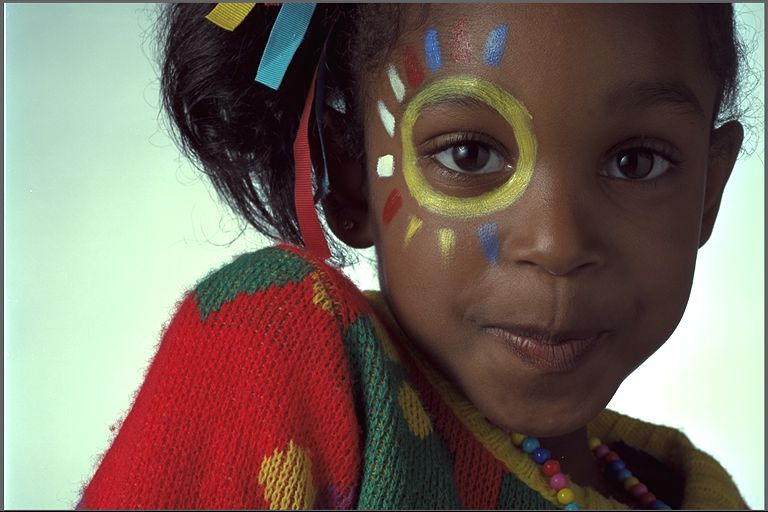} &
        \includegraphics[width=0.5\linewidth]{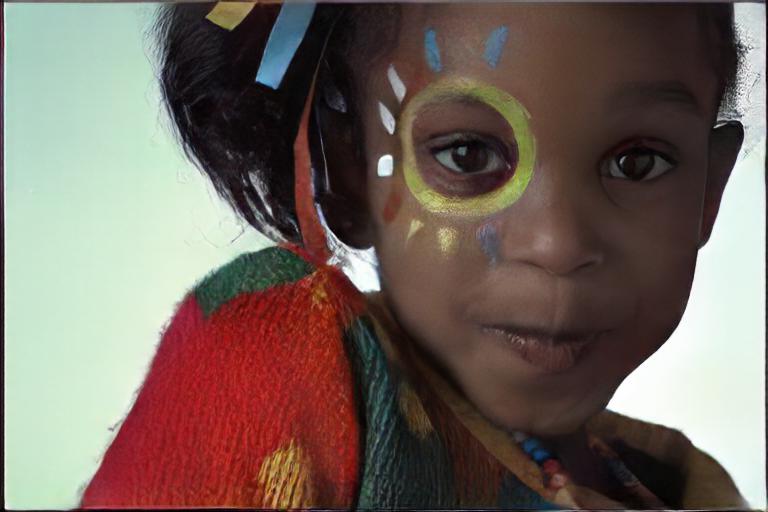} \\
        Original & Ours, $0.0685$bpp       \\[0.2cm] 
        \includegraphics[width=0.5\linewidth]{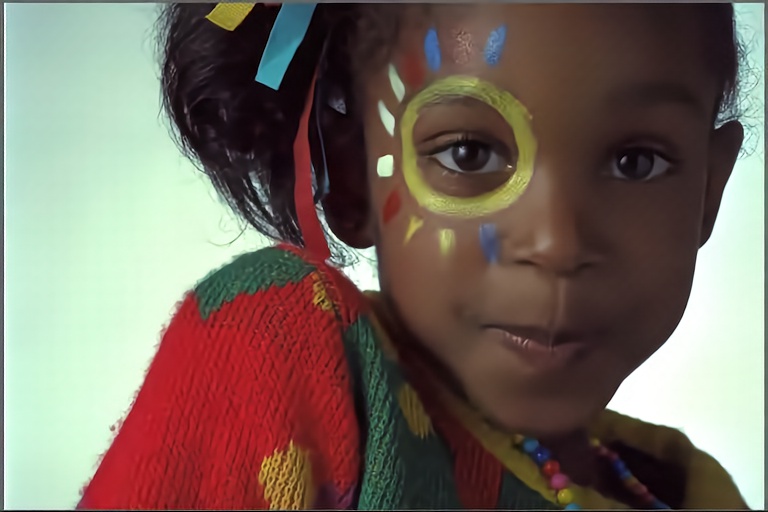} &
        \includegraphics[width=0.5\linewidth]{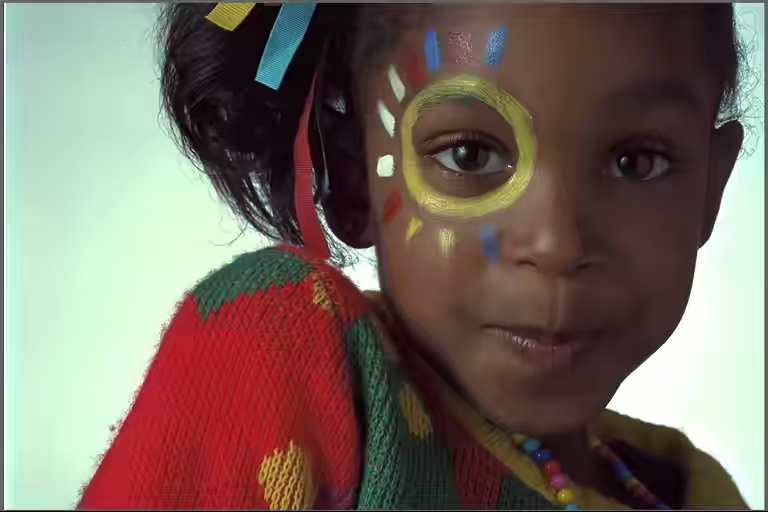}\\
        Minnen \etalnospace, $0.155$bpp, \emph{127\% larger}                    & BPG, $0.164$bpp    \\
    \end{tabular}}
    \vspace{0.2cm}
    \caption{\label{fig:minnen2}Our model produces an overall sharper face compared to \cite{minnen2018joint}, but the texture on the cloth deviates more from the original. Compared to BPG, Minnen \etalnospace\  has a less blurry face and fewer artifacts on the cheek.}
\end{figure}

\begin{figure}[hb!]
    \centering
    {\setlength{\tabcolsep}{1pt}
    \begin{tabular}{ll}
        \includegraphics[width=0.5\linewidth]{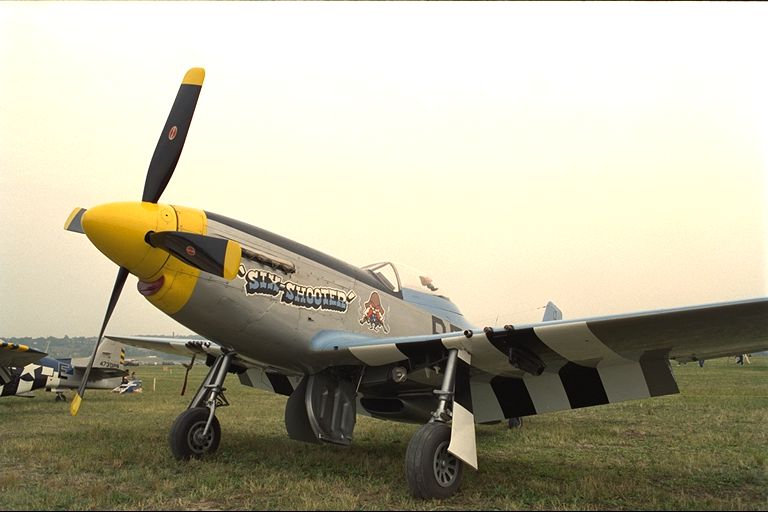} &
        \includegraphics[width=0.5\linewidth]{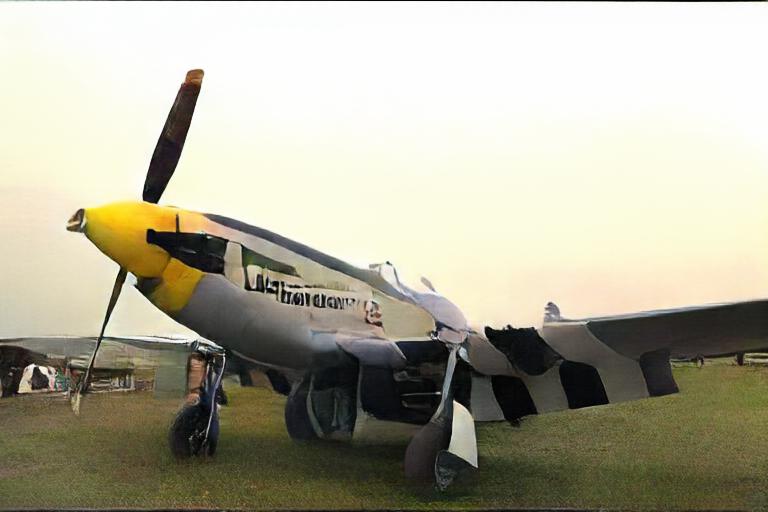} \\
        Original & Ours, $0.0328$bpp        \\[0.2cm] 
        \includegraphics[width=0.5\linewidth]{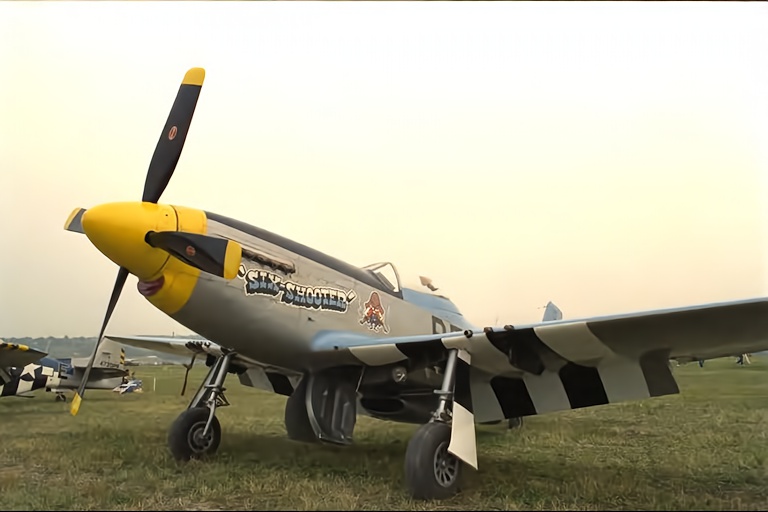} &
        \includegraphics[width=0.5\linewidth]{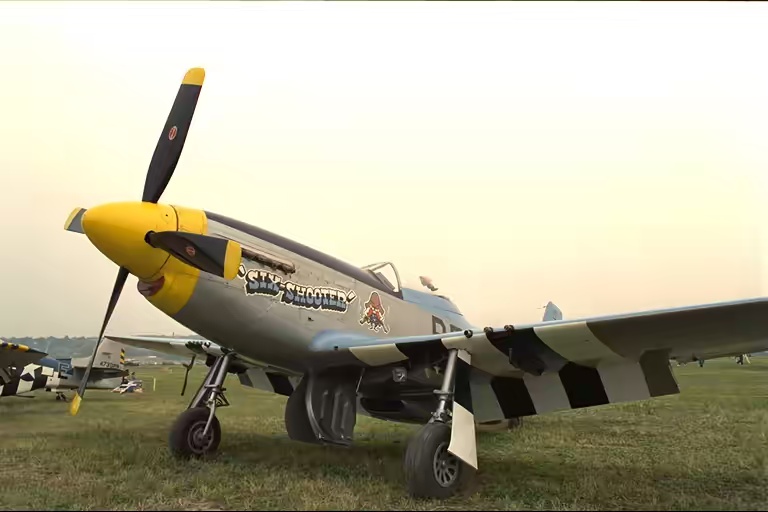}\\
        Minnen \etalnospace, $0.246$bpp, \emph{651\% larger}                   & BPG, $0.248$bpp  \\
    \end{tabular}}
    \vspace{0.2cm}
    \caption{\label{fig:minnen3}Here we obtain a significantly worse reconstruction than \cite{minnen2018joint}  and BPG, but use only a fraction of the bits. Between BPG and Minnen \etalnospace, it is hard to see any differences.}
\end{figure}

\begin{figure}[hb!]
    \centering
    {\setlength{\tabcolsep}{1pt}
    \begin{tabular}{ll}
        \includegraphics[width=0.5\linewidth]{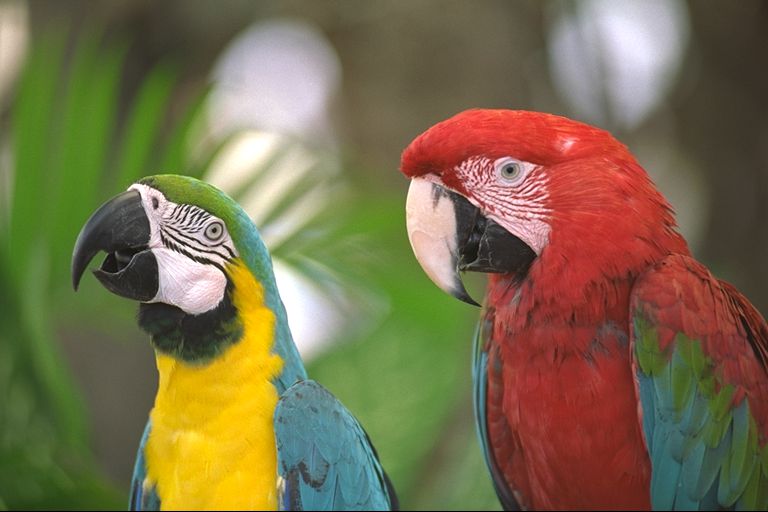} &
        \includegraphics[width=0.5\linewidth]{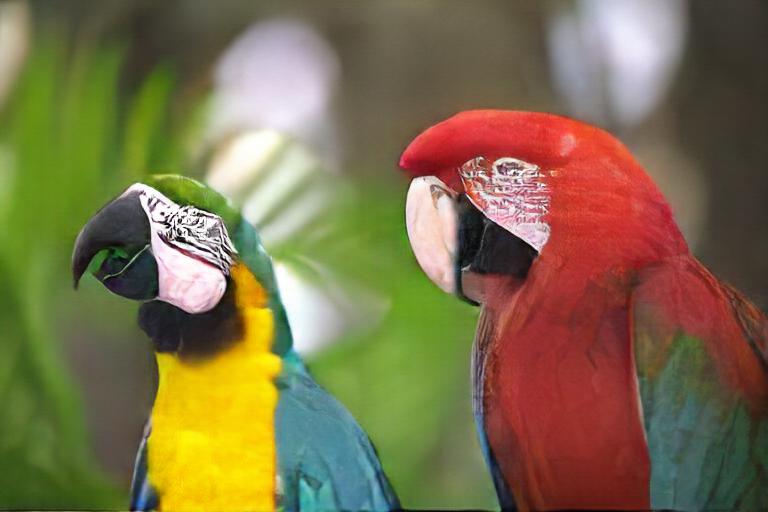}\\
        Original & Ours, $0.03418$bpp \\[0.2cm] 
        \includegraphics[width=0.5\linewidth]{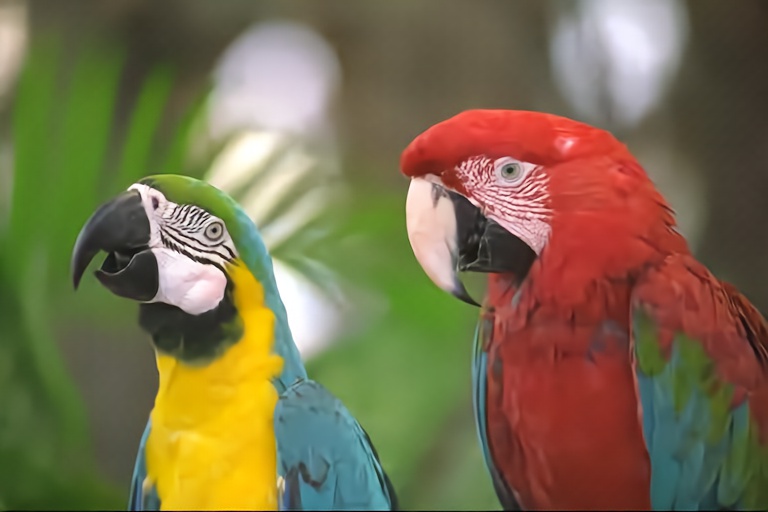} &
        \includegraphics[width=0.5\linewidth]{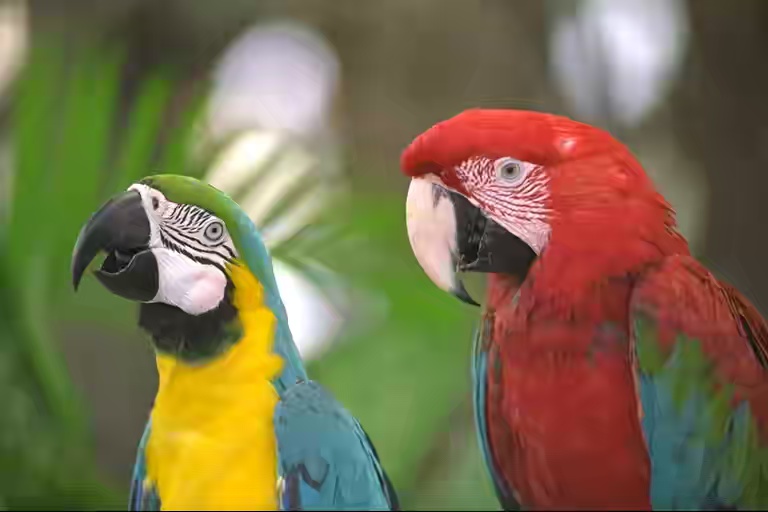}\\
        Minnen \etalnospace, $0.123$bpp, \emph{259\% larger},                    & BPG, $0.119$bpp  \\
    \end{tabular}}
    \vspace{0.2cm}
    \caption{\label{fig:minnen4}Here we obtain a significantly worse reconstruction compared to \cite{minnen2018joint} and BPG, but use only a fraction of the bits. Compared to BPG, Minnen \etalnospace  has a smoother background but less texture on the birds.}
\end{figure}

In Figs.~\ref{fig:minnen1}--\ref{fig:minnen4} in Sec.~\ref{sec:comp_minnen}, we show a qualitative comparison of our results to the images provided by the work of \cite{minnen2018joint}, as well as to BPG~\cite{bpgurl} on those images.
First, we see that BPG is still visually competitive with the current state-of-the-art, which is consistent with moderate $8.41\%$ bitrate savings being reported by \cite{minnen2018joint} in terms of PSNR.
Second, even though we use much fewer bits compared to the example images available from~\cite{minnen2018joint}, for some of them (Figs.~\ref{fig:minnen1} and \ref{fig:minnen2}) our method can still produce images of comparable visual quality.

\section{Training Details} \label{sec:training}

We employ the ADAM optimizer \cite{kingmaB14} with a learning rate of $0.0002$ and set the mini-batch size to $1$. Our networks are trained for 150000 iterations on Cityscapes and for 280000 iterations on Open Images. For normalization we used instance normalization~\cite{instancenorm}, except in the second half of the Open Images training, we train the generator/decoder with fixed batch statistics (as implemented in the test mode of batch normalization~\cite{ioffe2015batch}), since we found this reduced artifacts and color shift.

\section{Data set and Preprocessing Details} \label{sec:datadetails}

To train GC models (which do not require semantic label maps, neither during training nor for deployment) for compression of diverse natural images, we use 200k images sampled randomly from the \emph{Open Images} data set \cite{OpenImages} (9M images). The training images are rescaled so that the longer side has length $768$px, and images for which rescaling does not result in at least $1.25\times$ downscaling as well as high saturation images (average $\text{S}>0.9$ or $\text{V}>0.8$ in HSV color space) are discarded (resulting in an effective training set size of 188k).

We evaluate these models on the Kodak image compression data set \cite{kodakurl} (24 images, $768 \times 512$px), which has a long tradition in the image compression literature and is still the most frequently used data set for comparisons of learned image compression methods. Additionally, we evaluate our GC models on 20 randomly selected images from the \emph{RAISE1K} data set \cite{dang2015raise}, a real-world image data set consisting of 8156 high-resolution RAW images (we rescale the images such that the longer side has length $768$px). To investigate the benefits of having a somewhat constrained application domain and semantic labels at training time, we also train GC models with semantic label maps on the \emph{Cityscapes} data set \cite{2016cityscapes} (2975 training and 500 validation images, 34 classes, $2048 \times 1024$px resolution) consisting of street scene images and evaluate it on 20 randomly selected validation images (without semantic labels). Both training and validation images are rescaled to $1024 \times 512$px resolution.

To evaluate the proposed SC method (which requires semantic label maps for training and deployment) we again rely on the Cityscapes data set. Cityscapes was previously used to generate images form semantic label maps using GANs \cite{isola2017image,zhu2017unpaired}. The preprocessing for SC is the same as for GC.

\section{Compression Details} \label{app:compression_details}

We compress the semantic label map for SC by quantizing the coordinates in the vector graphic to the image grid and encoding coordinates relative to preceding coordinates when traversing object boundaries (rather than relative to the image frame). The so-obtained bitstream is then compressed using arithmetic coding.

To ensure fair comparison, we do not count header sizes for any of the baseline methods throughout.

\section{Architecture Details} \label{sec:net_arch}

For the GC, the encoder $E$ convolutionally processes the image $\x$ and optionally the label map $\s$, with spatial dimension $W\times H$, into a feature map of size $\sfrac{W}{16} \times \sfrac{H}{16} \times 960$ (with 6 layers, of which four have $2$-strided convolutions), which is then projected down to $C$ channels (where $C\in \{2,4,8\}$ is much smaller than $960$).
This results in a feature map $\w$ of dimension $\sfrac{W}{16} \times \sfrac{H}{16} \times C$, which is quantized over $L$ centers to obtain the discrete $\hat{\w}$.
The generator $G$ projects $\hat{\w}$ up to $960$ channels, processes these with $9$ residual units \cite{He_2016_CVPR} at dimension $\sfrac{W}{16} \times \sfrac{H}{16} \times 960$, and then mirrors $E$ by convolutionally processing the features back to spatial dimensions $W\times H$ (with transposed convolutions instead of strided ones).

Similar to $E$, the feature extractor $F$ for SC processes the semantic map $\s$ down to the spatial dimension of $\hat{\w}$, which is then concatenated to $\hat{\w}$ for generation. In this case, we consider slightly higher bitrates and downscale by $8\times$ instead of $16\times$ in the encoder $E$, such that $\text{dim}(\hat{\w})=\sfrac{W}{8} \times \sfrac{H}{8} \times C$. The generator then first processes $\hat{\w}$ down to $\sfrac{W}{16} \times \sfrac{H}{16} \times 960$ and then proceeds as for GC.

For both GC and SC, we use the multi-scale architecture of \cite{wang2018highres} for the discriminator $D$, which measures the divergence between $p_{\x}$ and $p_{G(\z)}$ both locally and globally.

We adopt the notation from \cite{wang2018highres} to describe our encoder and generator/decoder architectures and additionally use {\tt q} to denote the quantization layer (see Sec. \ref{sec:compression} for details). The output of {\tt q} is encoded and stored.

\begin{itemize}[leftmargin=*]
\item {\bf Encoder GC:} {\tt c7s1-60, d120, d240, d480, d960, c3s1-$C$, q}

\item {\bf Encoders SC:}
\begin{itemize}[leftmargin=*]
\item Semantic label map encoder:
{\tt c7s1-60, d120, d240, d480, d960}

\item Image encoder:
{\tt c7s1-60, d120, d240, d480, c3s1-$C$, q, c3s1-480, d960}
\end{itemize}

The outputs of the semantic label map encoder and the image encoder are concatenated and fed to the generator/decoder.

\item{\bf Generator/decoder:}
{\tt c3s1-960, R960, R960, R960, R960, R960, R960, R960, R960, R960, u480, u240, u120, u60, c7s1-3}
\end{itemize}

\begin{figure}[h]
    \centering
    \includegraphics[width=\linewidth]{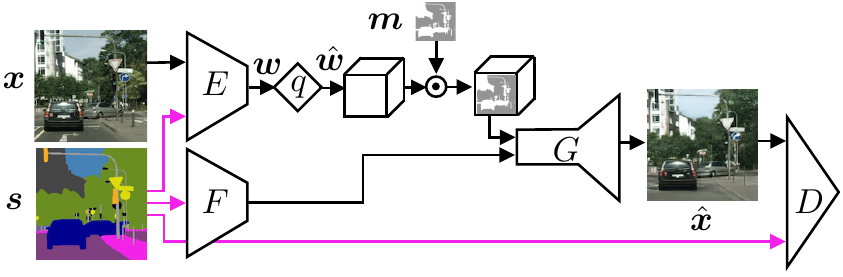}%
    \caption{\label{fig:model_sc}Structure of the proposed SC network. $E$ is the encoder for the image $\x$ and the semantic label map $\s$. $q$ quantizes the latent code $\w$ to $\hat \w$. The subsampled heatmap multiplies $\hat \w$ (pointwise) for spatial bit allocation. $G$ is the generator/decoder, producing the decompressed image $\hat \x$, and $D$ is the discriminator used for adversarial training. $F$ extracts features from $\s$ .}\end{figure}

\section{Visuals}\label{sec:visuals}

In the following Sections, \ref{sec:gc_kodak}, \ref{sec:gc_raise}, \ref{sec:gc_cs}, we show the first five images of each of the three data sets we used for the user study, next to the outputs of BPG at similar bitrates.

Secs.~\ref{sec:comp_rippel} and \ref{sec:comp_minnen} provide visual comparisons of our GC models with~\cite{rippel17a} and \cite{minnen2018joint}, respectively, on a subset of images form the Kodak data set.

In Sec.~\ref{sec:vis_repr}, we show visualizations of the latent representation of our GC models.

Finally, Sec.~\ref{sec:sccity} presents additional visual results for SC.

\clearpage
\onecolumn
\subsection{Generative Compression on Kodak}  \label{sec:gc_kodak}
\begin{figure}[h!]
\vspace{-0.3cm}
\centering
\resizebox{.79\linewidth}{!}{
\input{fig_appendix_kodak.tex}
}
\vspace{0.2cm}
\caption{First $5$ images of the Kodak data set, produced by our GC model with $C=4$ and BPG.}
\vspace{-2cm}
\end{figure}
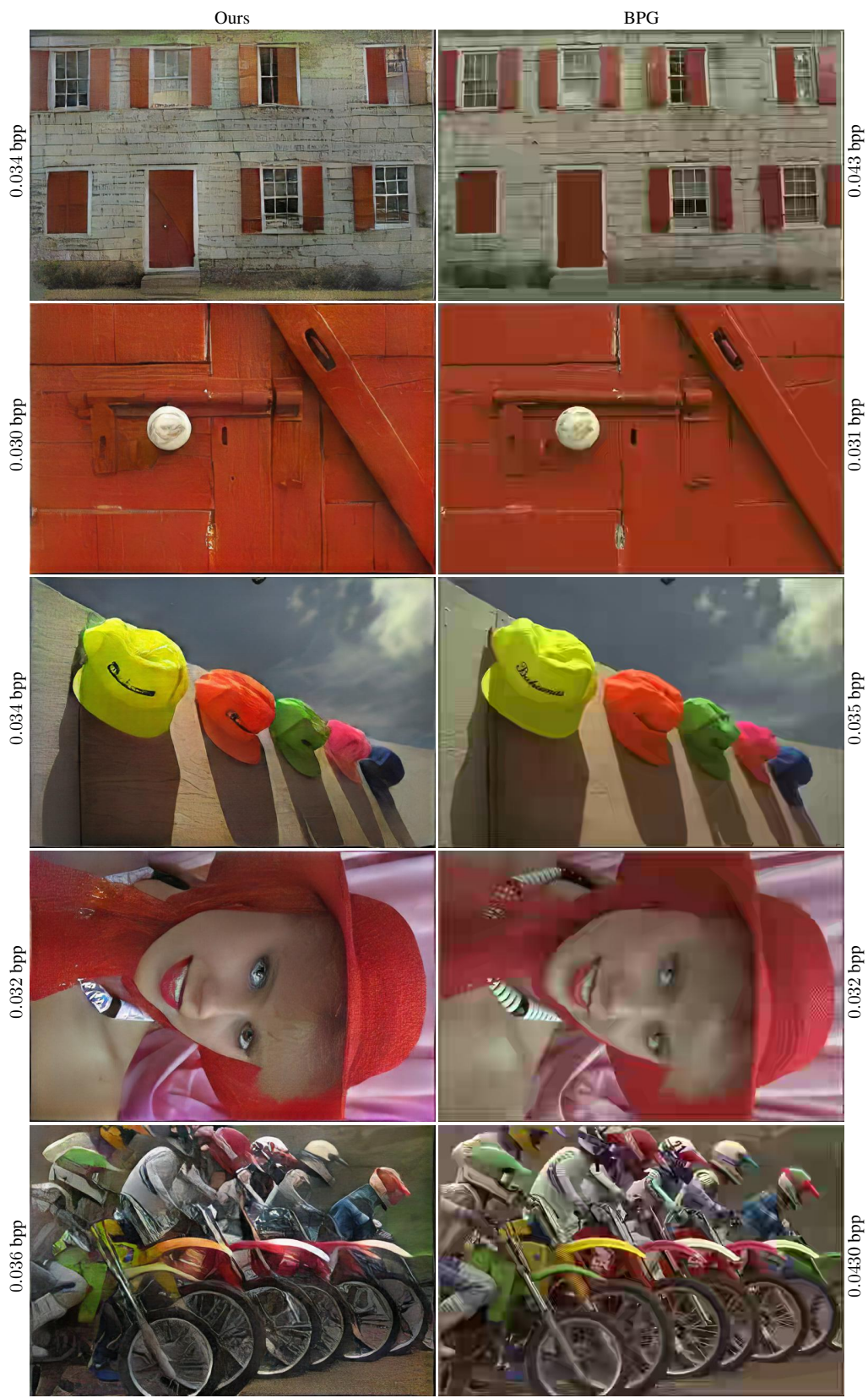

\clearpage
\subsection{Generative Compression on RAISE1k}  \label{sec:gc_raise}
\begin{figure}[h!]
\vspace{-0.3cm}
\centering
\resizebox{.79\linewidth}{!}{
\input{fig_appendix_raise1k.tex}}
\vspace{0.2cm}
\caption{First $5$ images of RAISE1k, produced by our GC model with $C=4$ and BPG.}
\vspace{-2cm}
\end{figure}
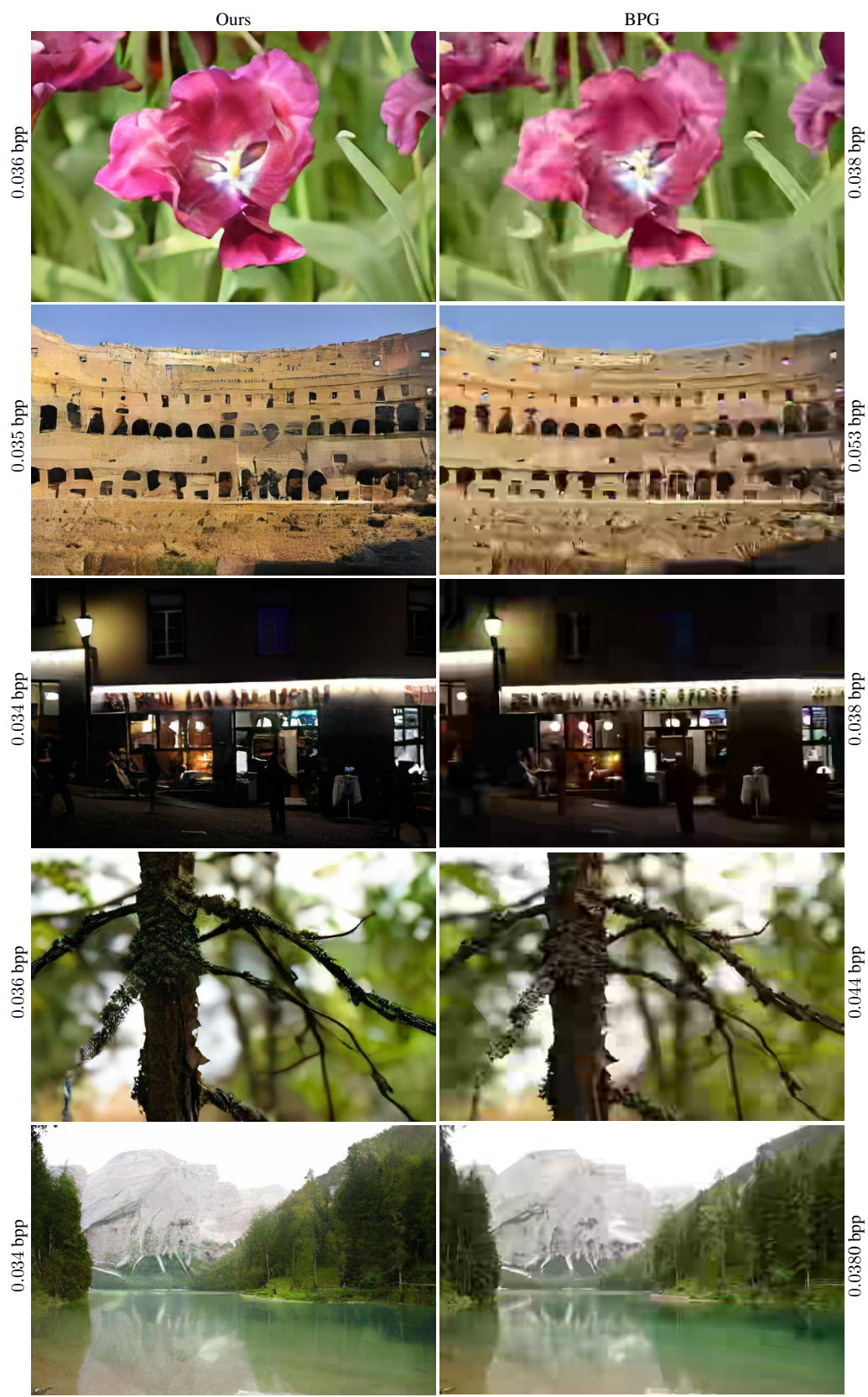

\clearpage
\subsection{Generative Compression on Cityscapes}  \label{sec:gc_cs}
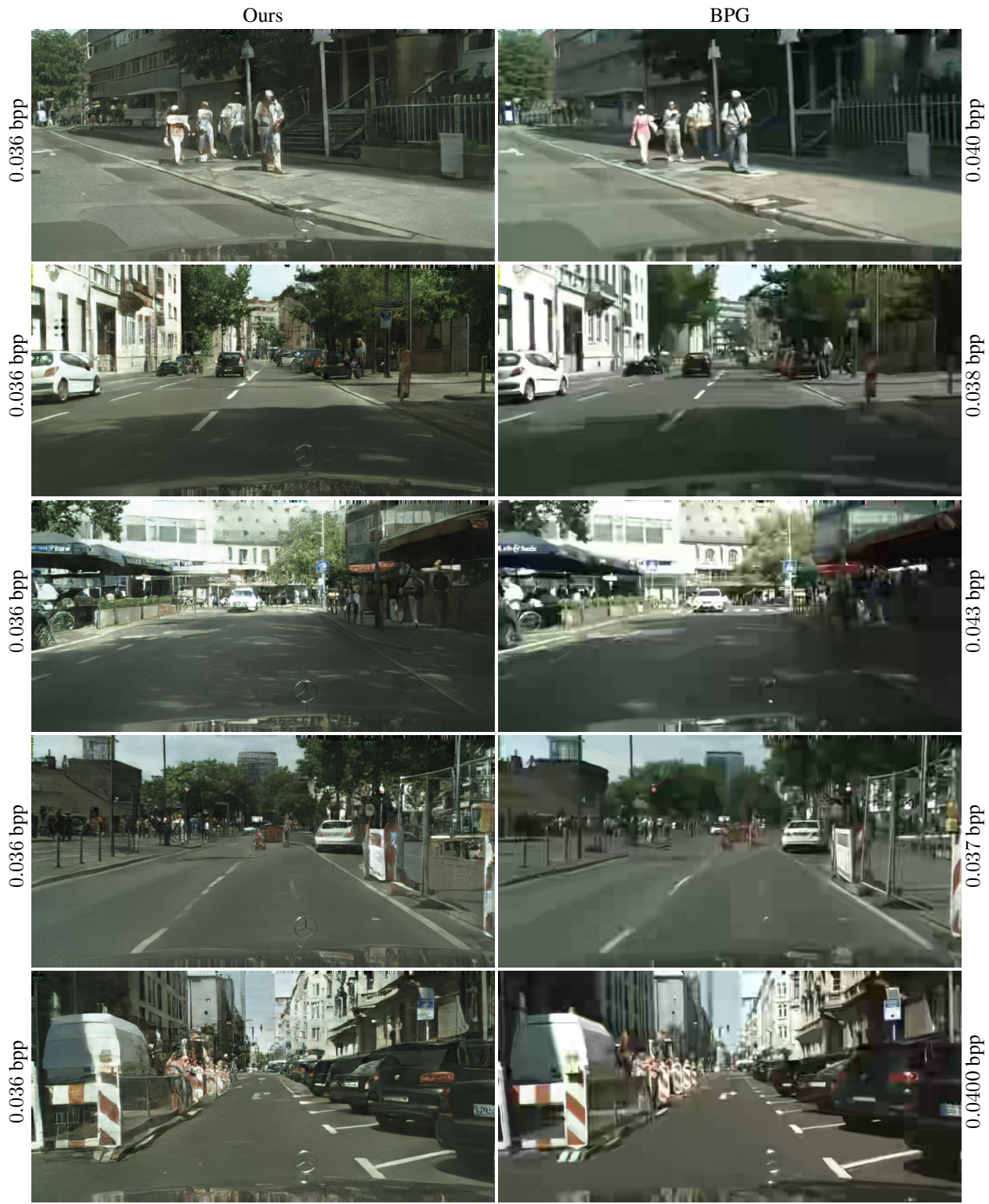
\begin{figure}[h!]
\centering
\input{fig_appendix_cityscapes.tex}
\vspace{0.2cm}
\caption{First $5$ images of Cityscapes, produced by our GC model with $C=4$ and BPG.}
\end{figure}

\clearpage
\subsection{Comparison with~\cite{rippel17a}} \label{sec:comp_rippel}

\clearpage
\subsection{Comparison with \cite{minnen2018joint}}\label{sec:comp_minnen}

\clearpage

\subsection{Sampling the compressed representations}\label{sec:vis_repr}

\begin{figure}[ht!]
    \centering
{\setlength{\tabcolsep}{1pt}
\begin{tabular}{ccc}
                \raisebox{2.5\normalbaselineskip}[0pt][0pt]{\parbox[t]{4mm}{\rotatebox[origin=l]{90}{Cityscapes}}}
        &\includegraphics[width=0.45\linewidth]{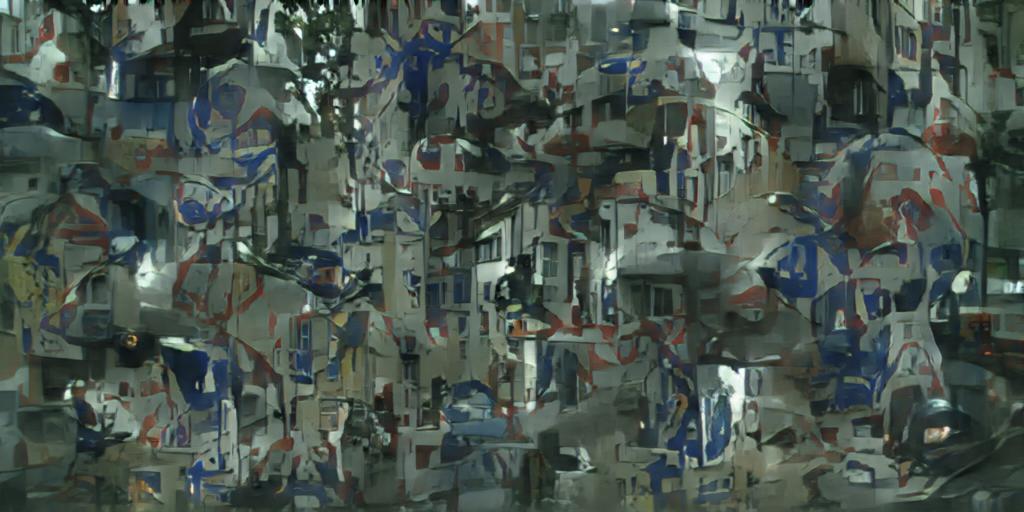} &
        \includegraphics[width=0.45\linewidth]{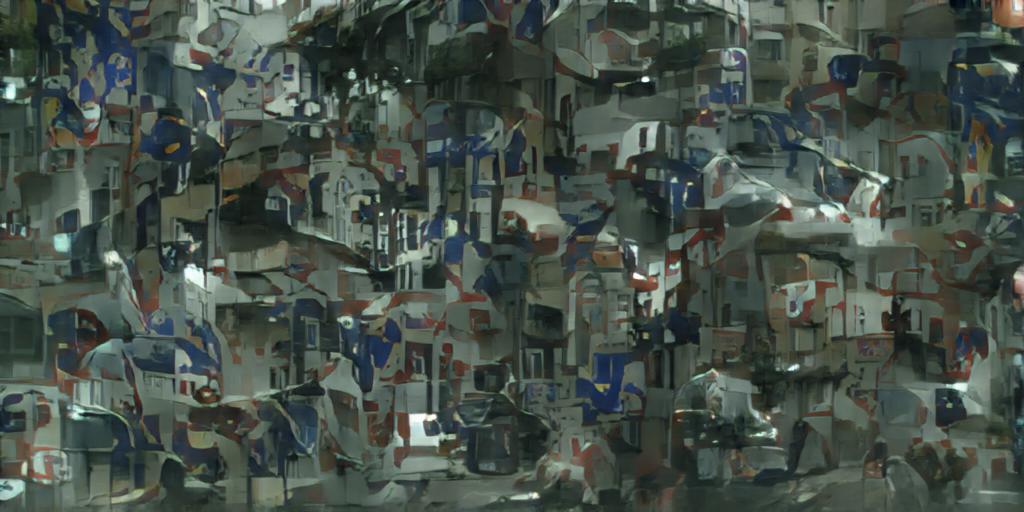} \\
        \raisebox{3.8\normalbaselineskip}[0pt][0pt]{\parbox[t]{4mm}{\rotatebox[origin=l]{90}{Open Images}}}
        &\includegraphics[width=0.45\linewidth]{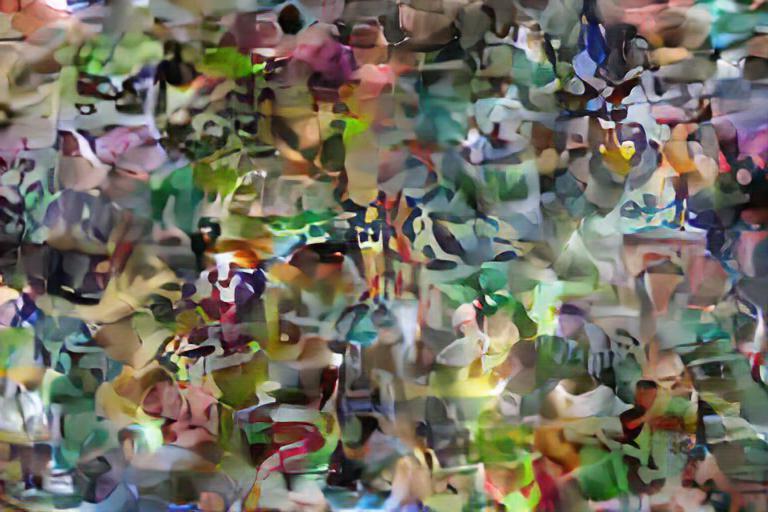} &
        \includegraphics[width=0.45\linewidth]{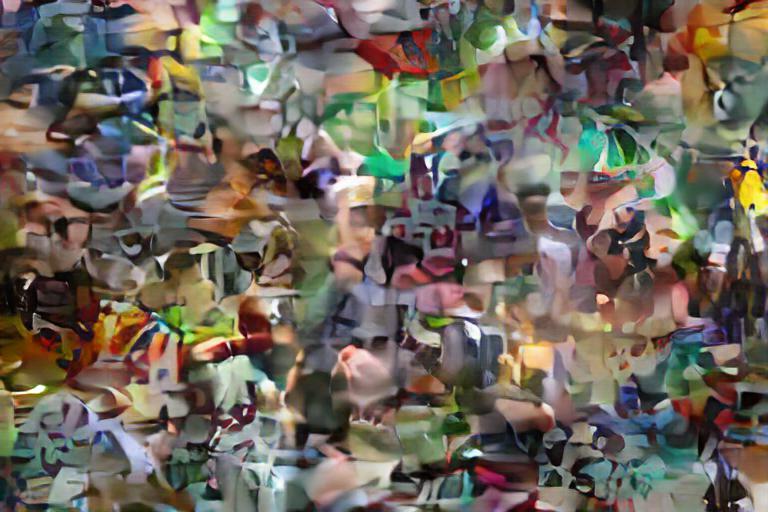} \\
        \end{tabular}}
        \vspace{0.2cm}
        \caption{\label{fig:sample_bn_openimages} We uniformly sample codes from the (discrete) latent space $\hat{\w}$ of our generative compression models (GC with $C=4$) trained on Cityscapes and Open Images. The Cityscapes model outputs domain specific patches (street signs, buildings, trees, road), whereas the Open Images samples are more colorful and consist of more generic visual patches.
}
\end{figure}

\begin{figure}[hb!]
    \centering
        \vspace{-1ex}
{\setlength{\tabcolsep}{1pt}
        \begin{tabular}{ll}
        \includegraphics[width=0.45\linewidth]{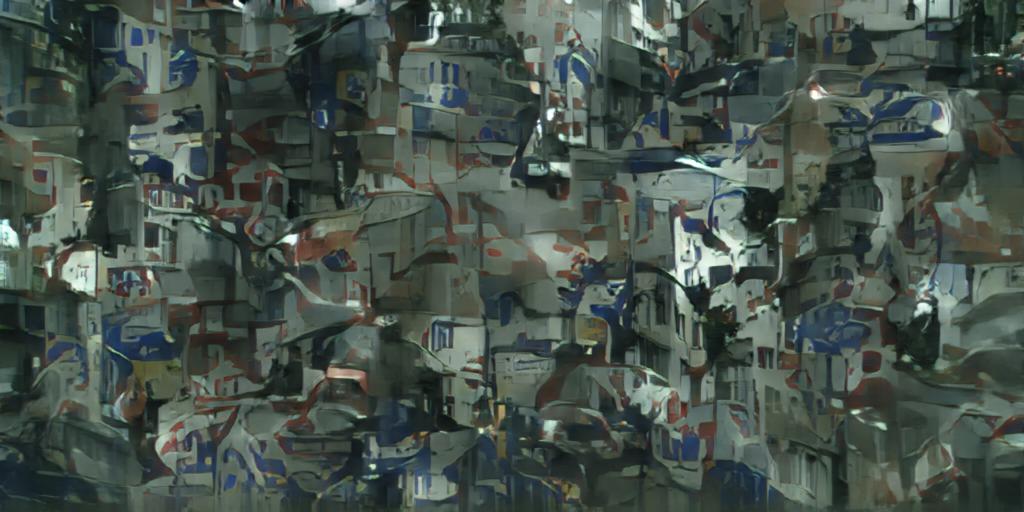}  &
        \includegraphics[width=0.45\linewidth]{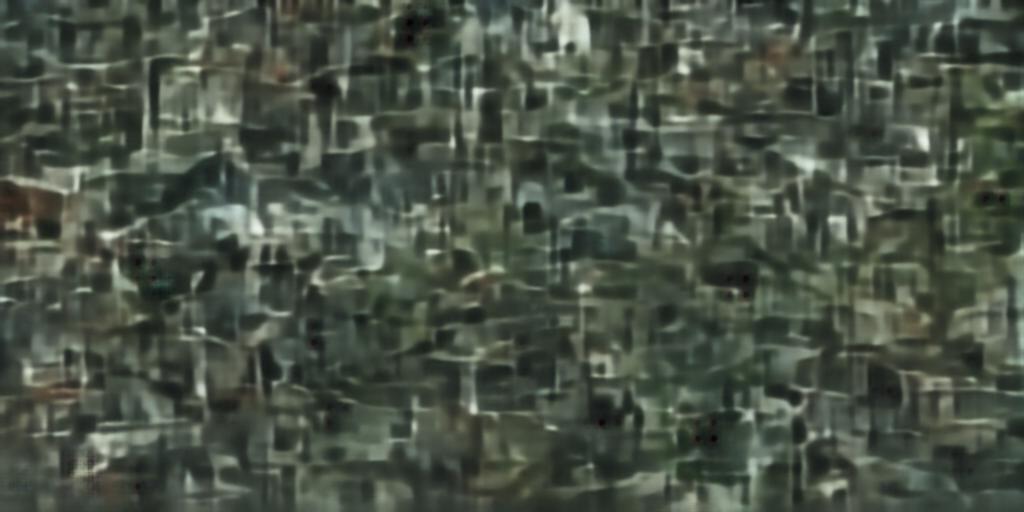} \\
        \includegraphics[width=0.45\linewidth]{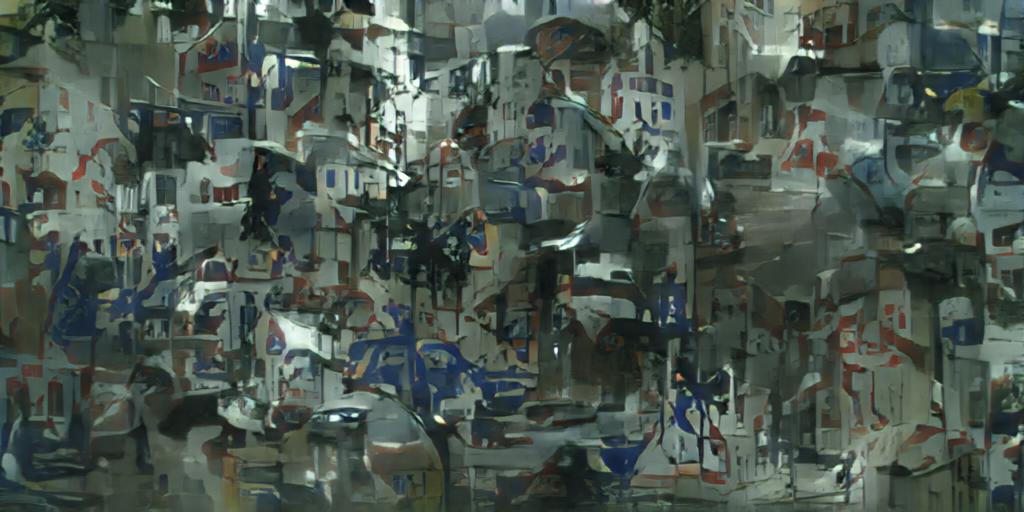}  &
        \includegraphics[width=0.45\linewidth]{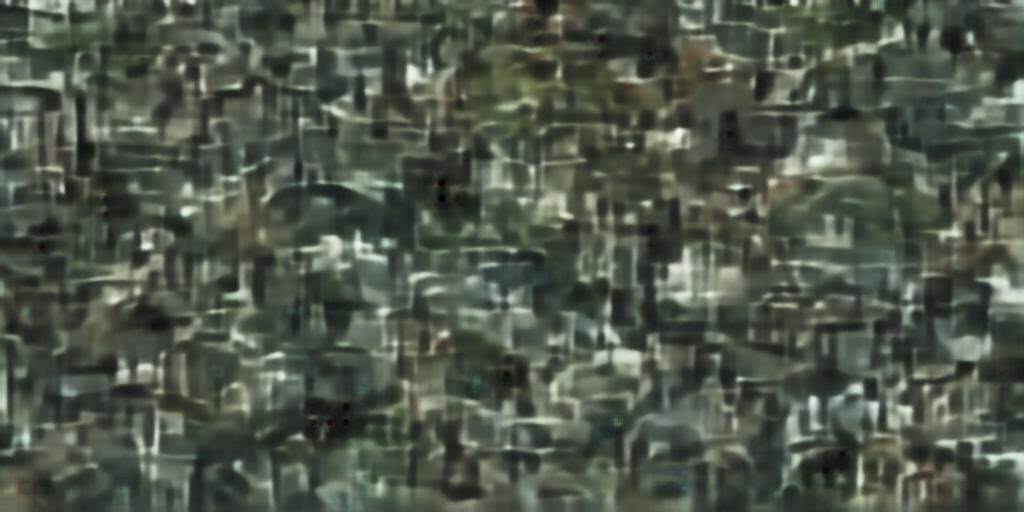} \\
        GC ($C=4$) &  MSE ($C=4$) \\
        \end{tabular}}
        \vspace{0.2cm}
        \caption{\label{fig:sample_bn_uniform} We train the same architecture with $C=4$ for MSE and for generative compression on Cityscapes.
        When uniformly sampling the (discrete) latent space $\hat{\w}$ of the models, we see stark differences between the decoded images $G(\hat{\w})$. The GC model produces patches that resemble parts of Cityscapes images (street signs, buildings, etc.), whereas the MSE model outputs looks like low-frequency noise.
        \vspace{-10ex}
}
\end{figure}

\begin{figure}[h]
    \centering
{\setlength{\tabcolsep}{1pt}
        \begin{tabular}{ll}
        \includegraphics[width=0.495\linewidth]{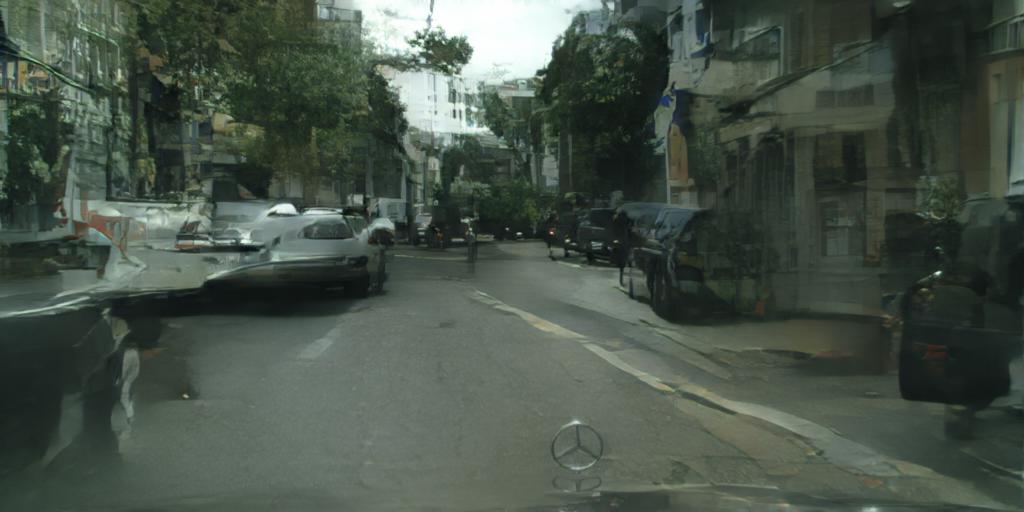} &
        \includegraphics[width=0.495\linewidth]{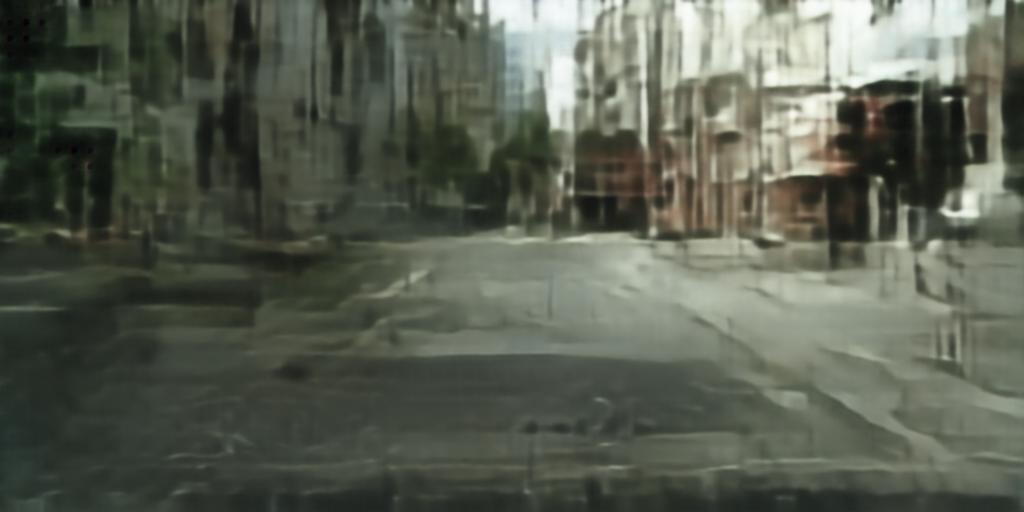} \\
        \includegraphics[width=0.495\linewidth]{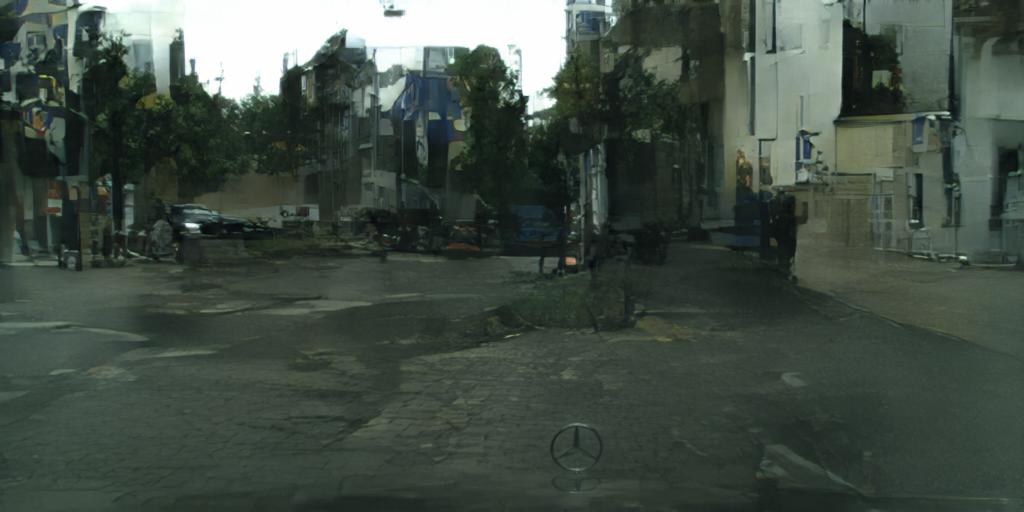} &
        \includegraphics[width=0.495\linewidth]{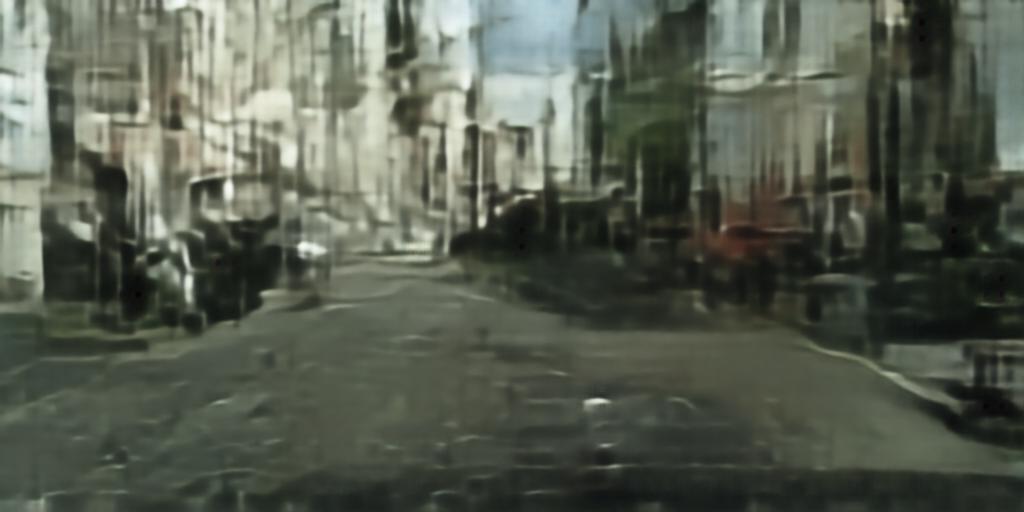} \\
        \includegraphics[width=0.495\linewidth]{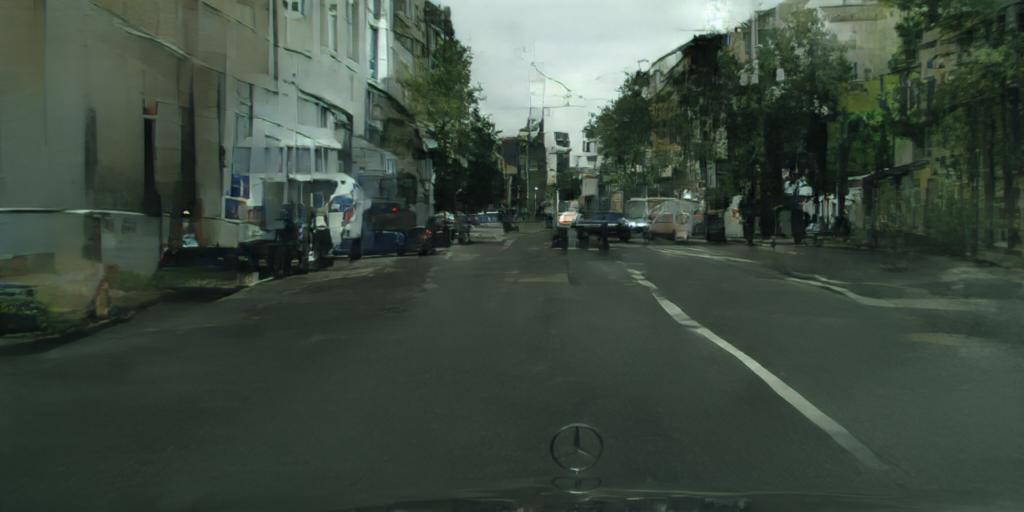} &
        \includegraphics[width=0.495\linewidth]{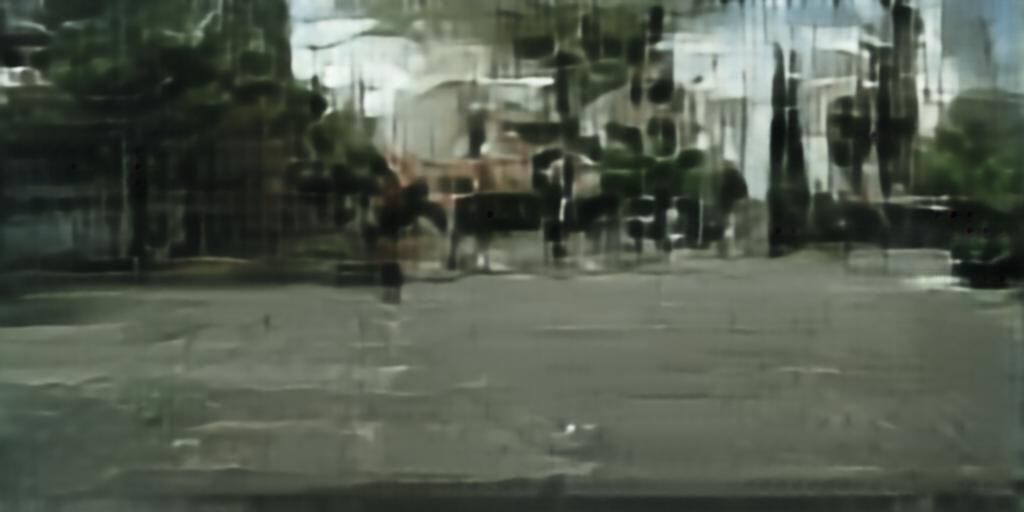} \\
        \includegraphics[width=0.495\linewidth]{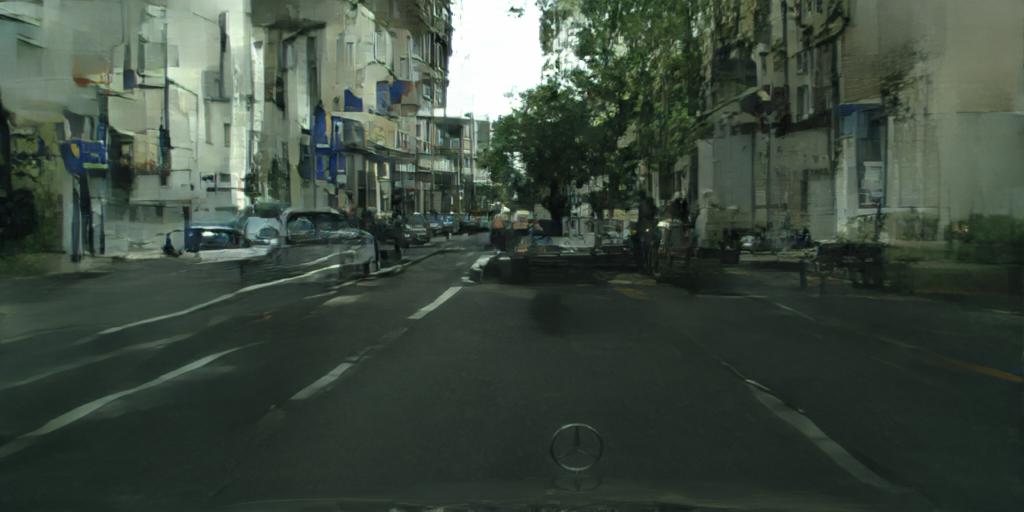} &
        \includegraphics[width=0.495\linewidth]{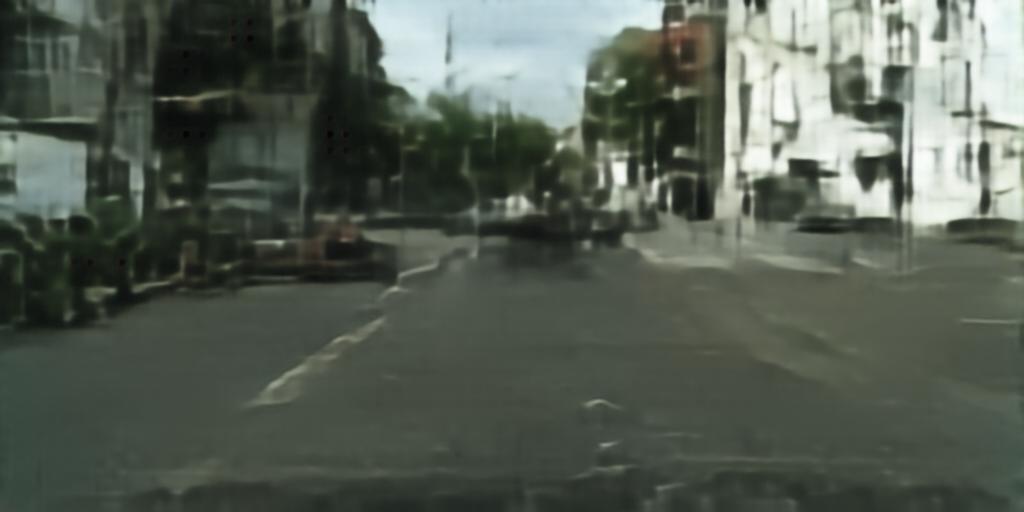} \\
        GC model with $C=4$ & MSE baseline model with $C=4$\\
        \end{tabular}}
        \caption{\label{fig:sample_bn_gan} We experiment with learning the distribution of $\hat \w = E(\x)$ by training an improved Wasserstein GAN \cite{gulrajani2017improved}. 
When sampling form the decoder/generator $G$ of our model by feeding it with samples from the improved WGAN generator, we obtain much sharper images than when we do the same with an MSE model.}
\end{figure}

\clearpage
\subsection{Selective Compression on Cityscapes} \label{sec:sccity}
\renewcommand{\arraystretch}{1.9}

\begin{figure}[h!]
\centering
    {\setlength{\tabcolsep}{1pt}
    \input{fig_full_classes_2.tex}}
    \vspace{0.2cm}
\caption{Synthesizing different classes for two different images from Cityscapes, using our SC network with $C=4$.
In each image except for \emph{no synthesis}, we additionally synthesize the classes \emph{vegetation, sky, sidewalk, ego vehicle, wall}. }
\end{figure}
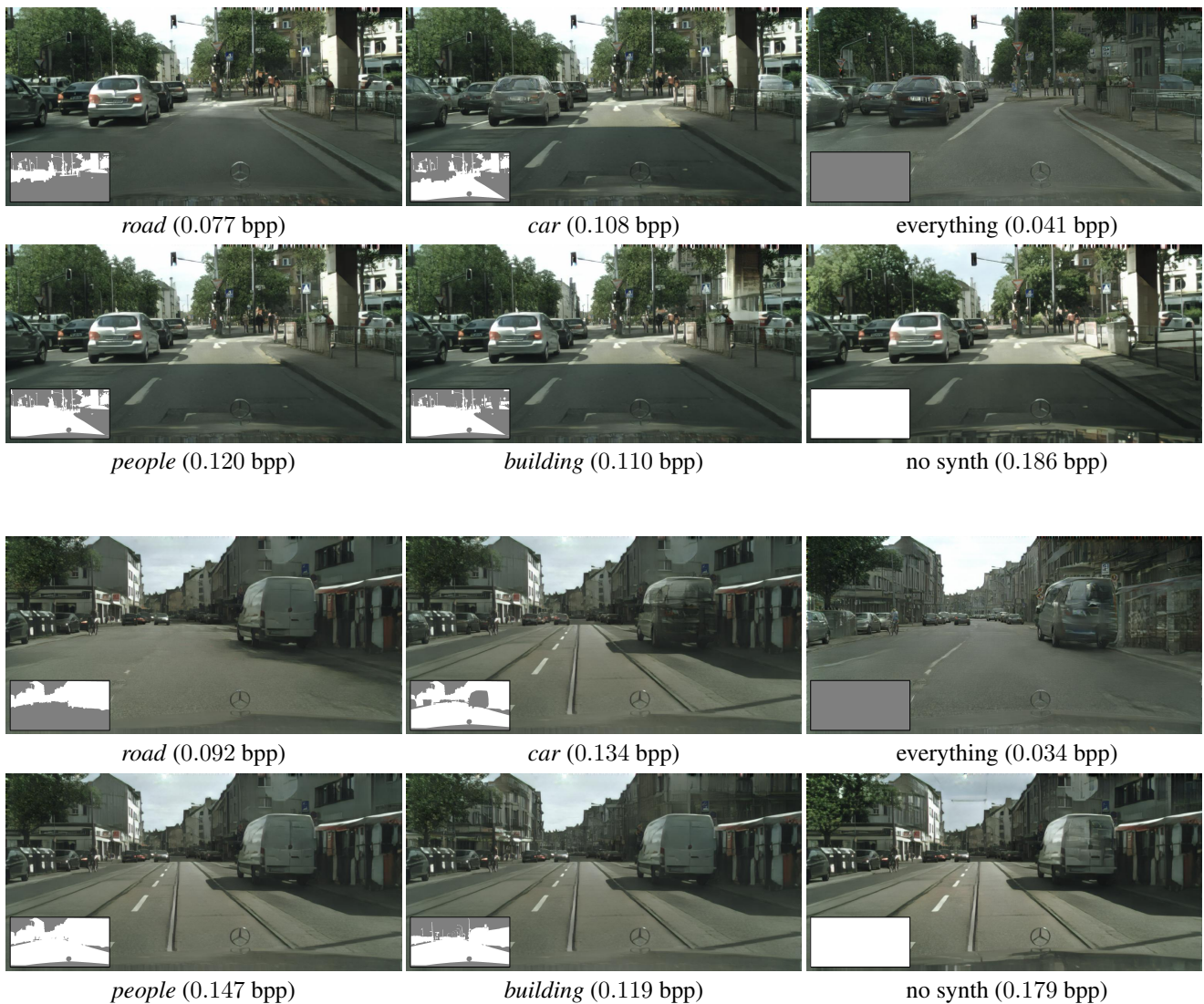

\clearpage

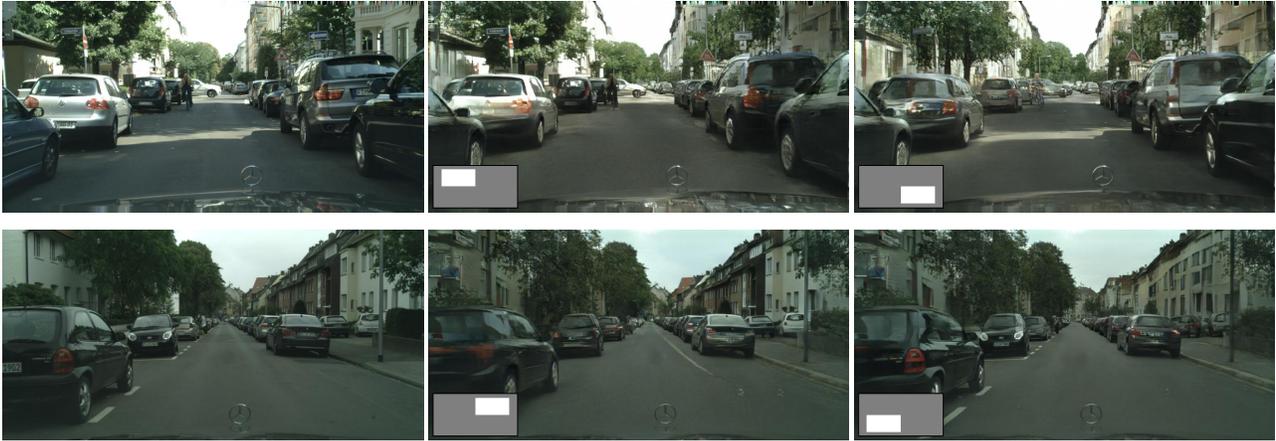
\begin{figure}[h!]
\centering
    {\setlength{\tabcolsep}{1pt}
    \input{fig_full_boxes_2.tex}}
\caption{Example images obtained by our SC network ($C=8$) preserving a box and synthesizing the rest of the image, on Cityscapes. The SC network seamlessly merges preserved and generated image content even in places where the box crosses object boundaries.}
\end{figure}

\begin{figure}[h!]
    \vspace{3ex}
\centering
    {\setlength{\tabcolsep}{1pt}
    \input{fig_predicted.tex}}
    \caption{Reconstructions obtained by our SC network using semantic label maps estimated from the input image via PSPNet \cite{zhao2017pspnet}.}
\end{figure}
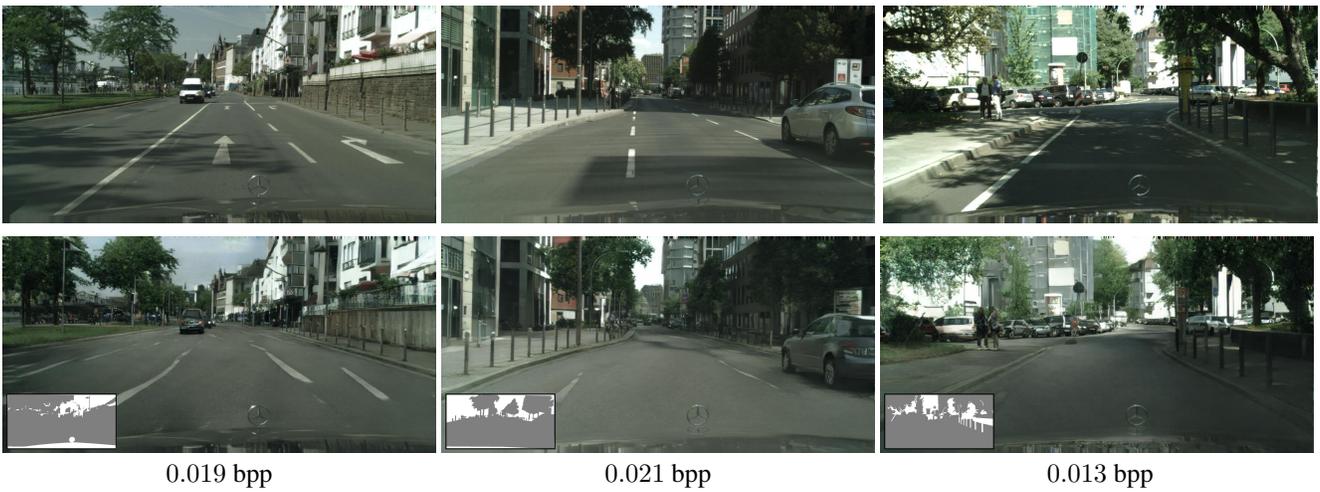

\end{document}

%% file: mymacros.tex
\newcommand{\safemath}[2]{\newcommand{#1}{\ensuremath{#2}\xspace}}

\newcommand{\m}{\bm{m}}

\newcommand{\s}{\bm{s}}

\newcommand{\x}{\bm{x}}

\newcommand{\z}{\bm{z}}
\newcommand{\w}{\bm{w}}
\renewcommand{\v}{\bm{v}}

\newcommand{\0}{\mathbf{0}}

\newcommand{\R}{\mathbb{R}}

\newcommand{\E}{\mathbb{E}}

\safemath{\Ab}{\mathbb{A}}
\safemath{\Bb}{\mathbb{B}}
\safemath{\Cb}{\mathbb{C}}
\safemath{\Db}{\mathbb{D}}
\safemath{\Eb}{\mathbb{E}}
\safemath{\Fb}{\mathbb{F}}
\safemath{\Gb}{\mathbb{G}}
\safemath{\Hb}{\mathbb{H}}
\safemath{\Ib}{\mathbb{I}}
\safemath{\Jb}{\mathbb{J}}
\safemath{\Kb}{\mathbb{K}}
\safemath{\Lb}{\mathbb{L}}
\safemath{\Mb}{\mathbb{M}}
\safemath{\Nb}{\mathbb{N}}
\safemath{\Ob}{\mathbb{O}}
\safemath{\Pb}{\mathbb{P}}
\safemath{\Qb}{\mathbb{Q}}
\safemath{\Rb}{\mathbb{R}}
\safemath{\Sb}{\mathbb{S}}
\safemath{\Tb}{\mathbb{T}}
\safemath{\Ub}{\mathbb{U}}
\safemath{\Vb}{\mathbb{V}}
\safemath{\Wb}{\mathbb{W}}
\safemath{\Xb}{\mathbb{X}}
\safemath{\Yb}{\mathbb{Y}}
\safemath{\Zb}{\mathbb{Z}}

\safemath{\Af}{\mathbf{A}}
\safemath{\Bf}{\mathbf{B}}
\safemath{\Df}{\mathbf{D}}
\safemath{\Ef}{\mathbf{E}}
\safemath{\Ff}{\mathbf{F}}
\safemath{\Gf}{\mathbf{G}}
\safemath{\Hf}{\mathbf{H}}
\safemath{\Jf}{\mathbf{J}}
\safemath{\Kf}{\mathbf{K}}
\safemath{\Lf}{\mathbf{L}}
\safemath{\Mf}{\mathbf{M}}
\safemath{\Nf}{\mathbf{N}}
\safemath{\Of}{\mathbf{O}}
\safemath{\Pf}{\mathbf{P}}
\safemath{\Qf}{\mathbf{Q}}
\safemath{\Rf}{\mathbf{R}}
\safemath{\Sf}{\mathbf{S}}
\safemath{\Tf}{\mathbf{T}}
\safemath{\Uf}{\mathbf{U}}
\safemath{\Vf}{\mathbf{V}}
\safemath{\Wf}{\mathbf{W}}
\safemath{\Xf}{\mathbf{X}}
\safemath{\Yf}{\mathbf{Y}}
\safemath{\Zf}{\mathbf{Z}}

\safemath{\af}{\mathbf{a}}
\safemath{\df}{\mathbf{d}}
\safemath{\ef}{\mathbf{e}}
\safemath{\ff}{\mathbf{f}}
\safemath{\gf}{\mathbf{g}}
\safemath{\hf}{\mathbf{h}}
\safemath{\jf}{\mathbf{j}}
\safemath{\kf}{\mathbf{k}}
\safemath{\lf}{\mathbf{l}}
\safemath{\mf}{\mathbf{m}}
\safemath{\nf}{\mathbf{n}}
\safemath{\of}{\mathbf{o}}
\safemath{\pf}{\mathbf{p}}
\safemath{\qf}{\mathbf{q}}
\safemath{\rf}{\mathbf{r}}
\safemath{\tf}{\mathbf{t}}
\safemath{\uf}{\mathbf{u}}
\safemath{\vf}{\mathbf{v}}
\safemath{\wf}{\mathbf{w}}
\safemath{\xf}{\mathbf{x}}
\safemath{\yf}{\mathbf{y}}
\safemath{\zf}{\mathbf{z}}

\safemath{\Ac}{\mathcal{A}}
\safemath{\Bc}{\mathcal{B}}
\safemath{\Cc}{\mathcal{C}}
\safemath{\Dc}{\mathcal{D}}
\safemath{\Ec}{\mathcal{E}}
\safemath{\Fc}{\mathcal{F}}
\safemath{\Gc}{\mathcal{G}}
\safemath{\Hc}{\mathcal{H}}
\safemath{\Ic}{\mathcal{I}}
\safemath{\Jc}{\mathcal{J}}
\safemath{\Kc}{\mathcal{K}}
\safemath{\Lc}{\mathcal{L}}
\safemath{\Mc}{\mathcal{M}}
\safemath{\Nc}{\mathcal{N}}
\safemath{\Oc}{\mathcal{O}}
\safemath{\Pc}{\mathcal{P}}
\safemath{\Qc}{\mathcal{Q}}
\safemath{\Rc}{\mathcal{R}}
\safemath{\Sc}{\mathcal{S}}
\safemath{\Tc}{\mathcal{T}}
\safemath{\Uc}{\mathcal{U}}
\safemath{\Vc}{\mathcal{V}}
\safemath{\Wc}{\mathcal{W}}
\safemath{\Xc}{\mathcal{X}}
\safemath{\Yc}{\mathcal{Y}}
\safemath{\Zc}{\mathcal{Z}}

\safemath{\As}{\mathsf{A}}
\safemath{\Bs}{\mathsf{B}}
\safemath{\Cs}{\mathsf{C}}
\safemath{\Ds}{\mathsf{D}}
\safemath{\Es}{\mathsf{E}}
\safemath{\Fs}{\mathsf{F}}
\safemath{\Gs}{\mathsf{G}}
\safemath{\Hs}{\mathsf{H}}
\safemath{\Is}{\mathsf{I}}
\safemath{\Js}{\mathsf{J}}
\safemath{\Ks}{\mathsf{K}}
\safemath{\Ls}{\mathsf{L}}
\safemath{\Ms}{\mathsf{M}}
\safemath{\Ns}{\mathsf{N}}
\safemath{\Os}{\mathsf{O}}
\safemath{\Ps}{\mathsf{P}}
\safemath{\Qs}{\mathsf{Q}}
\safemath{\Rs}{\mathsf{R}}
\safemath{\Ss}{\mathsf{S}}
\safemath{\Ts}{\mathsf{T}}
\safemath{\Us}{\mathsf{U}}
\safemath{\Vs}{\mathsf{V}}
\safemath{\Ws}{\mathsf{W}}
\safemath{\Xs}{\mathsf{X}}
\safemath{\Ys}{\mathsf{Y}}
\safemath{\Zs}{\mathsf{Z}}

\safemath{\as}{\mathsf{a}}
\safemath{\bs}{\mathsf{b}}
\safemath{\cs}{\mathsf{c}}
\safemath{\ds}{\mathsf{d}}
\safemath{\es}{\mathsf{e}}
\safemath{\fs}{\mathsf{f}}
\safemath{\gs}{\mathsf{g}}
\safemath{\hs}{\mathsf{h}}
\safemath{\is}{\mathsf{i}}
\safemath{\js}{\mathsf{j}}
\safemath{\ks}{\mathsf{k}}
\safemath{\ls}{\mathsf{l}}
\safemath{\ms}{\mathsf{m}}
\safemath{\ns}{\mathsf{n}}
\safemath{\os}{\mathsf{o}}
\safemath{\ps}{\mathsf{p}}
\safemath{\qs}{\mathsf{q}}
\safemath{\rs}{\mathsf{r}}
\safemath{\us}{\mathsf{u}}
\safemath{\xs}{\mathsf{x}}
\safemath{\ys}{\mathsf{y}}
\safemath{\zs}{\mathsf{z}}

\DeclareMathAlphabet{\mathbfsf}{\encodingdefault}{\sfdefault}{bx}{n}
\safemath{\Asf}{\mathbfsf{A}}
\safemath{\Bsf}{\mathbfsf{B}}
\safemath{\Csf}{\mathbfsf{C}}
\safemath{\Dsf}{\mathbfsf{D}}
\safemath{\Esf}{\mathbfsf{E}}
\safemath{\Fsf}{\mathbfsf{F}}
\safemath{\Gsf}{\mathbfsf{G}}
\safemath{\Hsf}{\mathbfsf{H}}
\safemath{\Isf}{\mathbfsf{I}}
\safemath{\Jsf}{\mathbfsf{J}}
\safemath{\Ksf}{\mathbfsf{K}}
\safemath{\Lsf}{\mathbfsf{L}}
\safemath{\Msf}{\mathbfsf{M}}
\safemath{\Nsf}{\mathbfsf{N}}
\safemath{\Osf}{\mathbfsf{O}}
\safemath{\Psf}{\mathbfsf{P}}
\safemath{\Qsf}{\mathbfsf{Q}}
\safemath{\Rsf}{\mathbfsf{R}}
\safemath{\Ssf}{\mathbfsf{S}}
\safemath{\Tsf}{\mathbfsf{T}}
\safemath{\Usf}{\mathbfsf{U}}
\safemath{\Vsf}{\mathbfsf{V}}
\safemath{\Wsf}{\mathbfsf{W}}
\safemath{\Xsf}{\mathbfsf{X}}
\safemath{\Ysf}{\mathbfsf{Y}}
\safemath{\Zsf}{\mathbfsf{Z}}

\safemath{\asf}{\mathbfsf{a}}
\safemath{\bsf}{\mathbfsf{b}}
\safemath{\csf}{\mathbfsf{c}}
\safemath{\dsf}{\mathbfsf{d}}
\safemath{\esf}{\mathbfsf{e}}
\safemath{\fsf}{\mathbfsf{f}}
\safemath{\gsf}{\mathbfsf{g}}
\safemath{\hsf}{\mathbfsf{h}}
\safemath{\isf}{\mathbfsf{i}}
\safemath{\jsf}{\mathbfsf{j}}
\safemath{\ksf}{\mathbfsf{k}}
\safemath{\lsf}{\mathbfsf{l}}
\safemath{\msf}{\mathbfsf{m}}
\safemath{\nsf}{\mathbfsf{n}}
\safemath{\osf}{\mathbfsf{o}}
\safemath{\psf}{\mathbfsf{p}}
\safemath{\qsf}{\mathbfsf{q}}
\safemath{\rsf}{\mathbfsf{r}}
\safemath{\ssf}{\mathbfsf{s}}
\safemath{\tsf}{\mathbfsf{t}}
\safemath{\usf}{\mathbfsf{u}}
\safemath{\vsf}{\mathbfsf{v}}
\safemath{\wsf}{\mathbfsf{w}}
\safemath{\xsf}{\mathbfsf{x}}
\safemath{\ysf}{\mathbfsf{y}}
\safemath{\zsf}{\mathbfsf{z}}

%% file: fig_full_classes.tex
\begin{tabular}{ccc}
\begin{overpic}[width=0.325\linewidth]{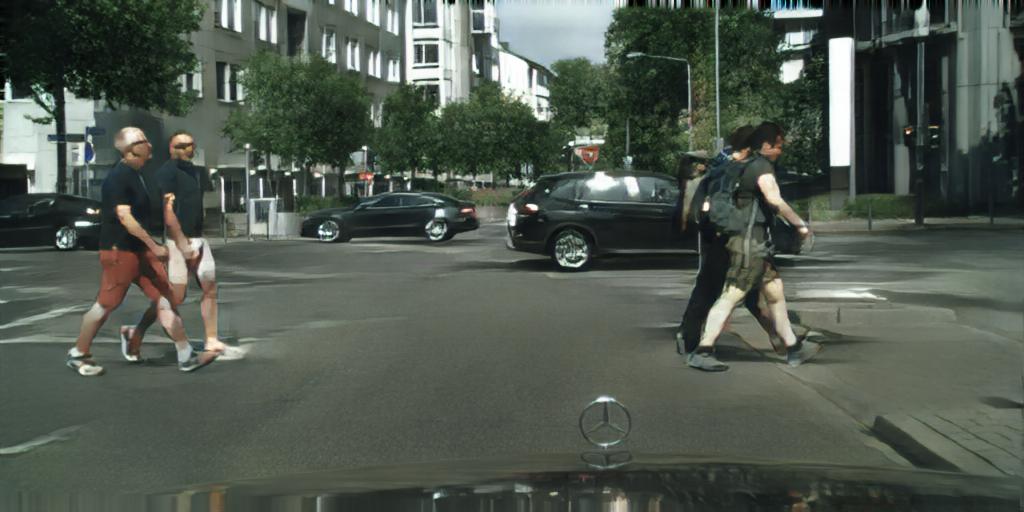}
\put(1,1){\setlength{\fboxsep}{0pt}\setlength{\fboxrule}{0.2pt}\fbox{\includegraphics[width=0.081250\linewidth]{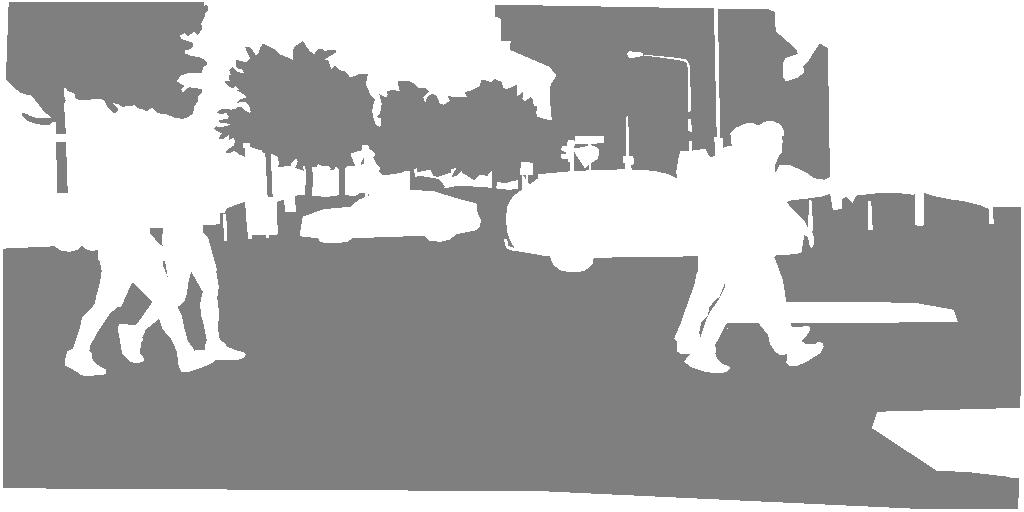}}}
\end{overpic} &
\begin{overpic}[width=0.325\linewidth]{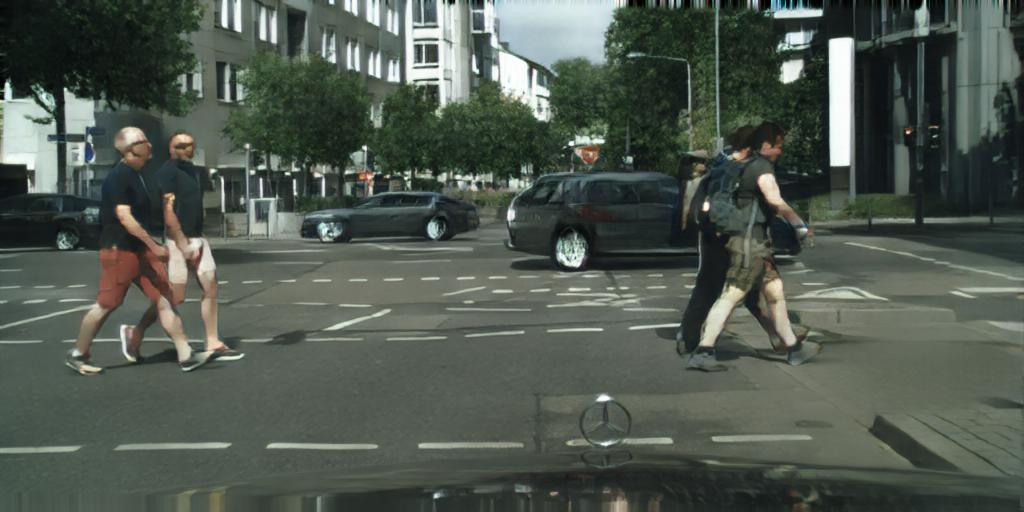}
\put(1,1){\setlength{\fboxsep}{0pt}\setlength{\fboxrule}{0.2pt}\fbox{\includegraphics[width=0.081250\linewidth]{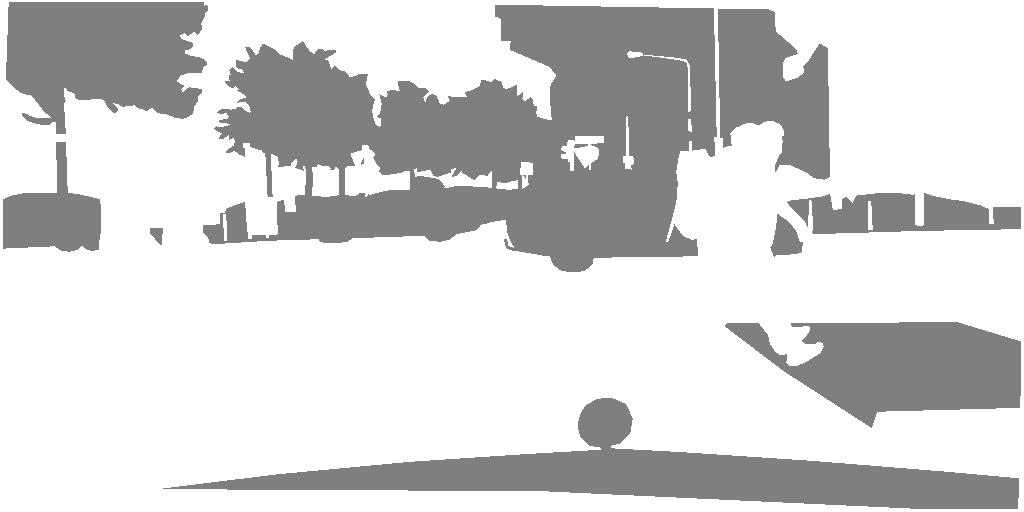}}}
\end{overpic} &
\begin{overpic}[width=0.325\linewidth]{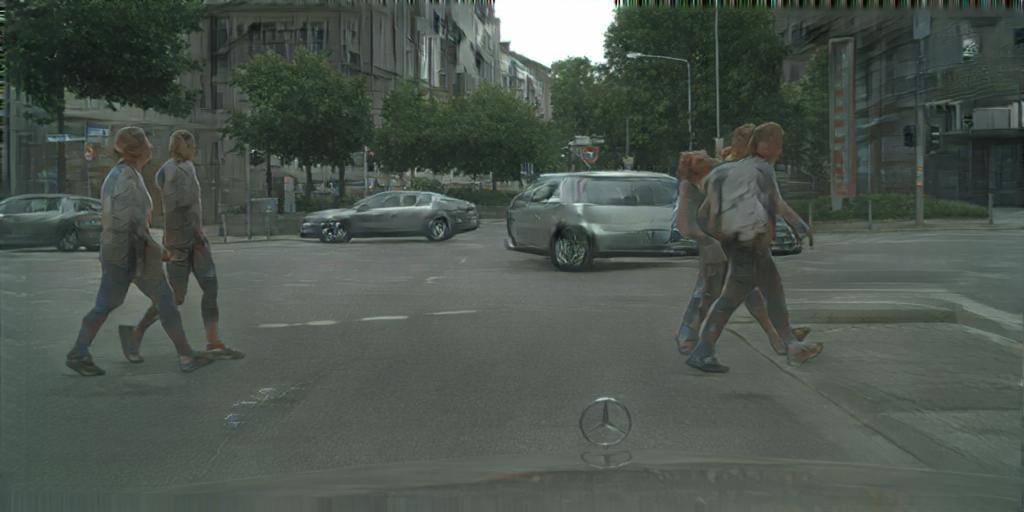}
\put(1,1){\setlength{\fboxsep}{0pt}\setlength{\fboxrule}{0.2pt}\fbox{\includegraphics[width=0.081250\linewidth]{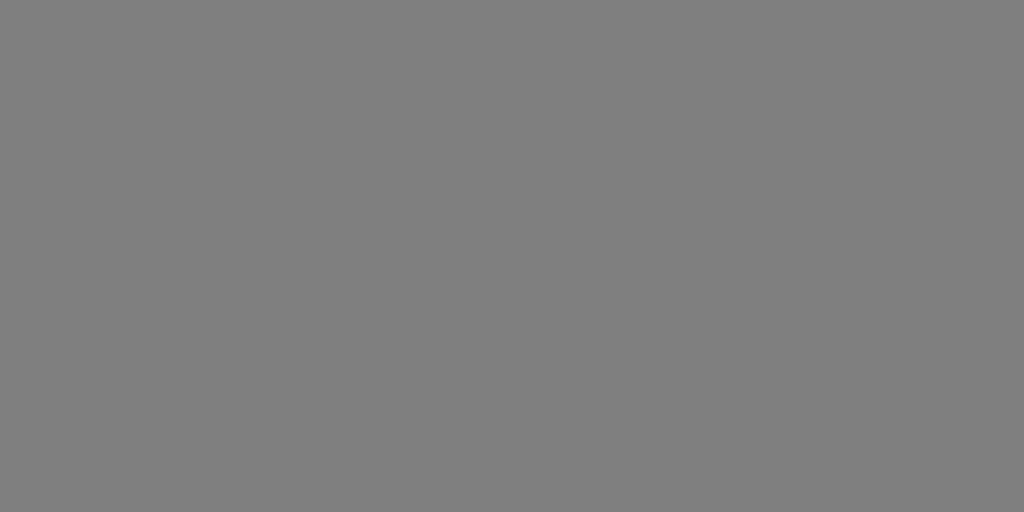}}}
\end{overpic} \\
    \emph{road} ($0.146$bpp, -55\%) & \emph{car} ($0.227$bpp, -15\%) & all synth. ($0.035$bpp, -89\%)   \\[3pt]
\begin{overpic}[width=0.325\linewidth]{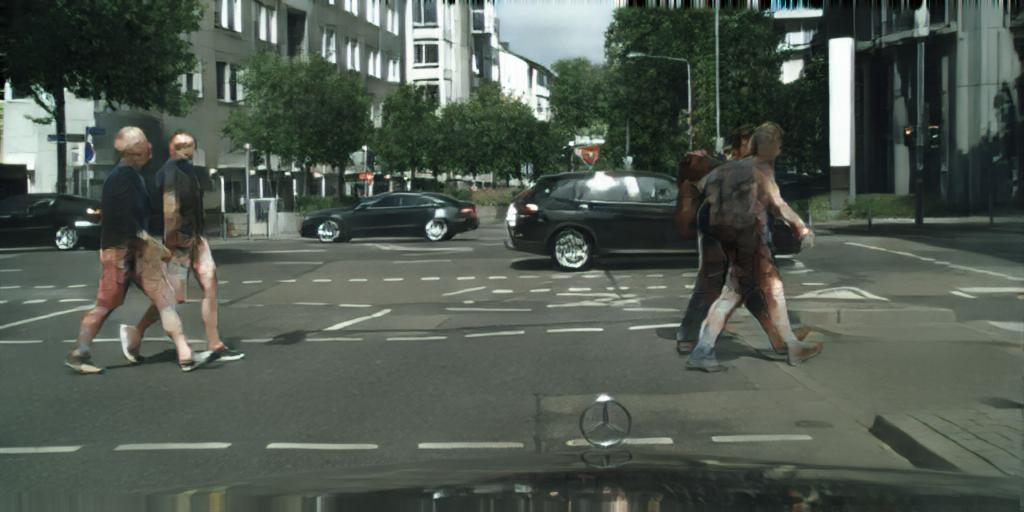}
\put(1,1){\setlength{\fboxsep}{0pt}\setlength{\fboxrule}{0.2pt}\fbox{\includegraphics[width=0.081250\linewidth]{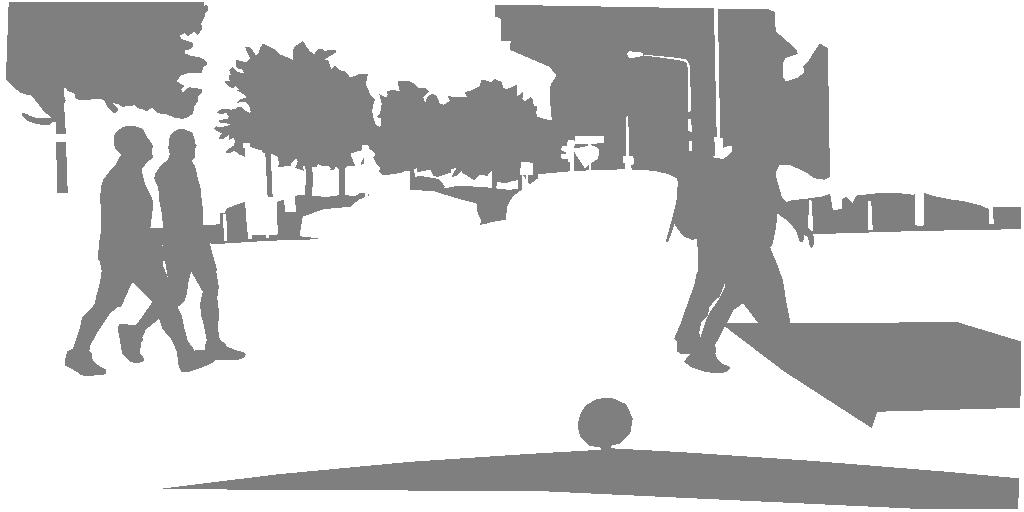}}}
\end{overpic} &
\begin{overpic}[width=0.325\linewidth]{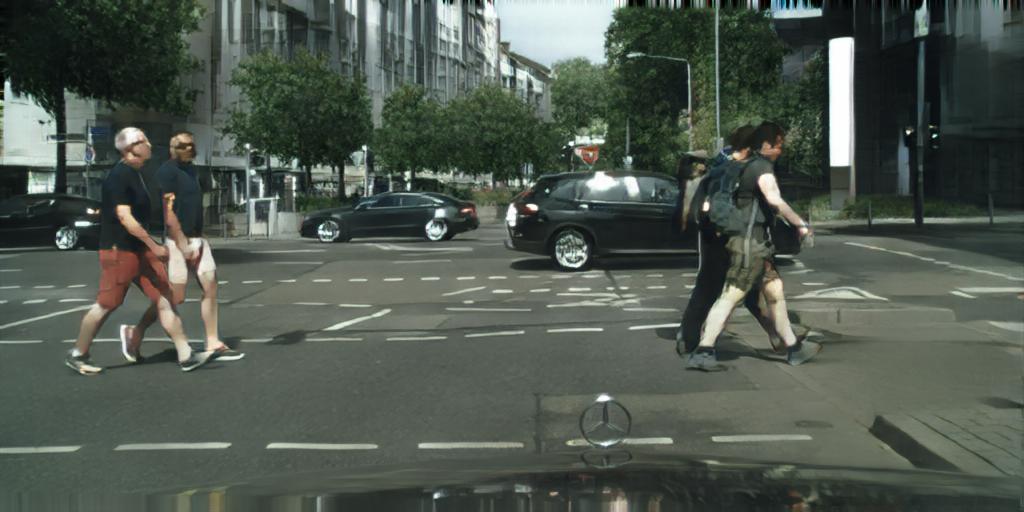}
\put(1,1){\setlength{\fboxsep}{0pt}\setlength{\fboxrule}{0.2pt}\fbox{\includegraphics[width=0.081250\linewidth]{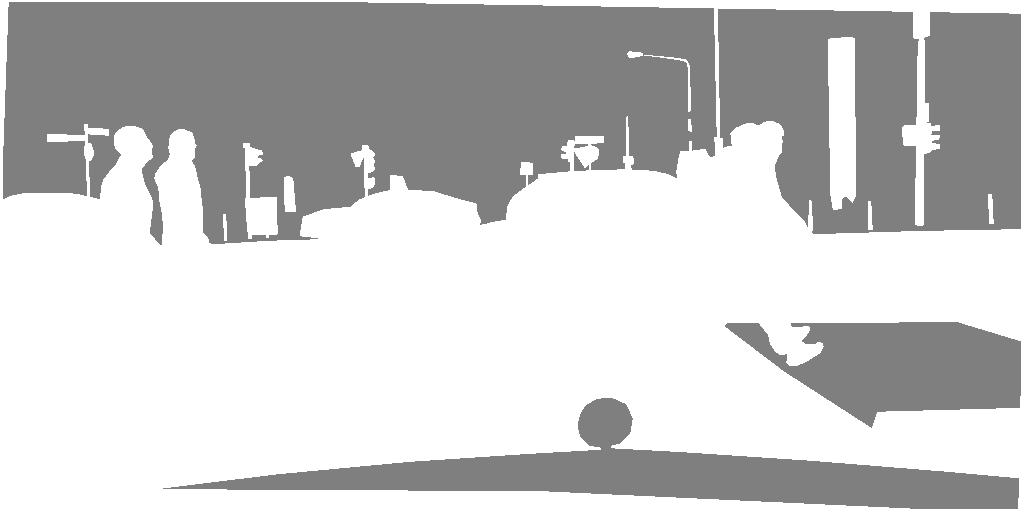}}}
\end{overpic} &
\begin{overpic}[width=0.325\linewidth]{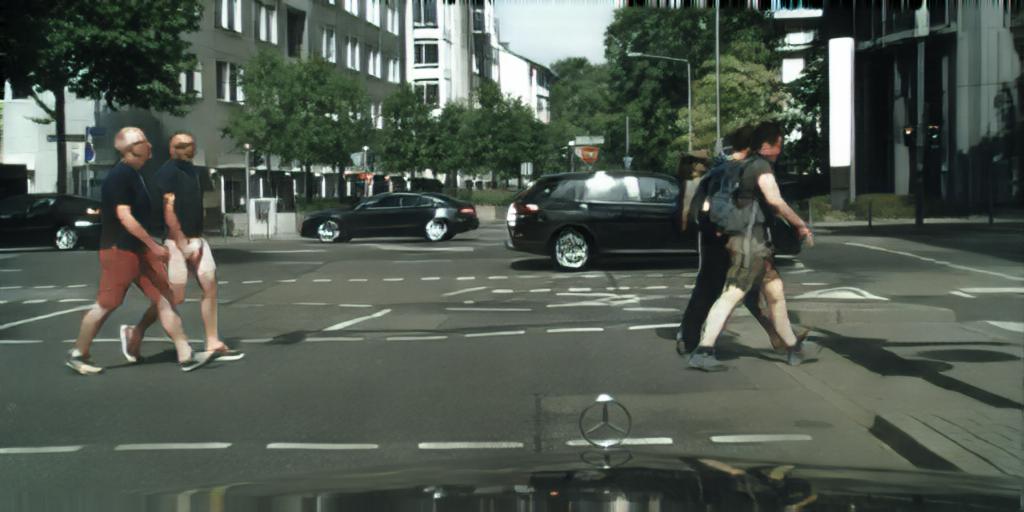}
\put(1,1){\setlength{\fboxsep}{0pt}\setlength{\fboxrule}{0.2pt}\fbox{\includegraphics[width=0.081250\linewidth]{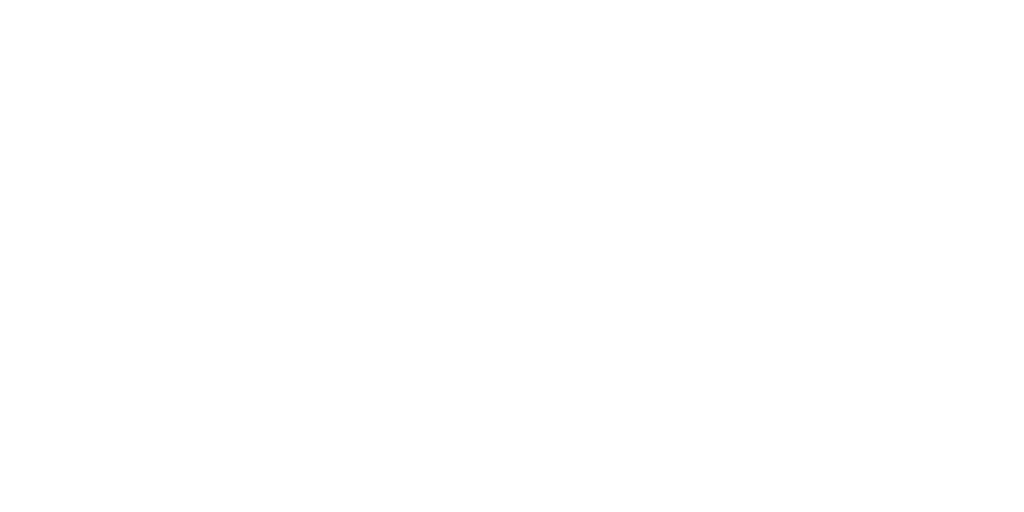}}}
\end{overpic}  \\
\emph{people} ($0.219$bpp, -33\%) & \emph{building} ($0.199$bpp, -39\%) & no synth.\ ($0.326$bpp, -0\%)
\end{tabular}

%% file: fig_appendix_kodak.tex
{\setlength{\tabcolsep}{1pt}
\begin{tabular}{rccl}
&Ours&BPG\\
\rotatebox[origin=c]{90}{$0.034$ bpp}&
\raisebox{-0.5\height}{\includegraphics[width=0.45\linewidth]{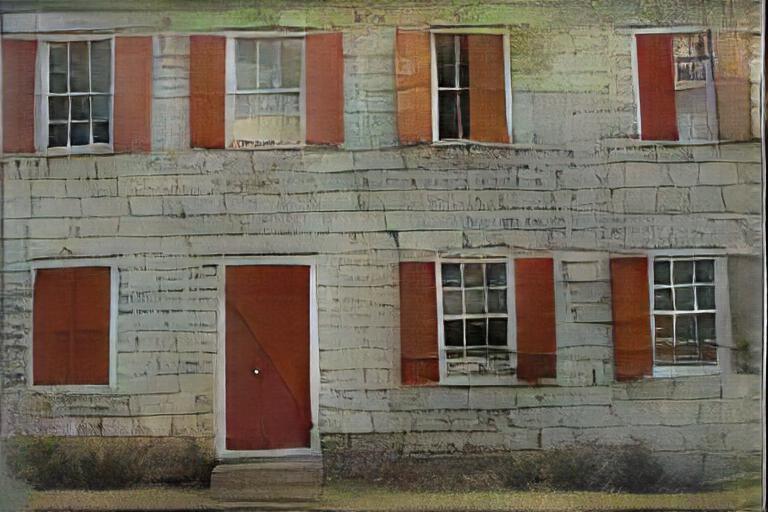}} &
\raisebox{-0.5\height}{\includegraphics[width=0.45\linewidth]{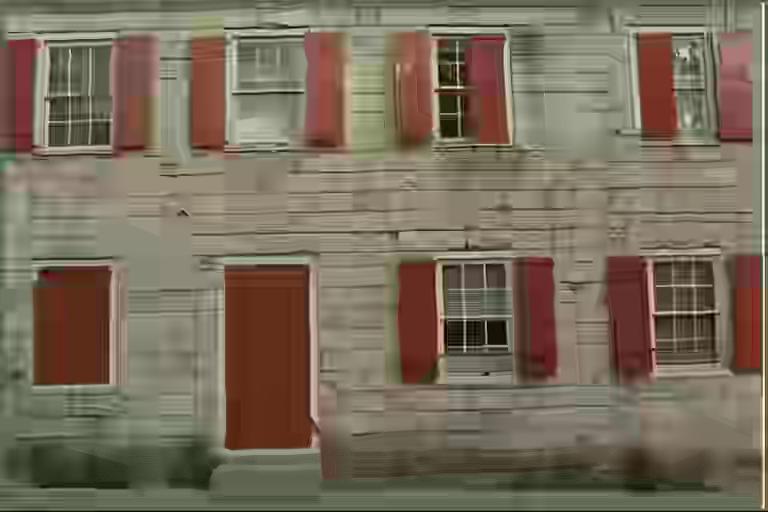}}&
\rotatebox[origin=c]{90}{$0.043$ bpp}\vspace{2pt} \\
\rotatebox[origin=c]{90}{$0.030$ bpp}&
\raisebox{-0.5\height}{\includegraphics[width=0.45\linewidth]{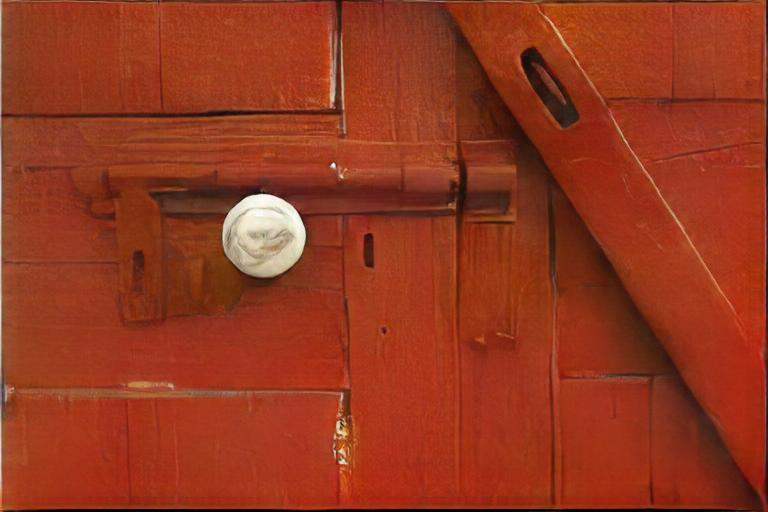}} &
\raisebox{-0.5\height}{\includegraphics[width=0.45\linewidth]{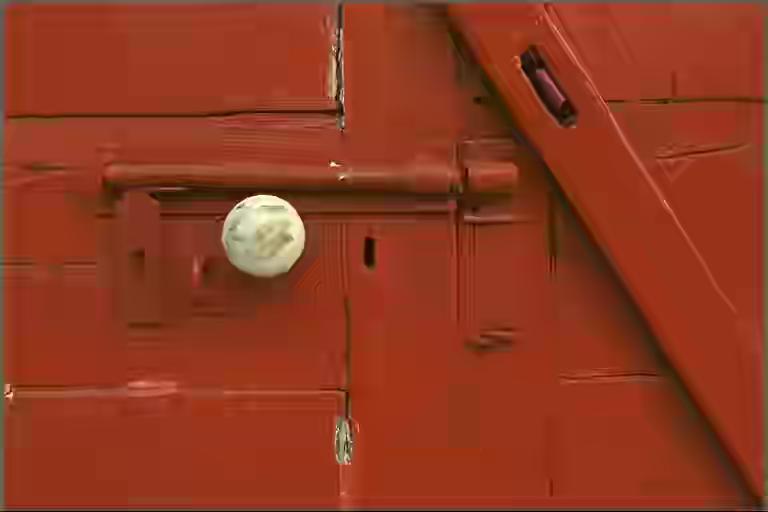}}&
\rotatebox[origin=c]{90}{$0.031$ bpp}\vspace{2pt} \\
\rotatebox[origin=c]{90}{$0.034$ bpp}&
\raisebox{-0.5\height}{\includegraphics[width=0.45\linewidth]{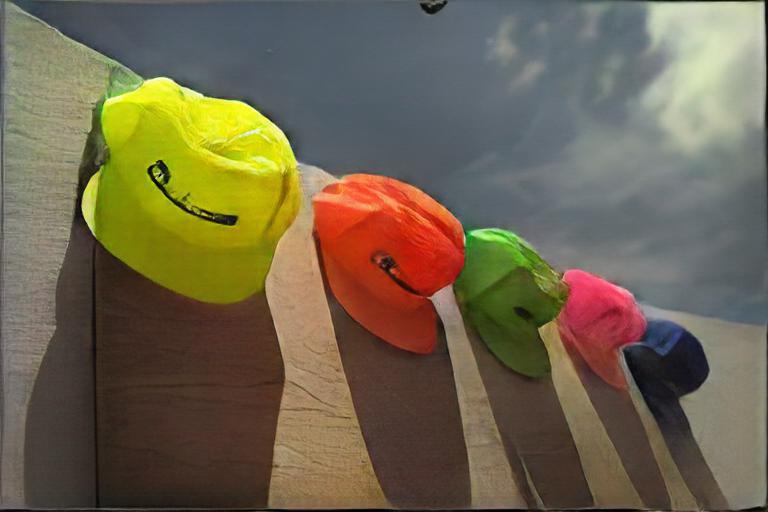}} &
\raisebox{-0.5\height}{\includegraphics[width=0.45\linewidth]{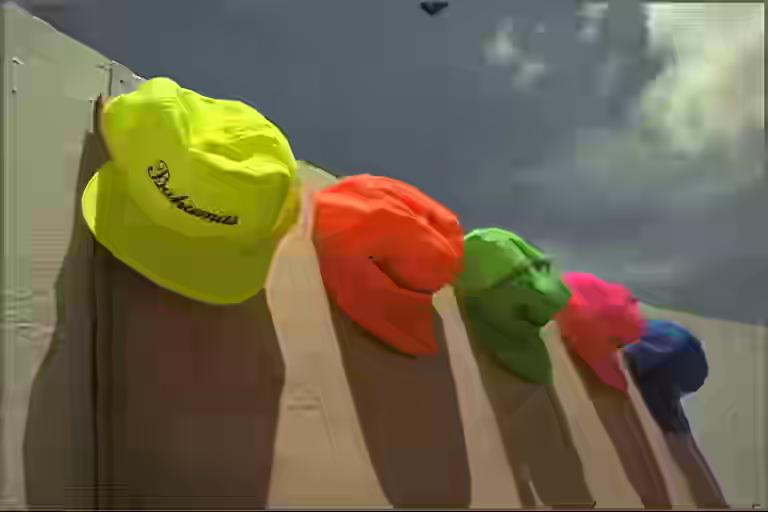}}&
\rotatebox[origin=c]{90}{$0.035$ bpp}\vspace{2pt} \\
\rotatebox[origin=c]{90}{$0.032$ bpp}&
\raisebox{-0.5\height}{\includegraphics[width=0.45\linewidth]{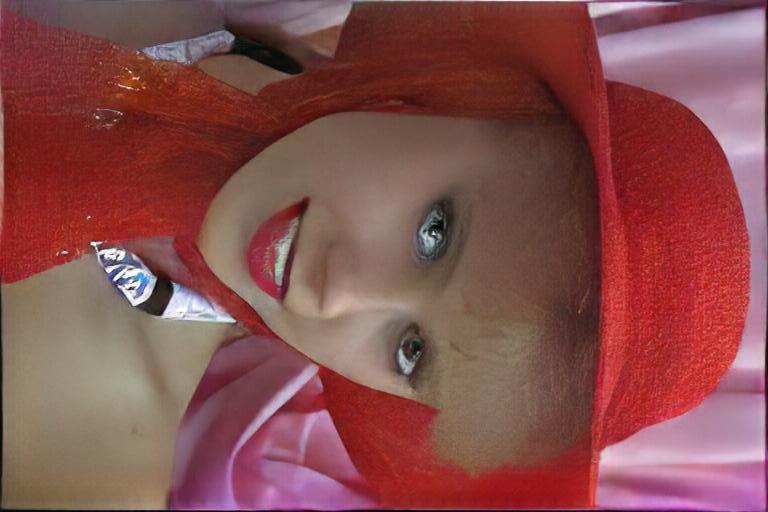}} &
\raisebox{-0.5\height}{\includegraphics[width=0.45\linewidth]{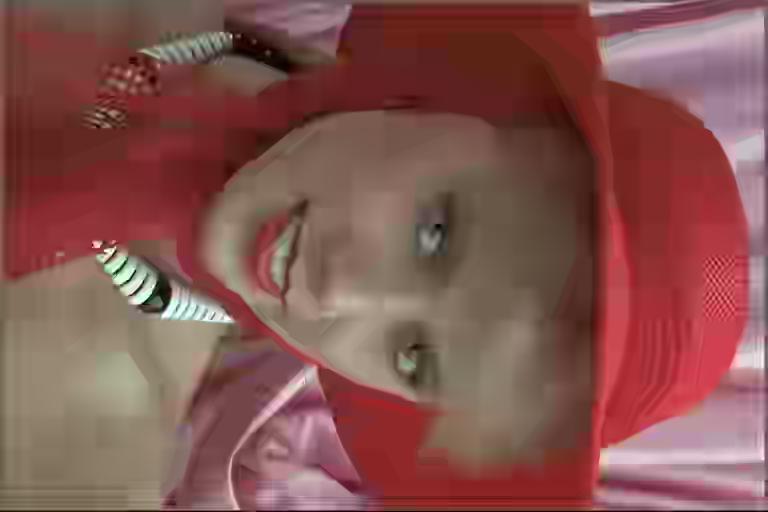}}&
\rotatebox[origin=c]{90}{$0.032$ bpp}\vspace{2pt} \\
\rotatebox[origin=c]{90}{$0.036$ bpp}&
\raisebox{-0.5\height}{\includegraphics[width=0.45\linewidth]{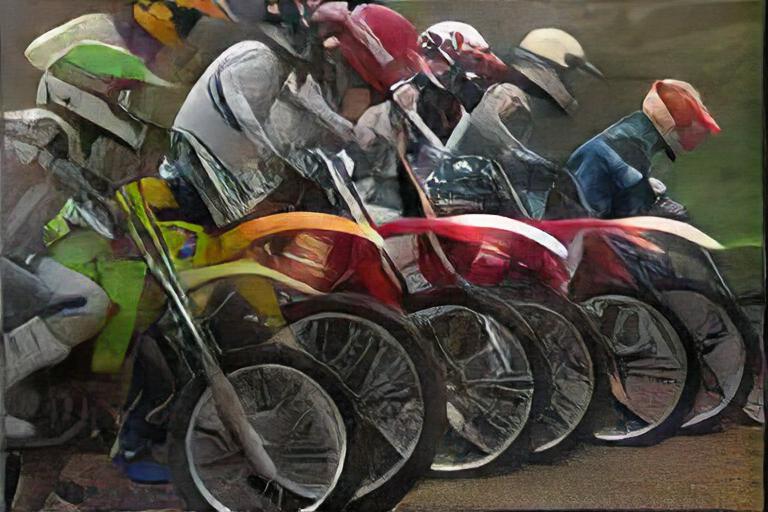}} &
\raisebox{-0.5\height}{\includegraphics[width=0.45\linewidth]{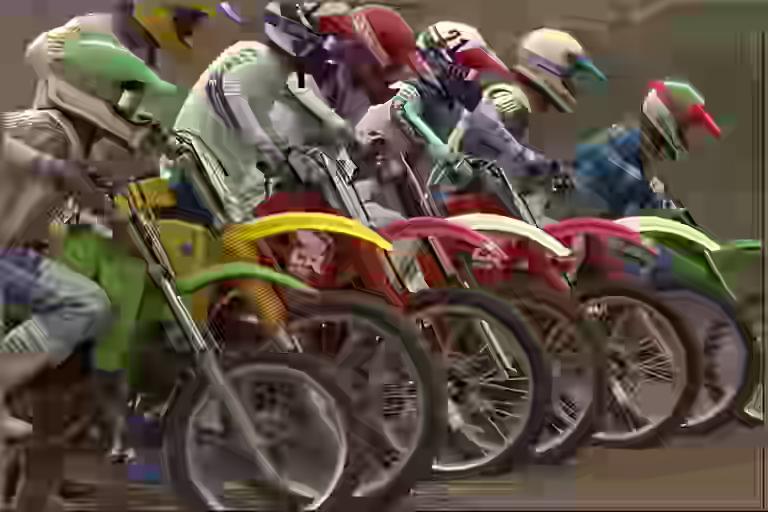}}&
\rotatebox[origin=c]{90}{$0.043$0 bpp}\vspace{2pt} \\
\end{tabular}}

%% file: fig_appendix_raise1k.tex
{\setlength{\tabcolsep}{1pt}
\begin{tabular}{rccl}
&Ours&BPG\\
\rotatebox[origin=c]{90}{$0.036$ bpp}&
\raisebox{-0.5\height}{\includegraphics[width=0.45\linewidth]{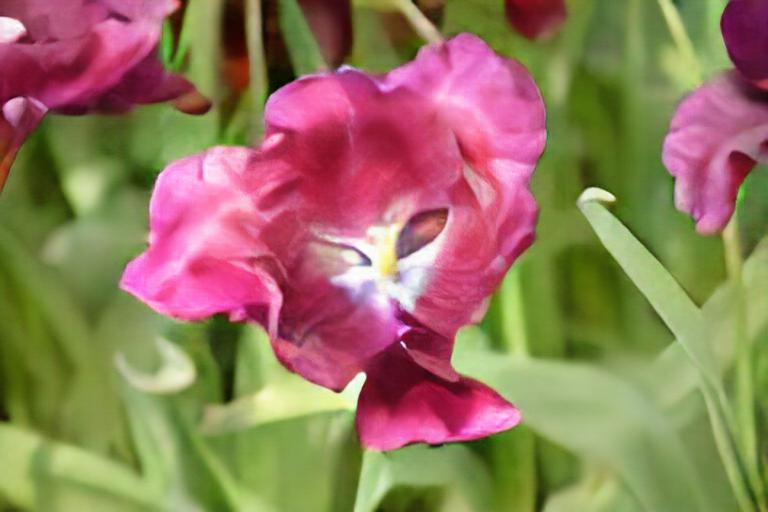}} &
\raisebox{-0.5\height}{\includegraphics[width=0.45\linewidth]{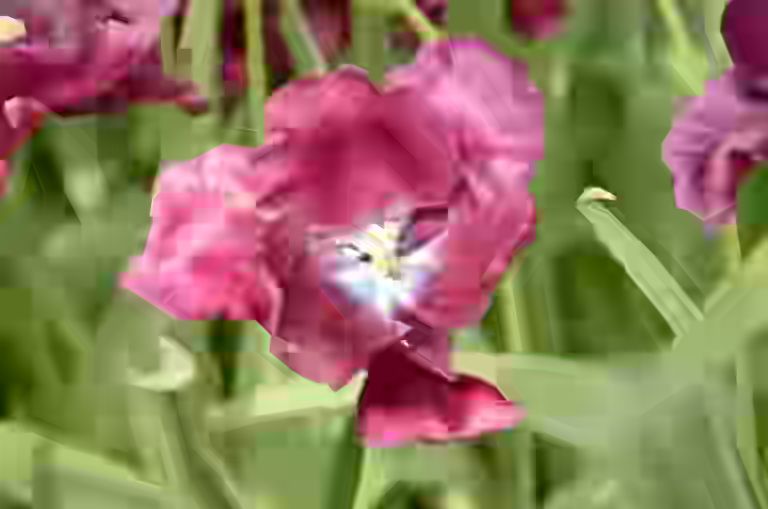}}&
\rotatebox[origin=c]{90}{$0.038$ bpp}\vspace{2pt} \\
\rotatebox[origin=c]{90}{$0.035$ bpp}&
\raisebox{-0.5\height}{\includegraphics[width=0.45\linewidth]{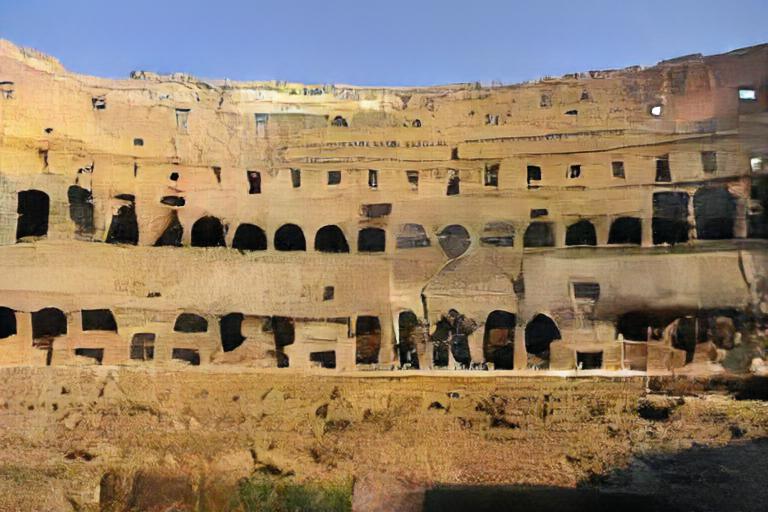}} &
\raisebox{-0.5\height}{\includegraphics[width=0.45\linewidth]{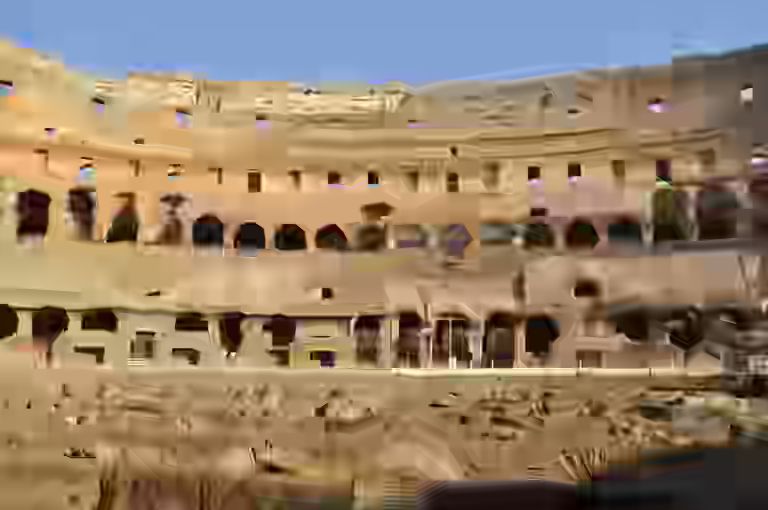}}&
\rotatebox[origin=c]{90}{$0.053$ bpp}\vspace{2pt} \\
\rotatebox[origin=c]{90}{$0.034$ bpp}&
\raisebox{-0.5\height}{\includegraphics[width=0.45\linewidth]{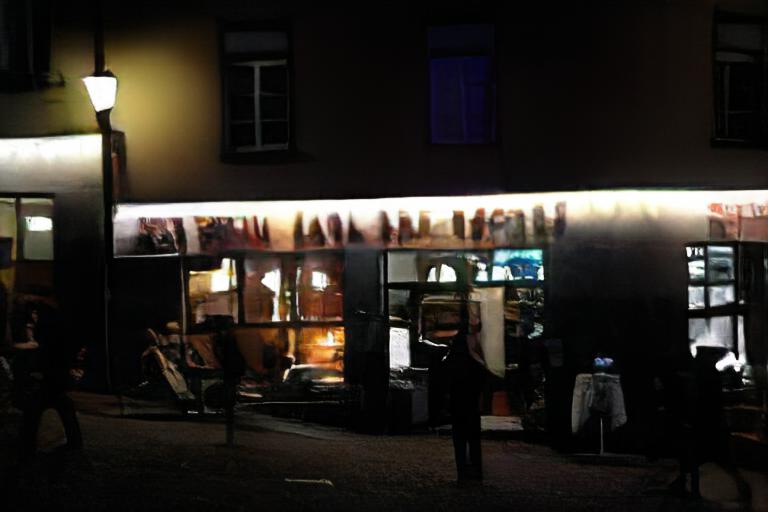}} &
\raisebox{-0.5\height}{\includegraphics[width=0.45\linewidth]{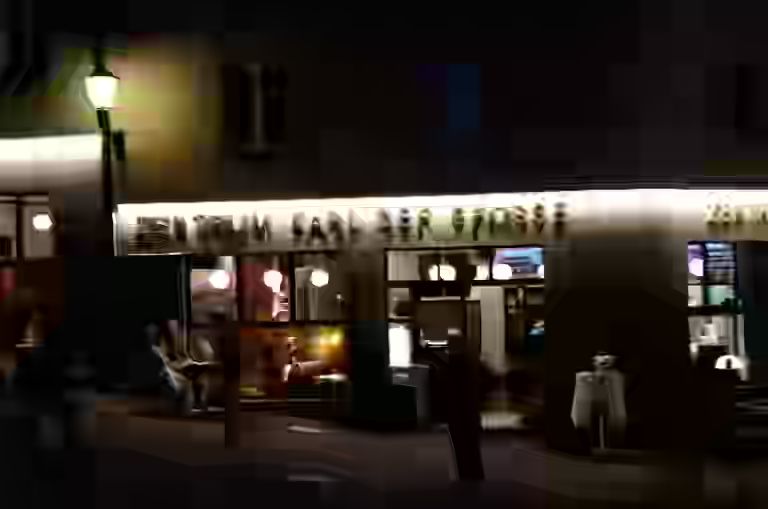}}&
\rotatebox[origin=c]{90}{$0.038$ bpp}\vspace{2pt} \\
\rotatebox[origin=c]{90}{$0.036$ bpp}&
\raisebox{-0.5\height}{\includegraphics[width=0.45\linewidth]{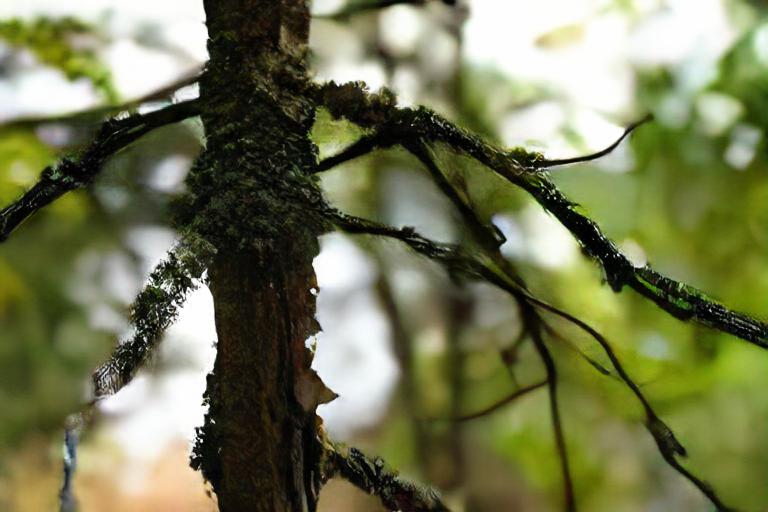}} &
\raisebox{-0.5\height}{\includegraphics[width=0.45\linewidth]{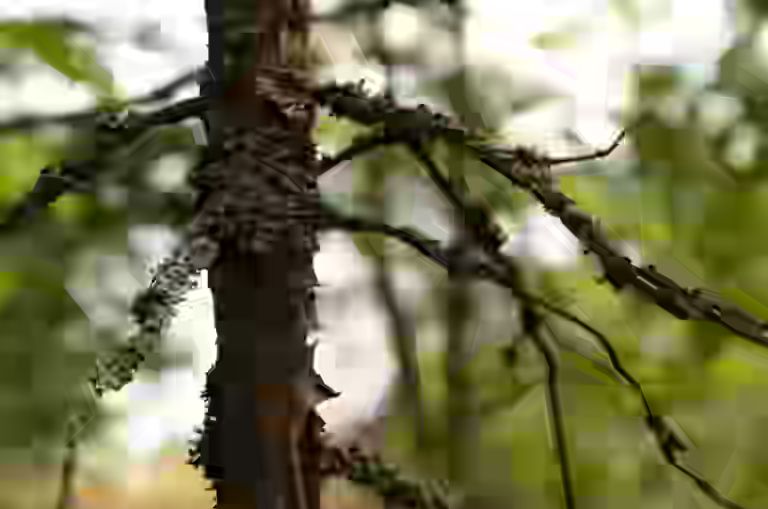}}&
\rotatebox[origin=c]{90}{$0.044$ bpp}\vspace{2pt} \\
\rotatebox[origin=c]{90}{$0.034$ bpp}&
\raisebox{-0.5\height}{\includegraphics[width=0.45\linewidth]{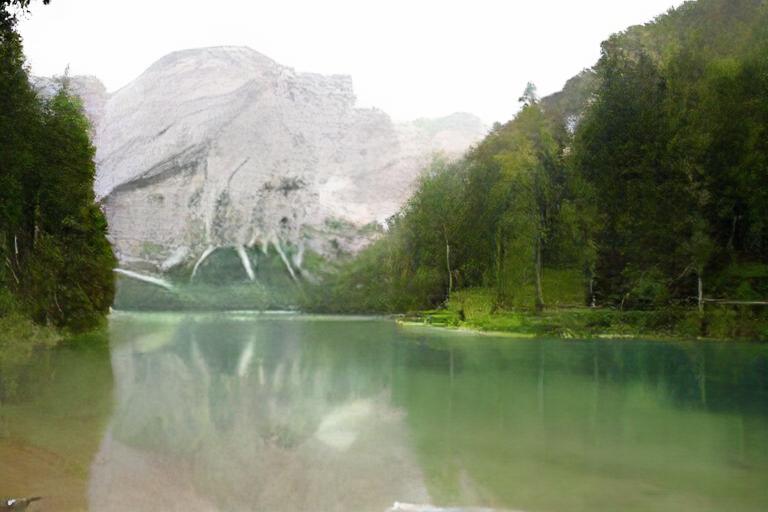}} &
\raisebox{-0.5\height}{\includegraphics[width=0.45\linewidth]{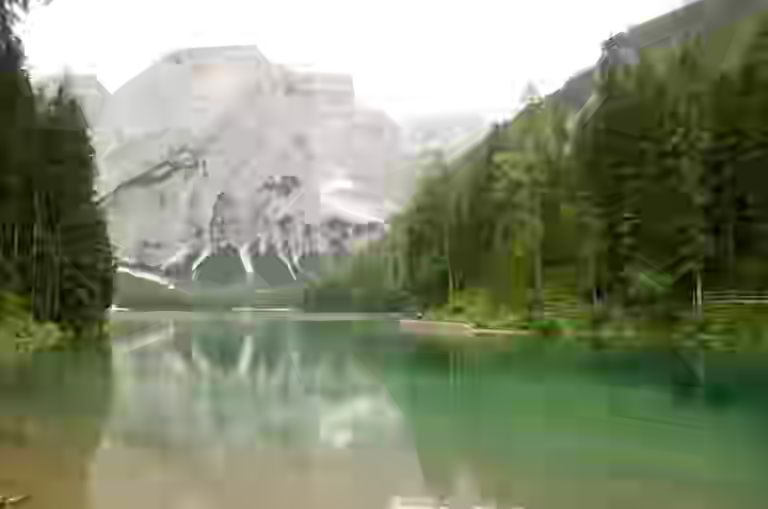}}&
\rotatebox[origin=c]{90}{$0.038$0 bpp}\vspace{2pt} \\
\end{tabular}}

%% file: fig_appendix_cityscapes.tex
{\setlength{\tabcolsep}{1pt}
\begin{tabular}{rccl}
&Ours&BPG\\
\rotatebox[origin=c]{90}{$0.036$ bpp}&
\raisebox{-0.5\height}{\includegraphics[width=0.45\linewidth]{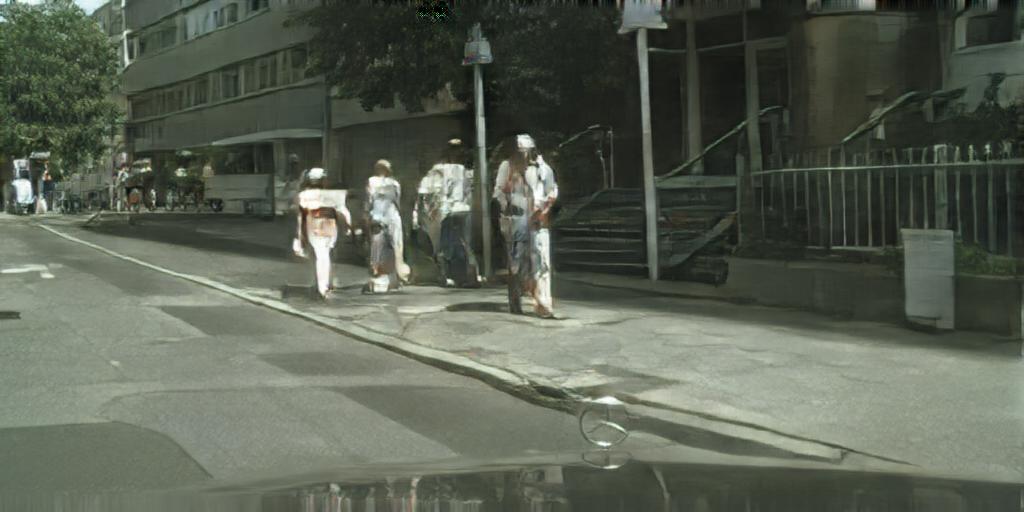}} &
\raisebox{-0.5\height}{\includegraphics[width=0.45\linewidth]{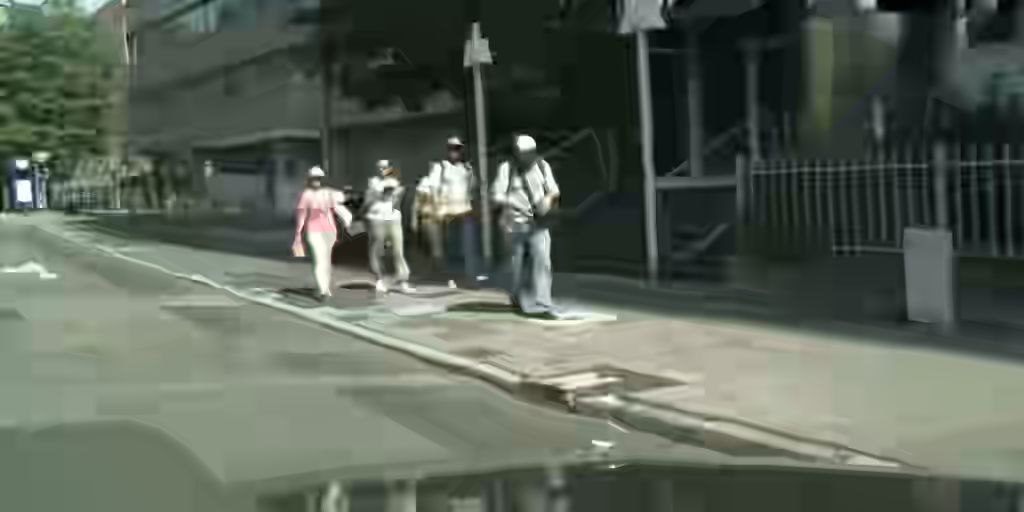}}&
\rotatebox[origin=c]{90}{$0.040$ bpp}\vspace{2pt} \\
\rotatebox[origin=c]{90}{$0.036$ bpp}&
\raisebox{-0.5\height}{\includegraphics[width=0.45\linewidth]{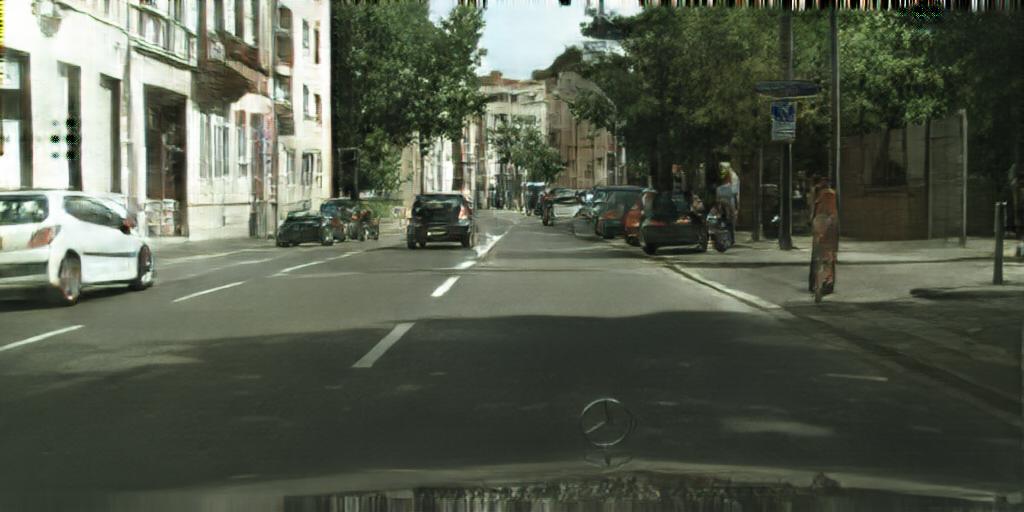}} &
\raisebox{-0.5\height}{\includegraphics[width=0.45\linewidth]{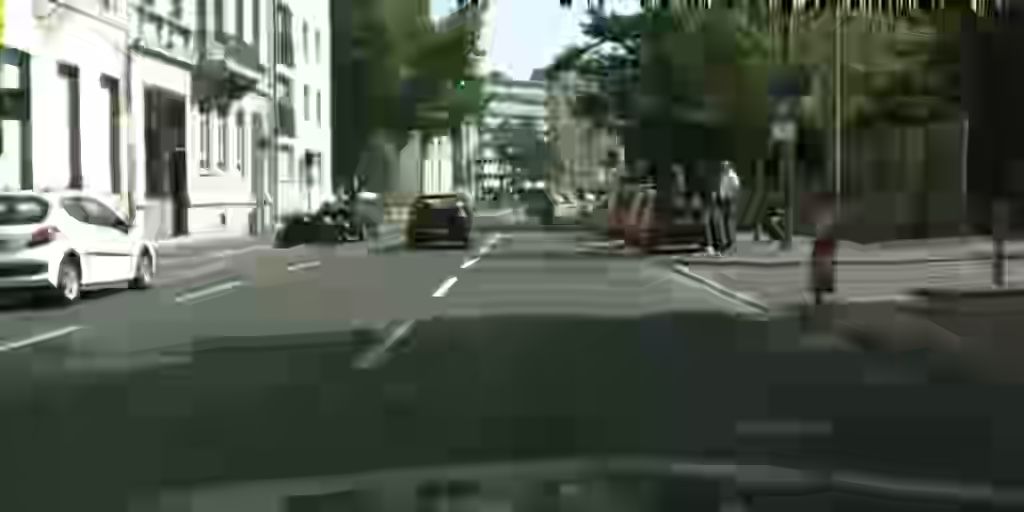}}&
\rotatebox[origin=c]{90}{$0.038$ bpp}\vspace{2pt} \\
\rotatebox[origin=c]{90}{$0.036$ bpp}&
\raisebox{-0.5\height}{\includegraphics[width=0.45\linewidth]{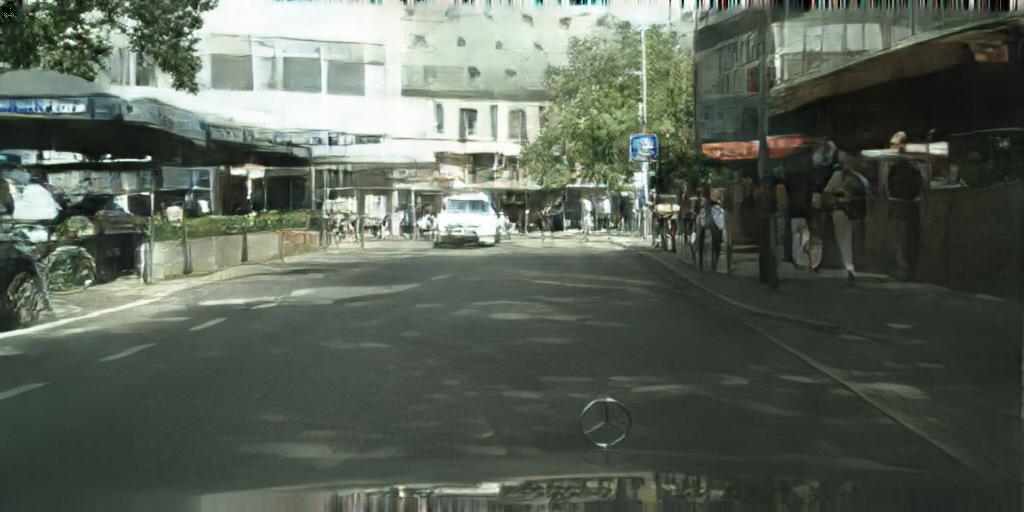}} &
\raisebox{-0.5\height}{\includegraphics[width=0.45\linewidth]{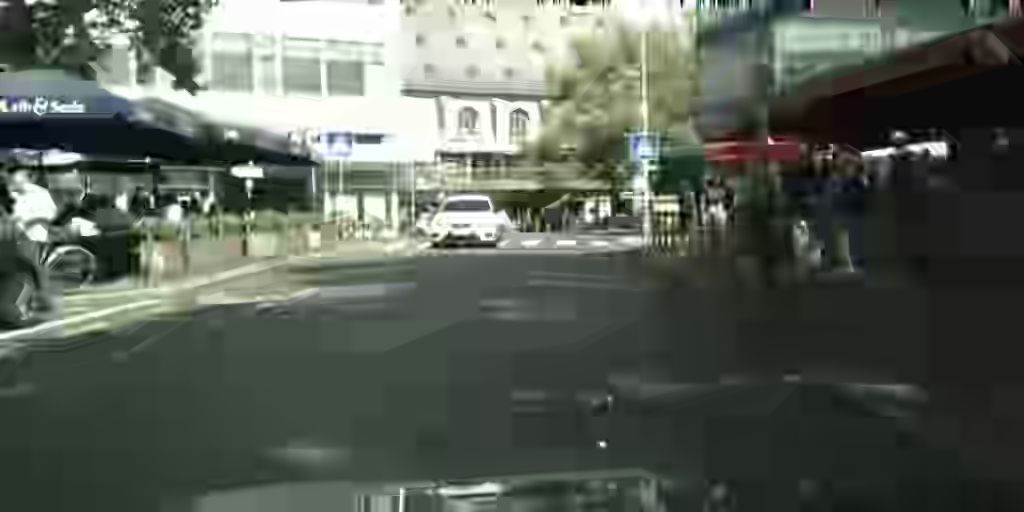}}&
\rotatebox[origin=c]{90}{$0.043$ bpp}\vspace{2pt} \\
\rotatebox[origin=c]{90}{$0.036$ bpp}&
\raisebox{-0.5\height}{\includegraphics[width=0.45\linewidth]{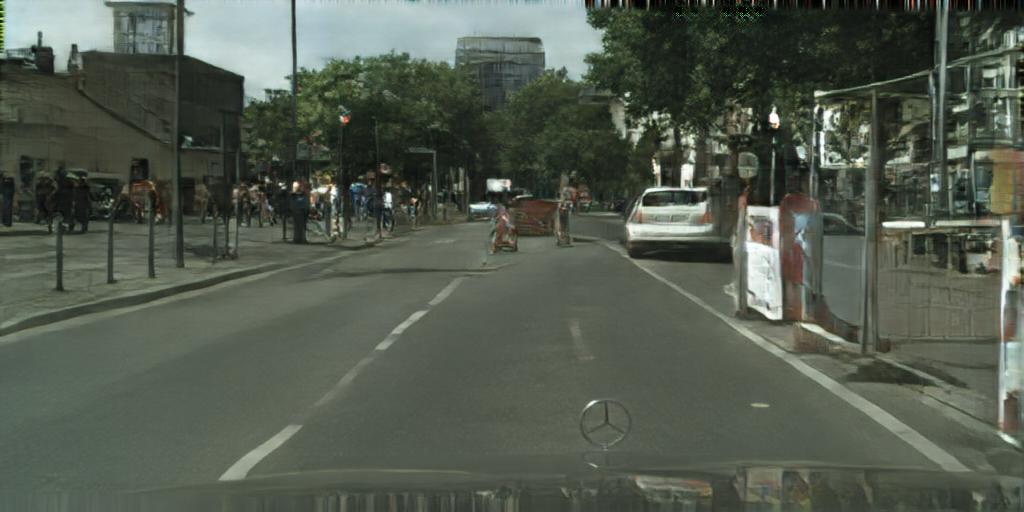}} &
\raisebox{-0.5\height}{\includegraphics[width=0.45\linewidth]{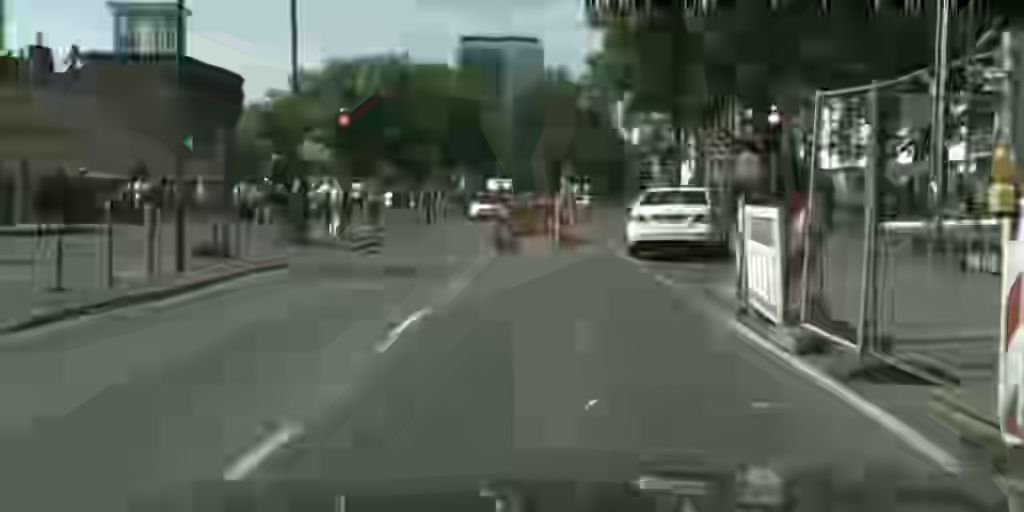}}&
\rotatebox[origin=c]{90}{$0.037$ bpp}\vspace{2pt} \\
\rotatebox[origin=c]{90}{$0.036$ bpp}&
\raisebox{-0.5\height}{\includegraphics[width=0.45\linewidth]{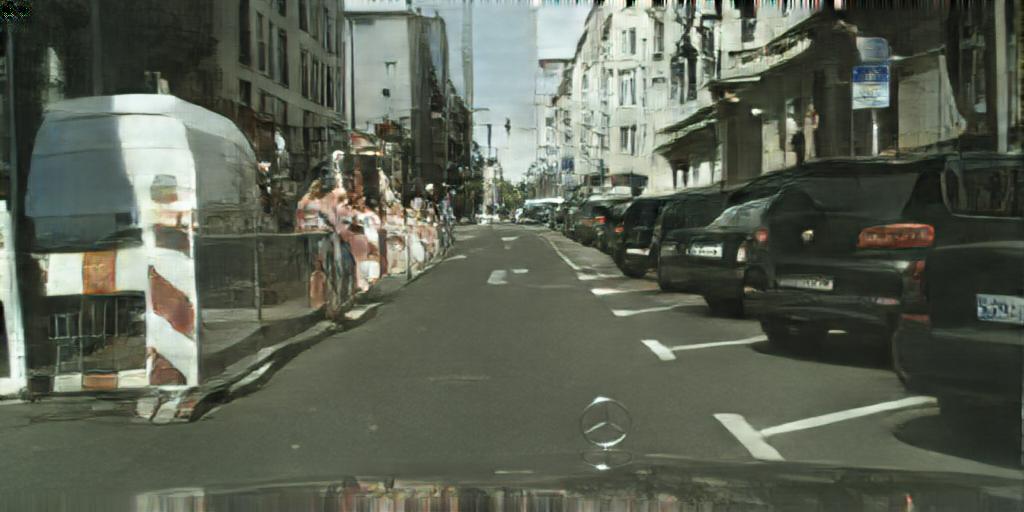}} &
\raisebox{-0.5\height}{\includegraphics[width=0.45\linewidth]{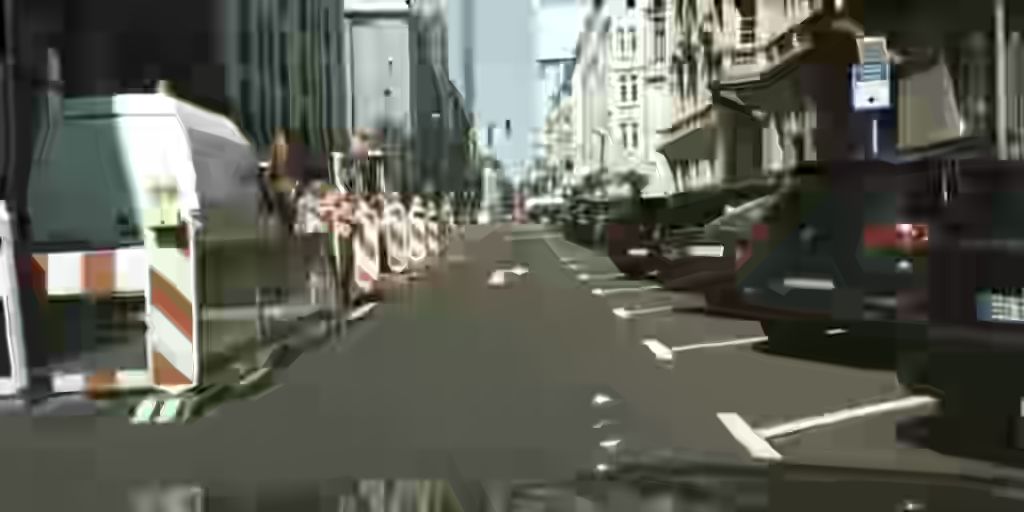}}&
\rotatebox[origin=c]{90}{$0.040$0 bpp}\vspace{2pt} \\
\end{tabular}}

%% file: fig_full_classes_2.tex
\begin{tabular}{ccc}
\begin{overpic}[width=0.33\linewidth]{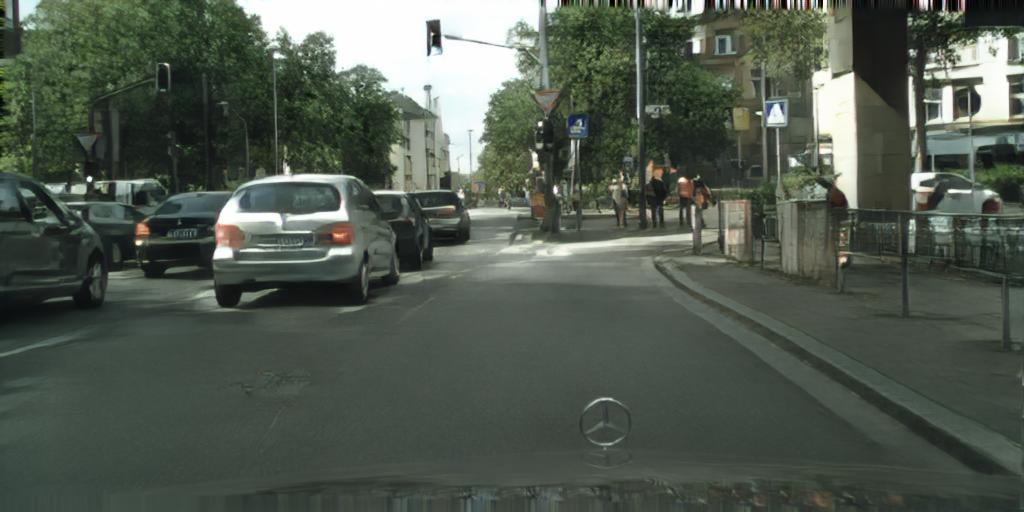}
\put(1,1){\setlength{\fboxsep}{0pt}\setlength{\fboxrule}{0.2pt}\fbox{\includegraphics[width=0.082500\linewidth]{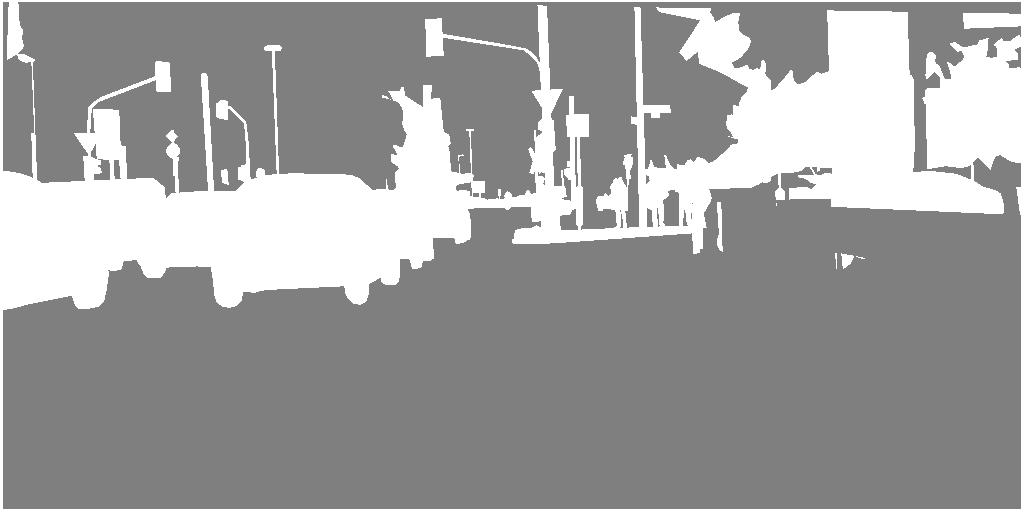}}}
\end{overpic} &
\begin{overpic}[width=0.33\linewidth]{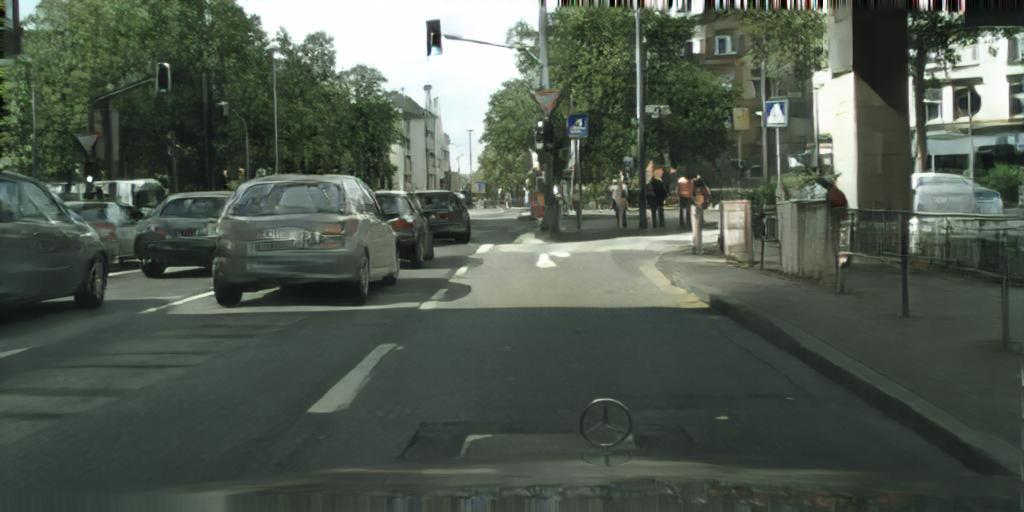}
\put(1,1){\setlength{\fboxsep}{0pt}\setlength{\fboxrule}{0.2pt}\fbox{\includegraphics[width=0.082500\linewidth]{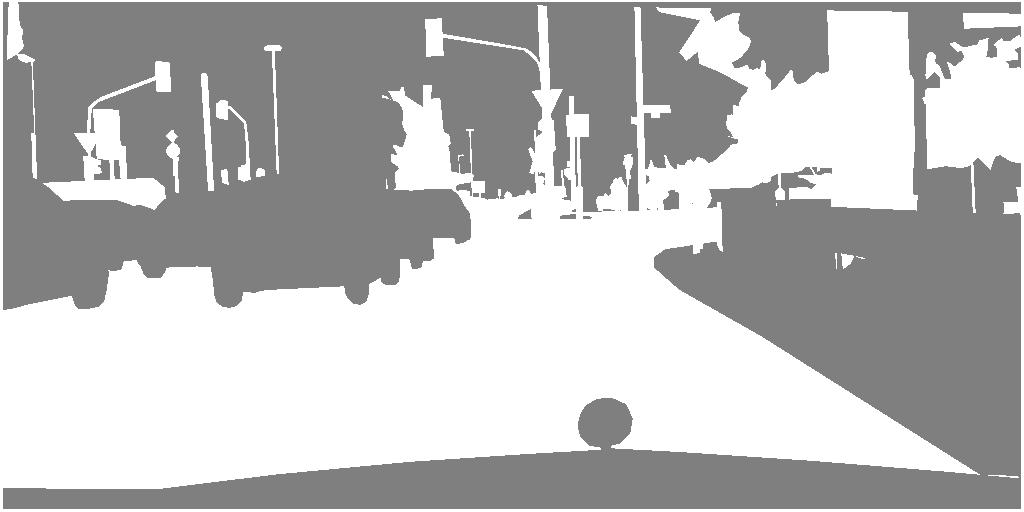}}}
\end{overpic} &
\begin{overpic}[width=0.33\linewidth]{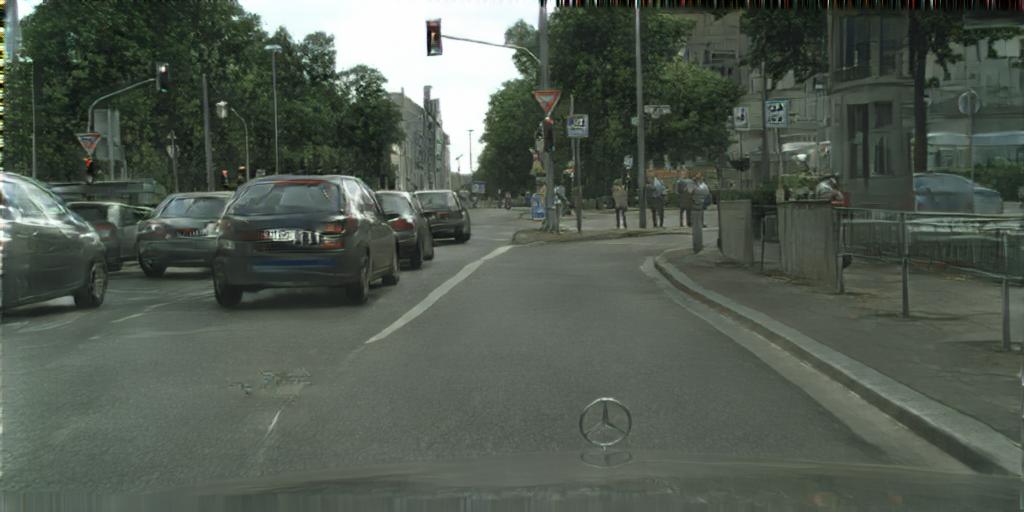}
\put(1,1){\setlength{\fboxsep}{0pt}\setlength{\fboxrule}{0.2pt}\fbox{\includegraphics[width=0.082500\linewidth]{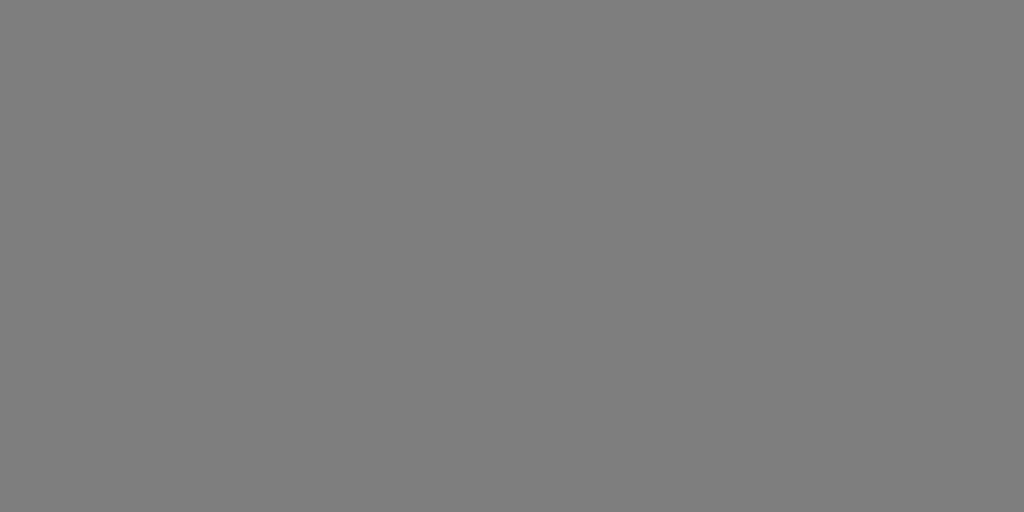}}}
\end{overpic} \vspace{-2.6ex} \\
\emph{road} ($0.077$ bpp) & \emph{car}  ($0.108$ bpp) & everything ($0.041$ bpp)    \\
%
\begin{overpic}[width=0.33\linewidth]{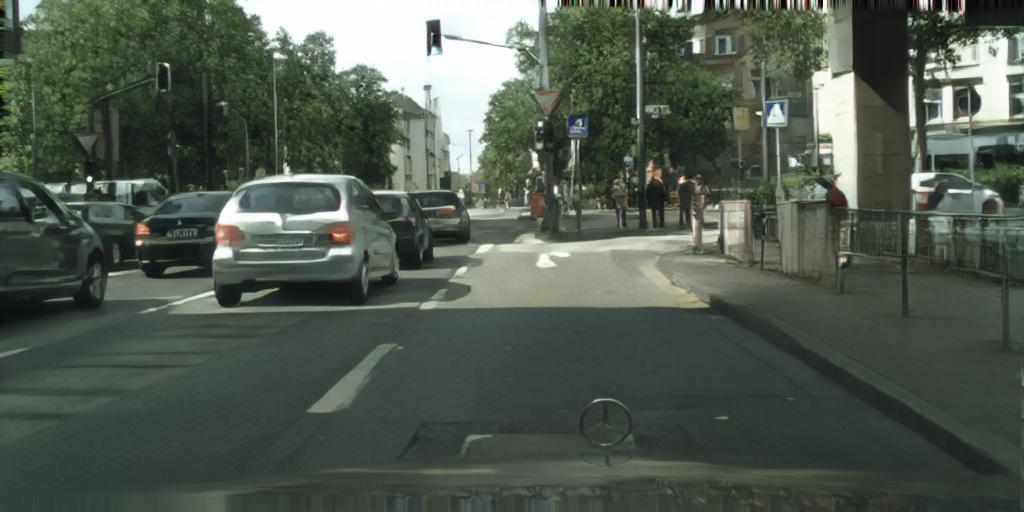}
\put(1,1){\setlength{\fboxsep}{0pt}\setlength{\fboxrule}{0.2pt}\fbox{\includegraphics[width=0.082500\linewidth]{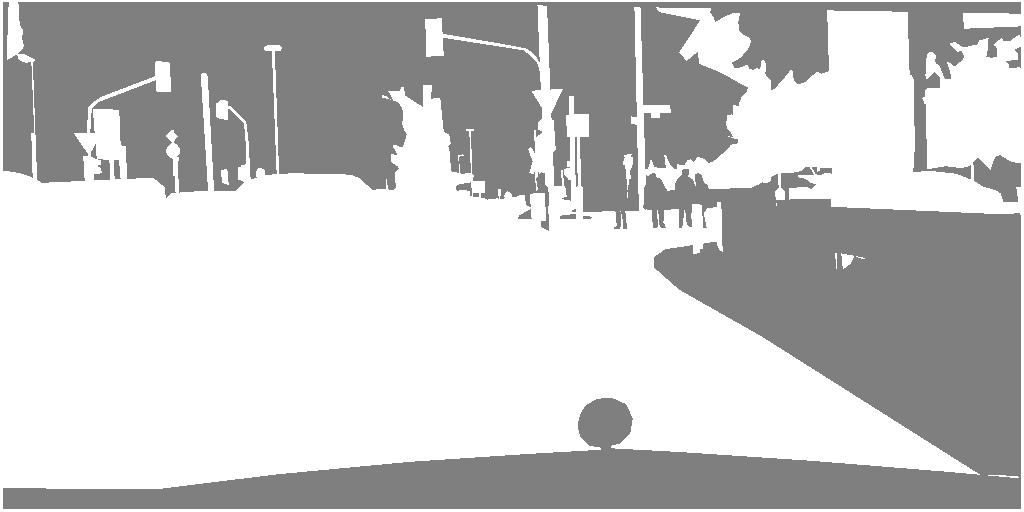}}}
\end{overpic} &
\begin{overpic}[width=0.33\linewidth]{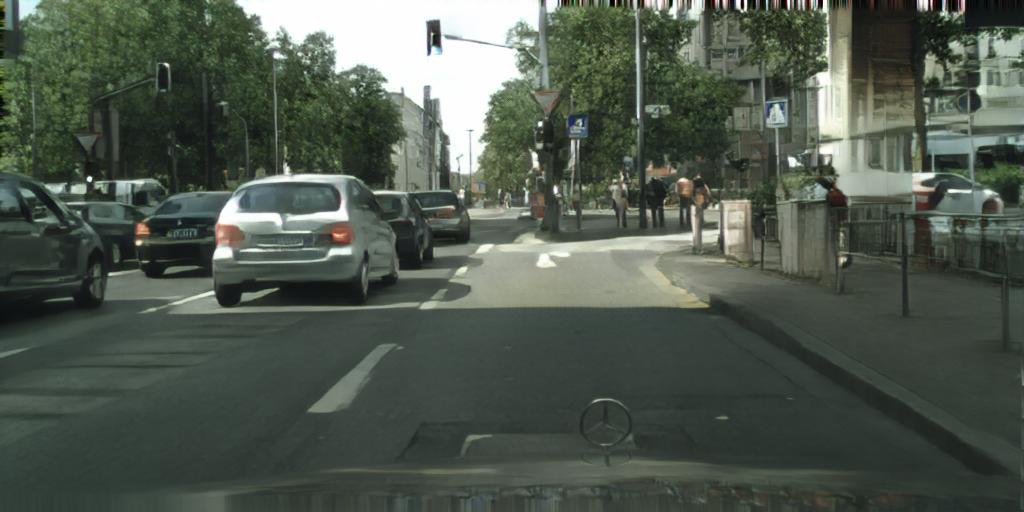}
\put(1,1){\setlength{\fboxsep}{0pt}\setlength{\fboxrule}{0.2pt}\fbox{\includegraphics[width=0.082500\linewidth]{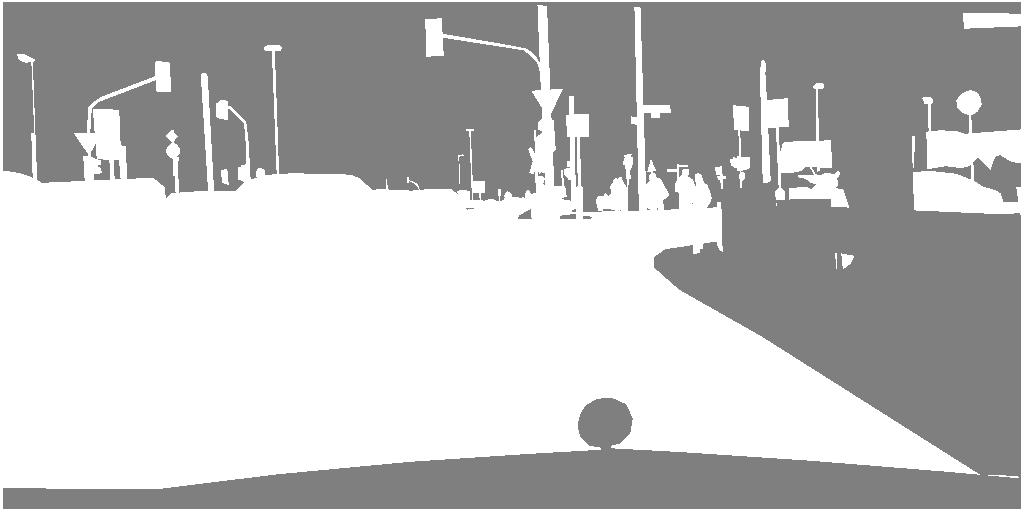}}}
\end{overpic} &
\begin{overpic}[width=0.33\linewidth]{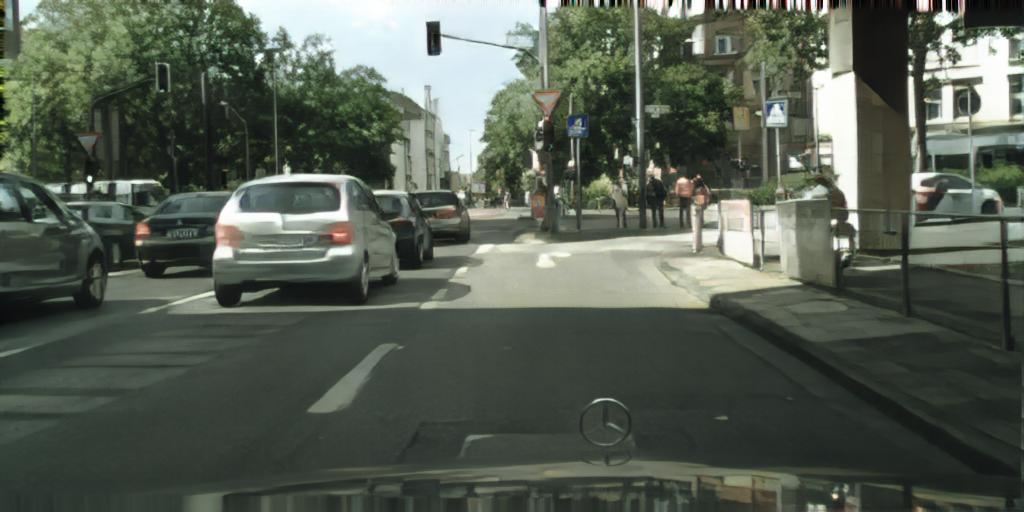}
\put(1,1){\setlength{\fboxsep}{0pt}\setlength{\fboxrule}{0.2pt}\fbox{\includegraphics[width=0.082500\linewidth]{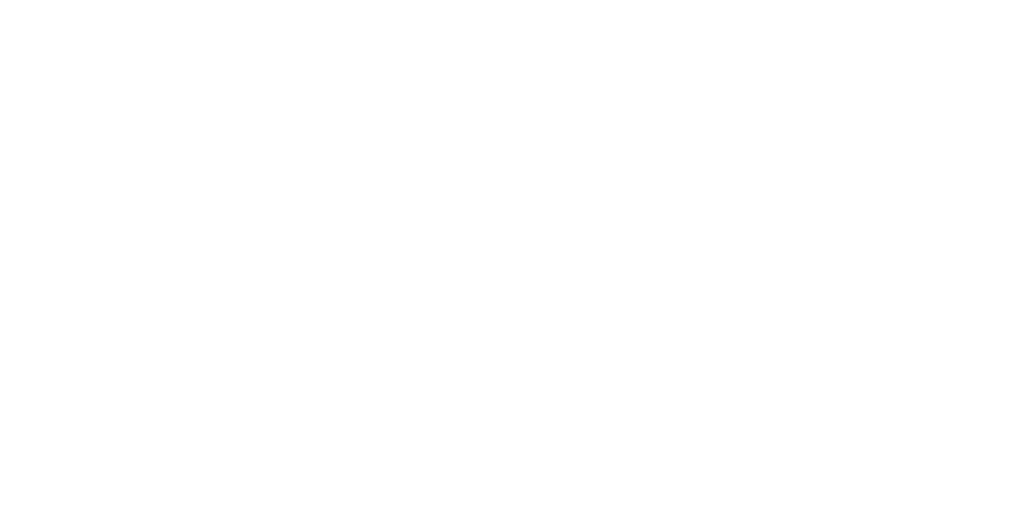}}}
\end{overpic} \vspace{-2.6ex} \\
\emph{people} ($0.120$ bpp) & \emph{building}  ($0.110$ bpp) & no synth ($0.186$ bpp) 
\\
\\
\begin{overpic}[width=0.33\linewidth]{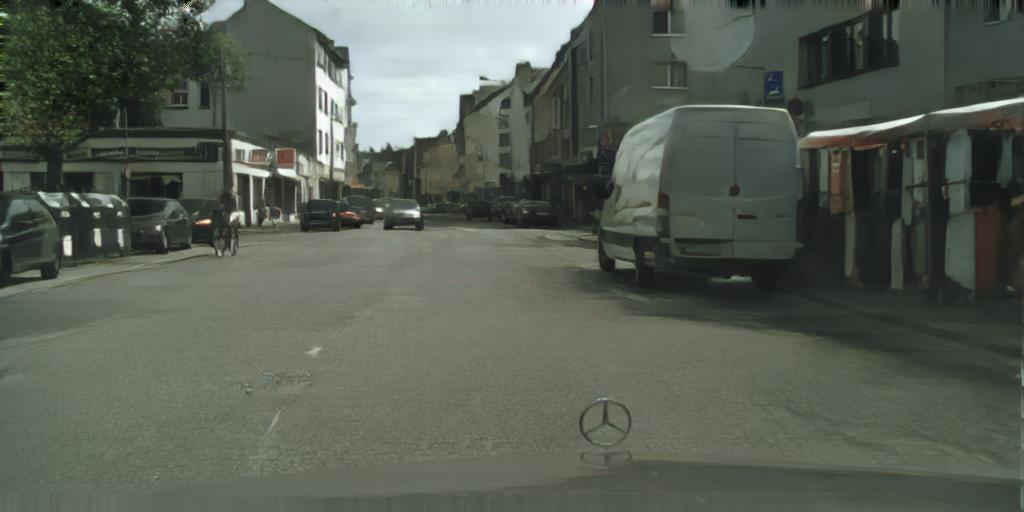}
\put(1,1){\setlength{\fboxsep}{0pt}\setlength{\fboxrule}{0.2pt}\fbox{\includegraphics[width=0.082500\linewidth]{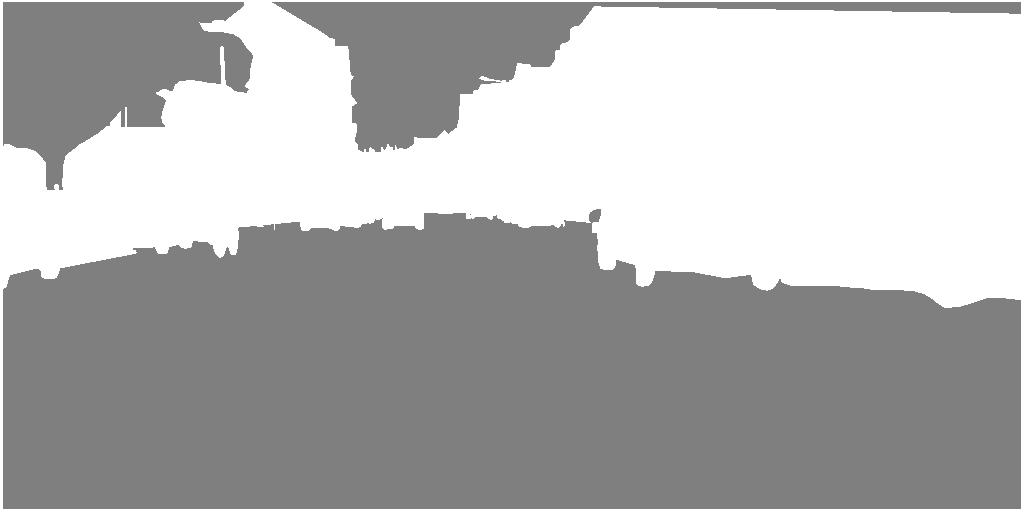}}}
\end{overpic} &
\begin{overpic}[width=0.33\linewidth]{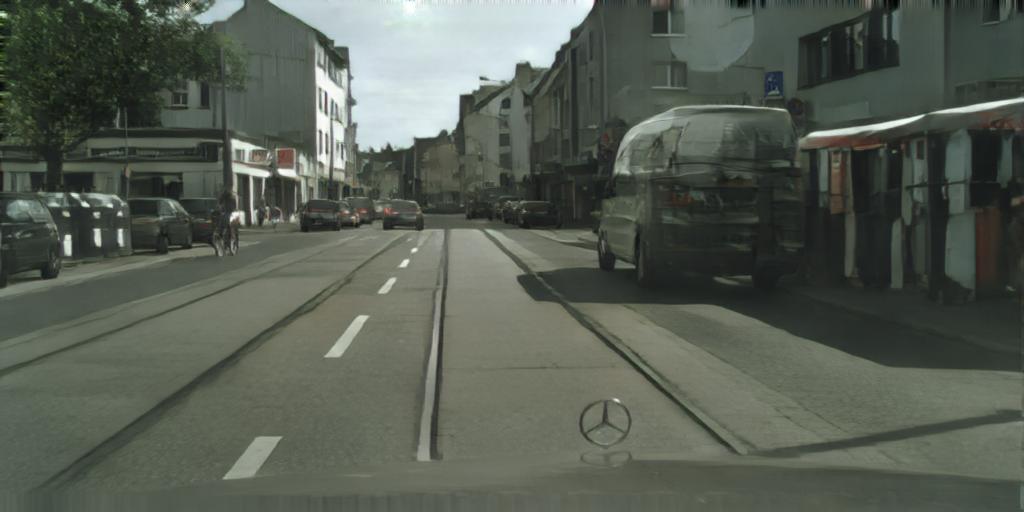}
\put(1,1){\setlength{\fboxsep}{0pt}\setlength{\fboxrule}{0.2pt}\fbox{\includegraphics[width=0.082500\linewidth]{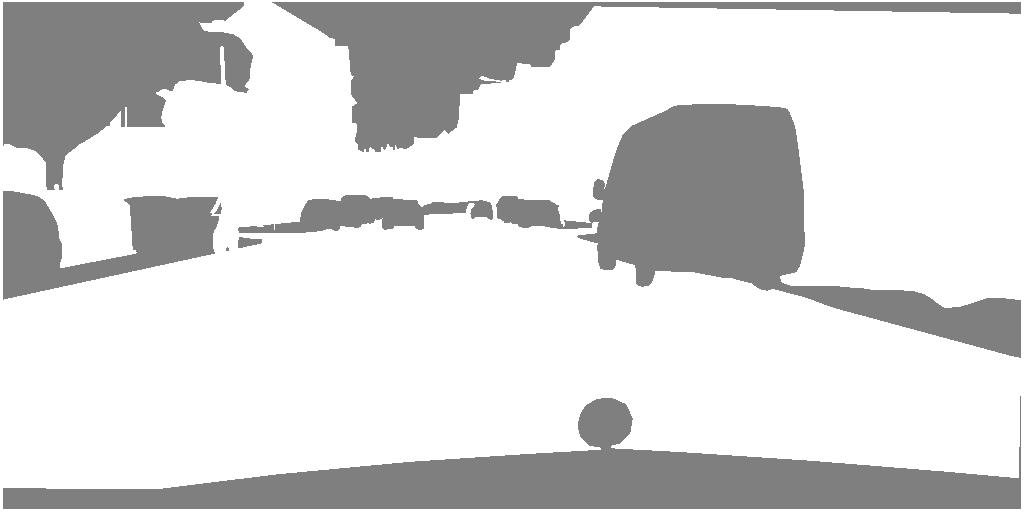}}}
\end{overpic} &
\begin{overpic}[width=0.33\linewidth]{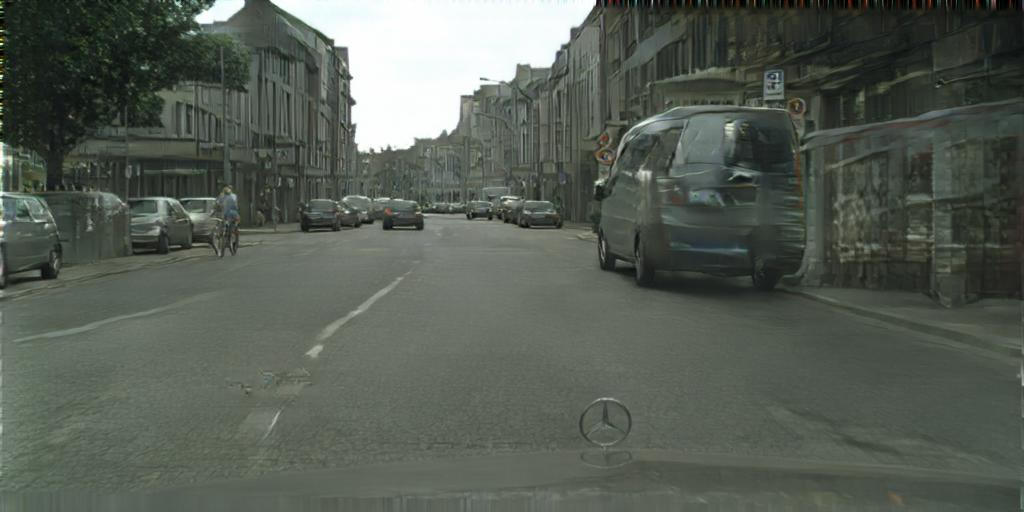}
\put(1,1){\setlength{\fboxsep}{0pt}\setlength{\fboxrule}{0.2pt}\fbox{\includegraphics[width=0.082500\linewidth]{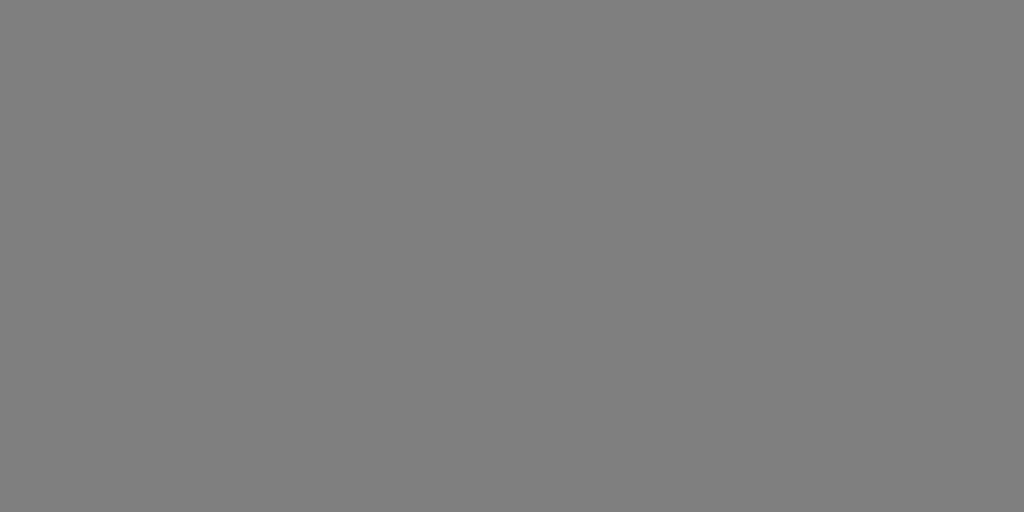}}}
\end{overpic} \vspace{-2.6ex} \\
\emph{road} ($0.092$ bpp) & \emph{car}  ($0.134$ bpp) & everything ($0.034$ bpp)    \\
%
\begin{overpic}[width=0.33\linewidth]{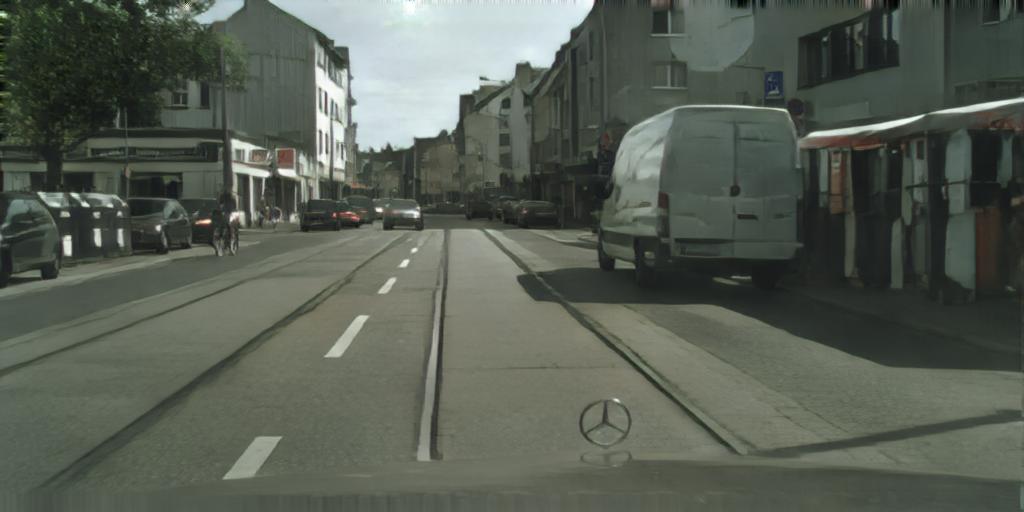}
\put(1,1){\setlength{\fboxsep}{0pt}\setlength{\fboxrule}{0.2pt}\fbox{\includegraphics[width=0.082500\linewidth]{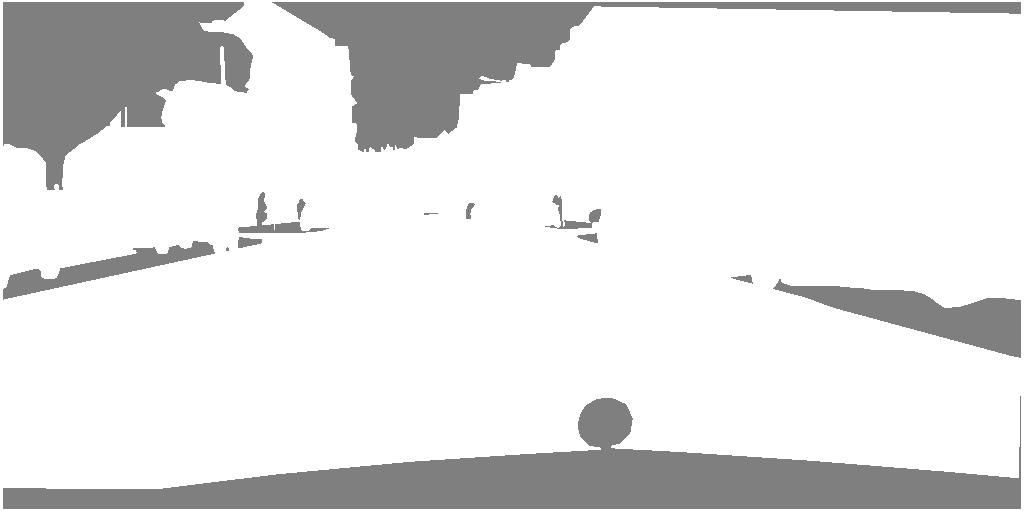}}}
\end{overpic} &
\begin{overpic}[width=0.33\linewidth]{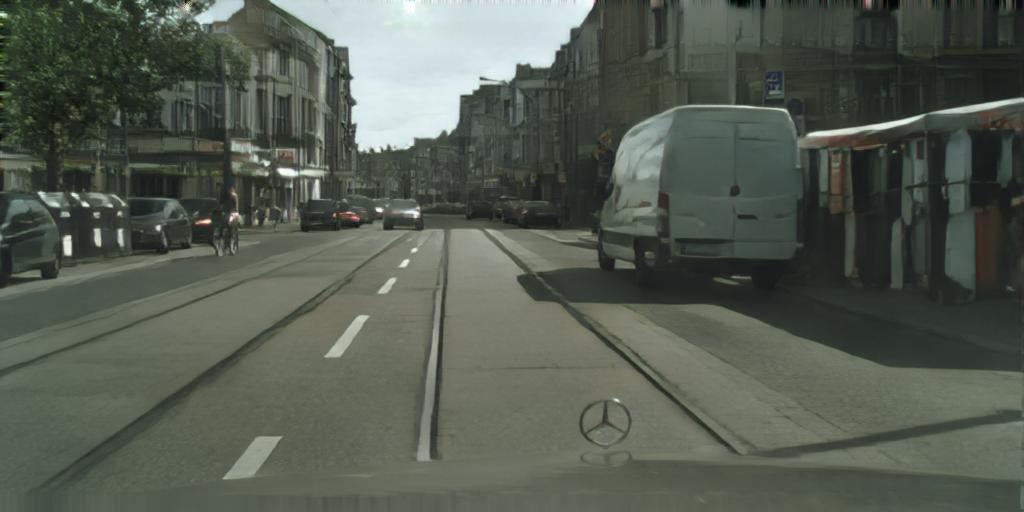}
\put(1,1){\setlength{\fboxsep}{0pt}\setlength{\fboxrule}{0.2pt}\fbox{\includegraphics[width=0.082500\linewidth]{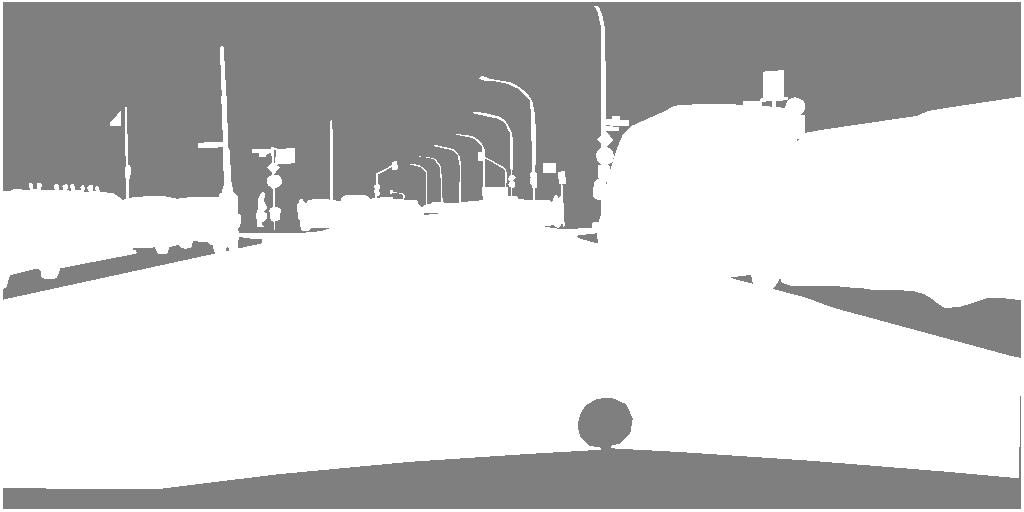}}}
\end{overpic} &
\begin{overpic}[width=0.33\linewidth]{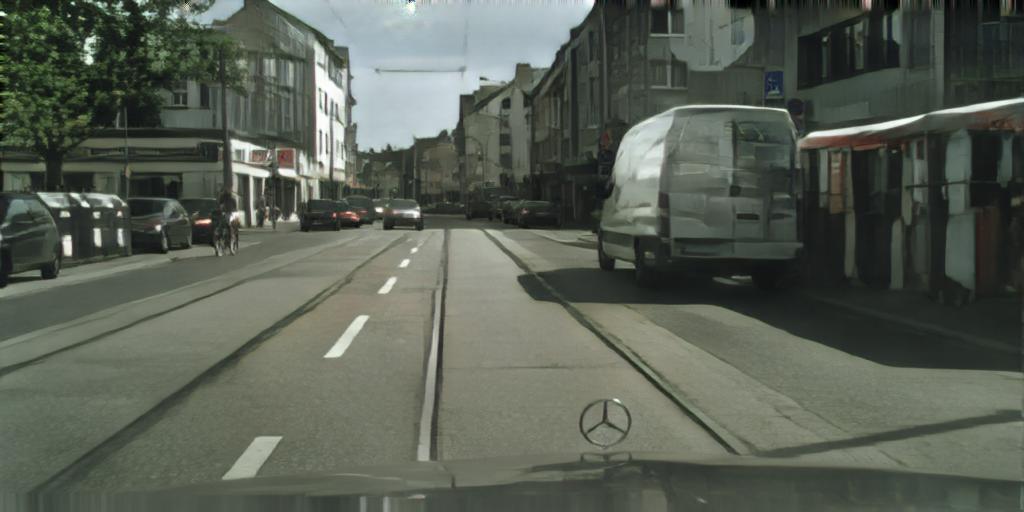}
\put(1,1){\setlength{\fboxsep}{0pt}\setlength{\fboxrule}{0.2pt}\fbox{\includegraphics[width=0.082500\linewidth]{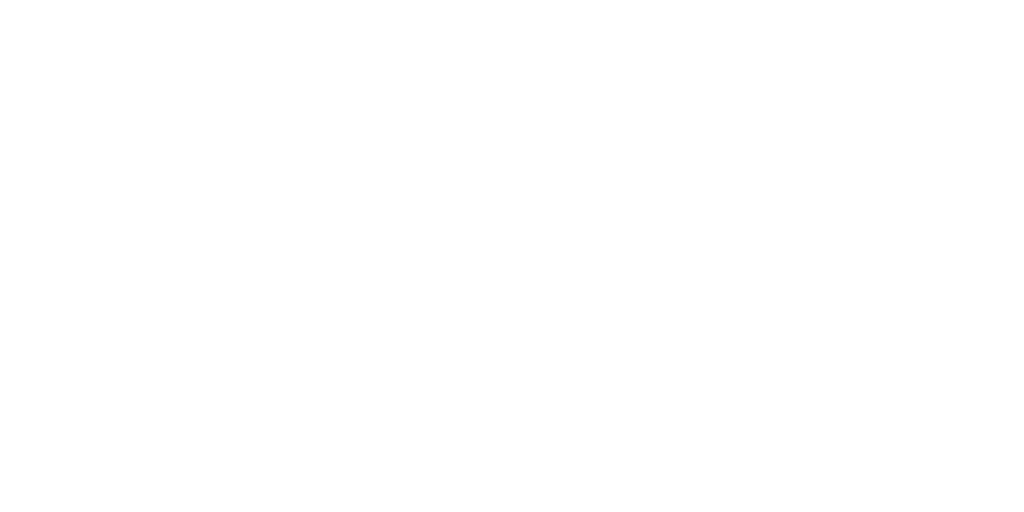}}}
\end{overpic} \vspace{-2.6ex} \\
\emph{people} ($0.147$ bpp) & \emph{building}  ($0.119$ bpp) & no synth ($0.179$ bpp) \\
\end{tabular}

%% file: fig_full_boxes_2.tex
\begin{tabular}{ccc}
\includegraphics[width=0.32\linewidth]{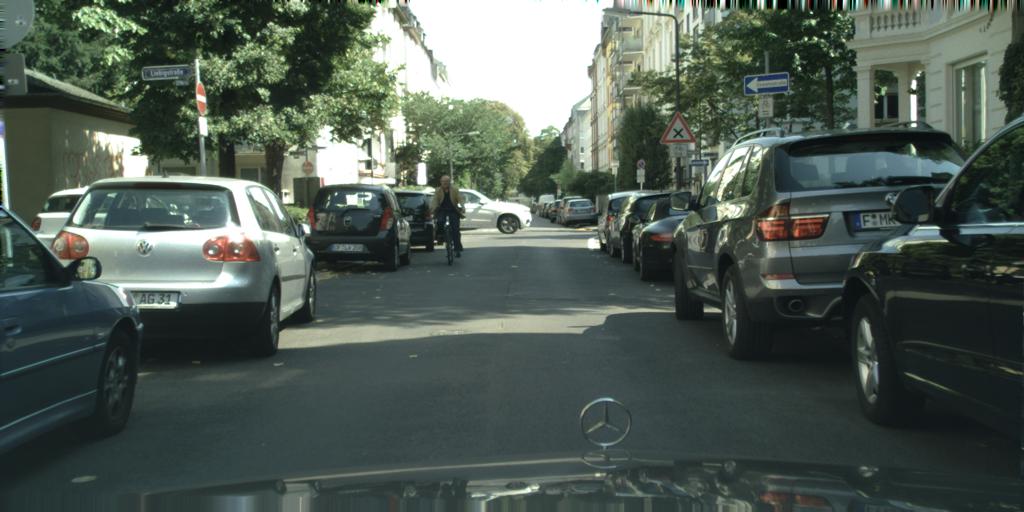} &
\begin{overpic}[width=0.32\linewidth]{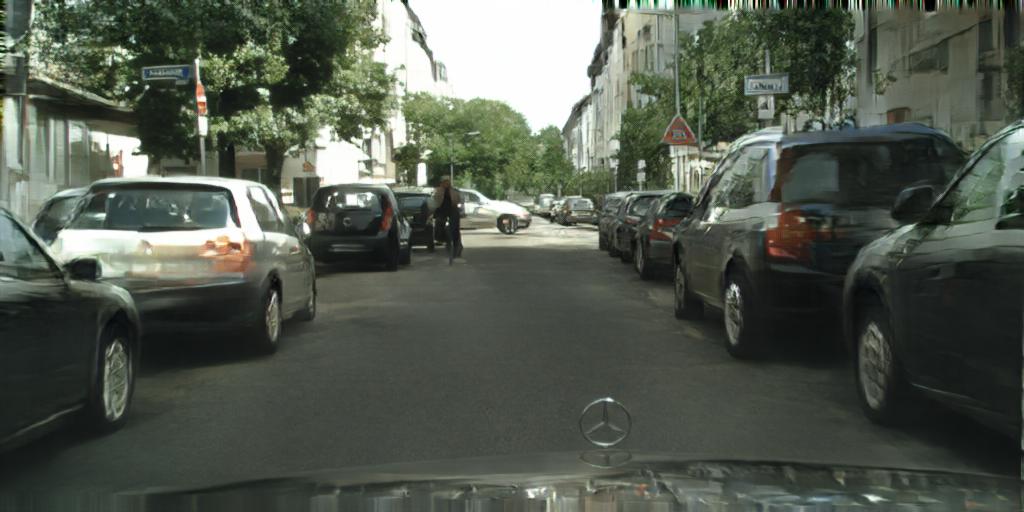}
\put(1,1){\setlength{\fboxsep}{0pt}\setlength{\fboxrule}{0.2pt}\fbox{\includegraphics[width=0.064000\linewidth]{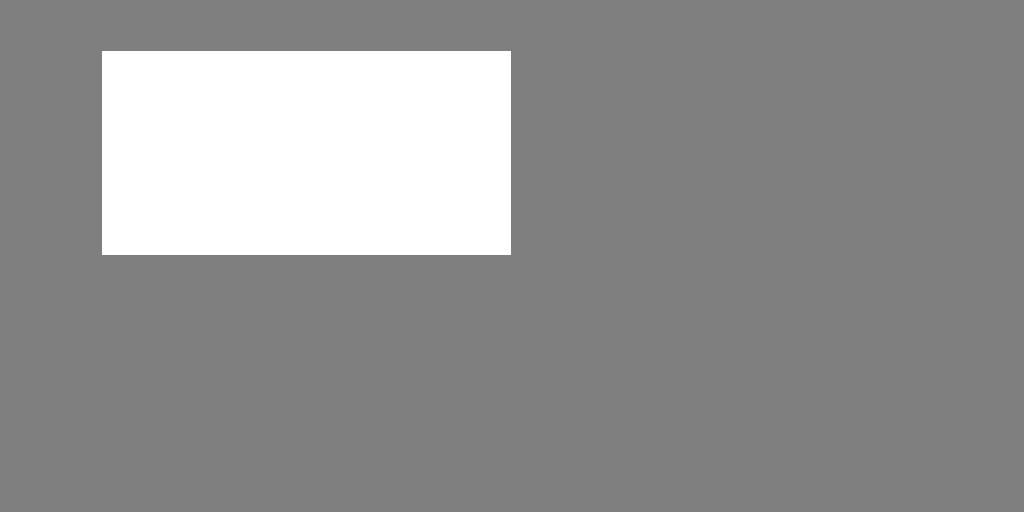}}}
\end{overpic} &
\begin{overpic}[width=0.32\linewidth]{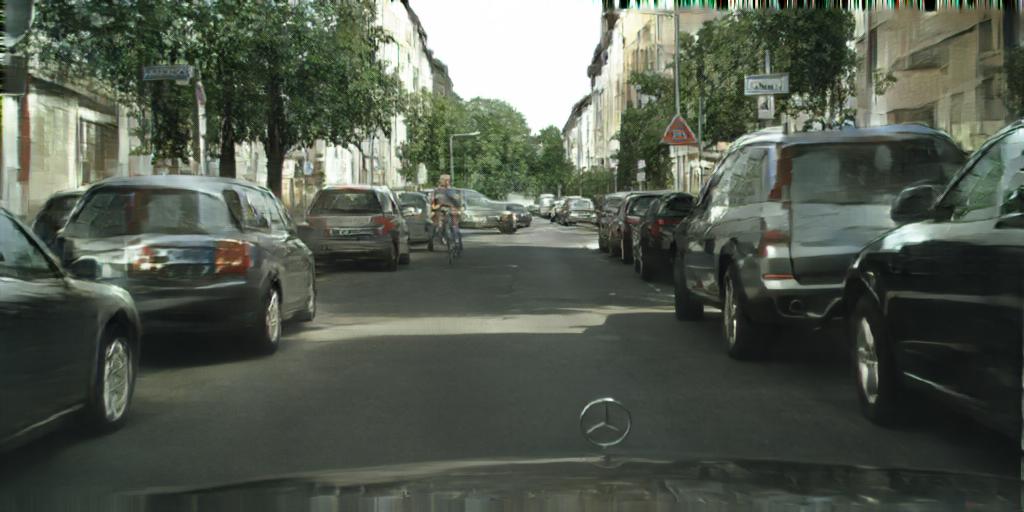}
\put(1,1){\setlength{\fboxsep}{0pt}\setlength{\fboxrule}{0.2pt}\fbox{\includegraphics[width=0.064000\linewidth]{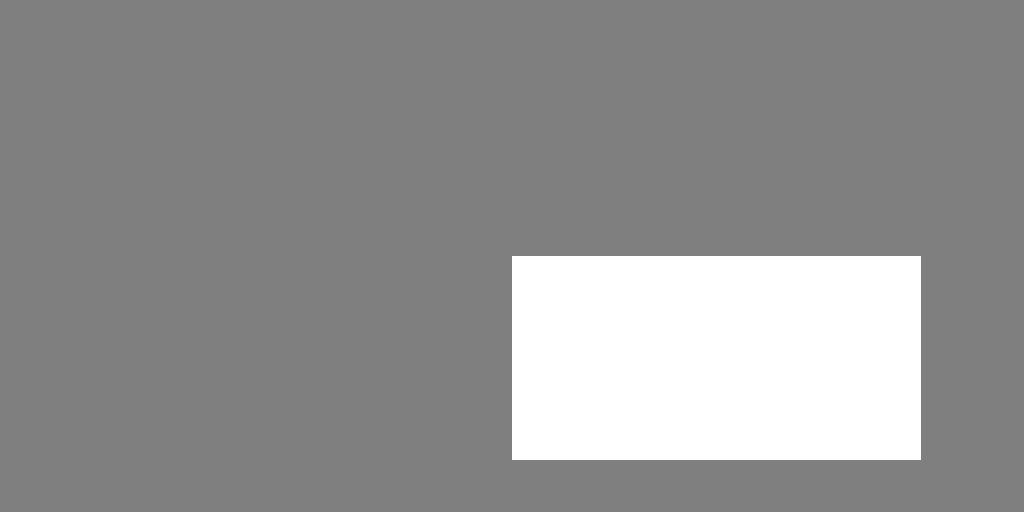}}}
\end{overpic} \\
%
%
\includegraphics[width=0.32\linewidth]{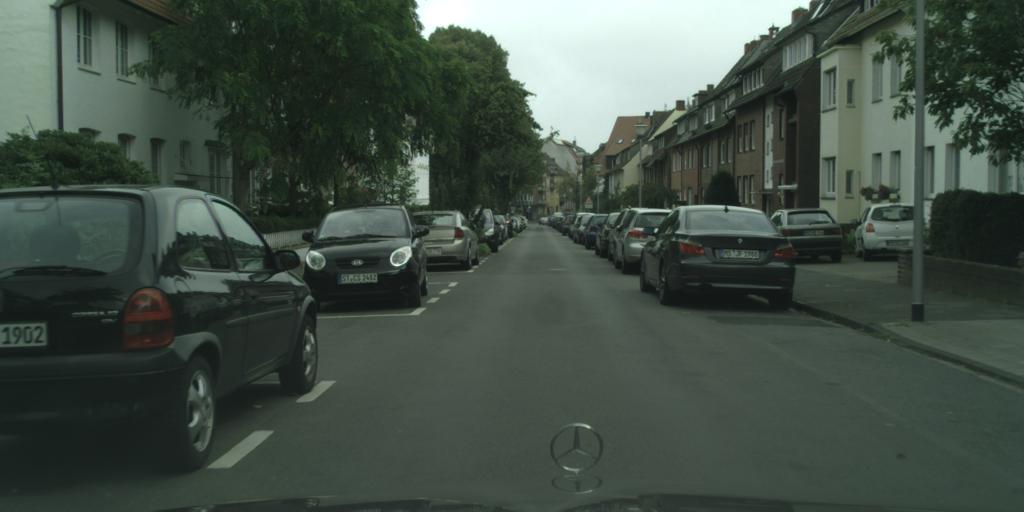} &
\begin{overpic}[width=0.32\linewidth]{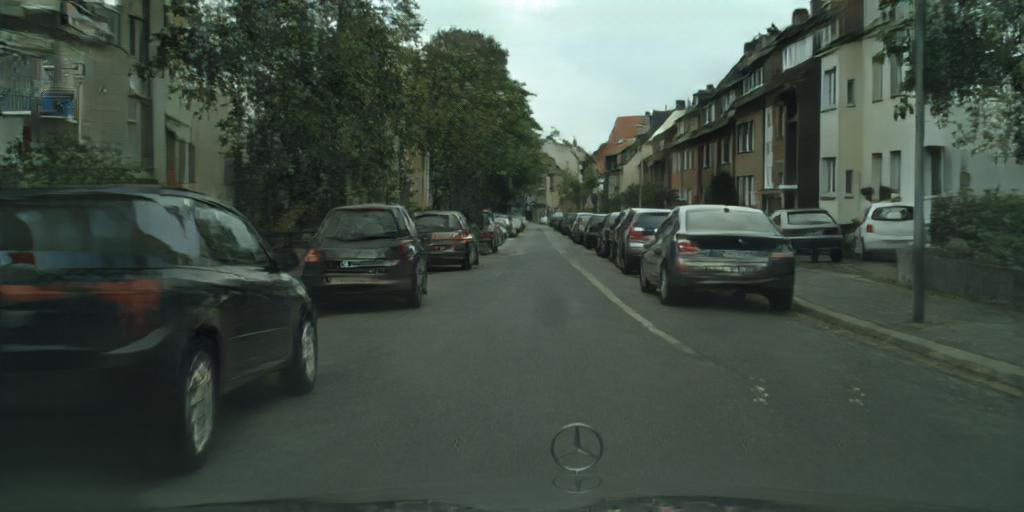}
\put(1,1){\setlength{\fboxsep}{0pt}\setlength{\fboxrule}{0.2pt}\fbox{\includegraphics[width=0.064000\linewidth]{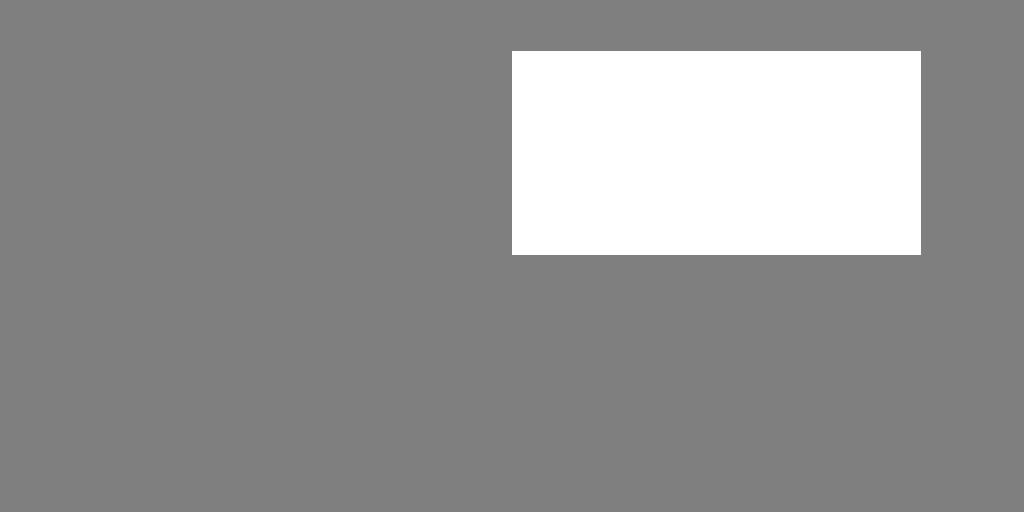}}}
\end{overpic} &
\begin{overpic}[width=0.32\linewidth]{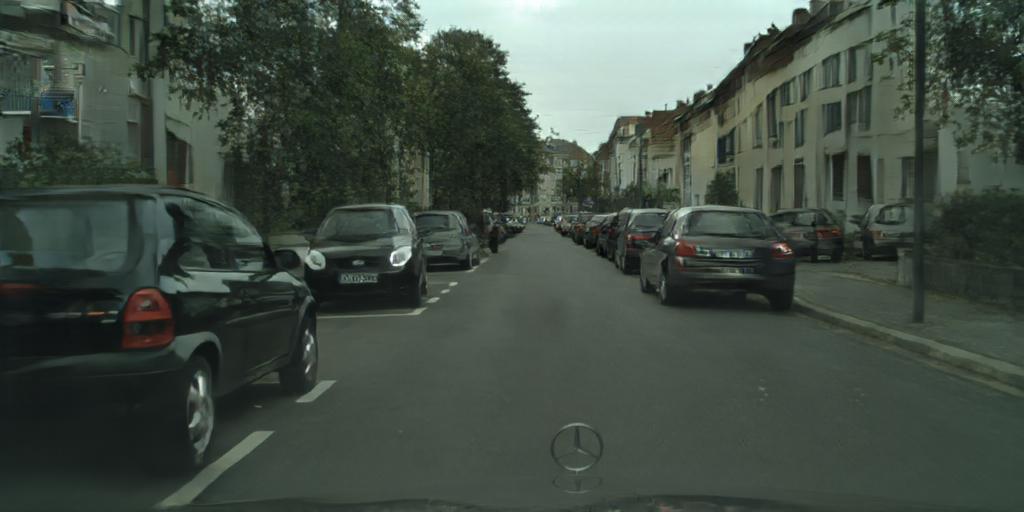}
\put(1,1){\setlength{\fboxsep}{0pt}\setlength{\fboxrule}{0.2pt}\fbox{\includegraphics[width=0.064000\linewidth]{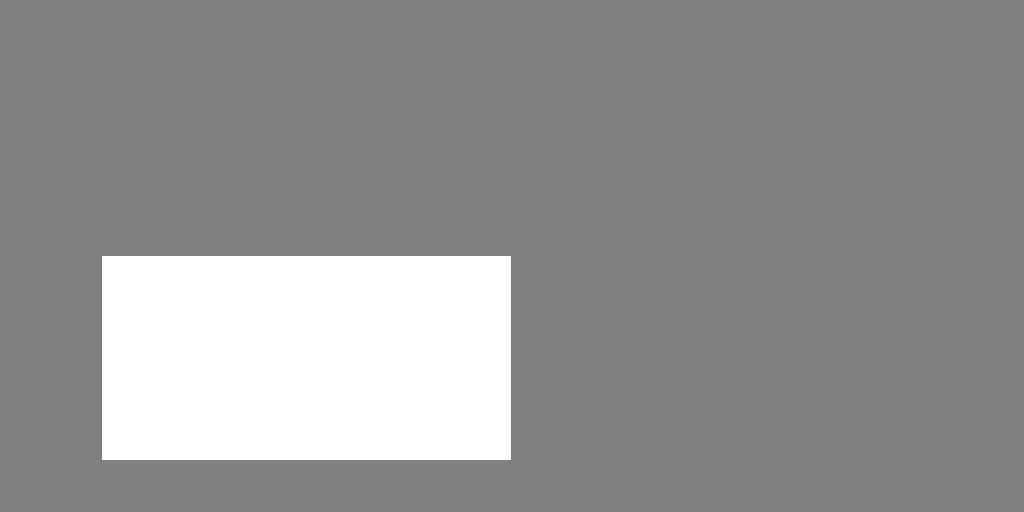}}}
\end{overpic} \\
%
%
\end{tabular}

%% file: fig_predicted.tex
\begin{tabular}{ccc}
\includegraphics[width=0.33\linewidth]{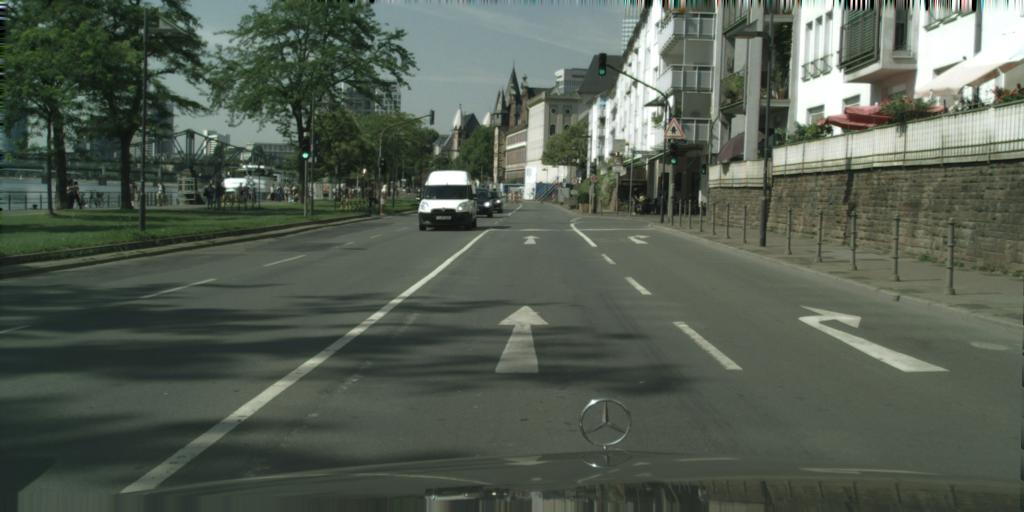} &
\includegraphics[width=0.33\linewidth]{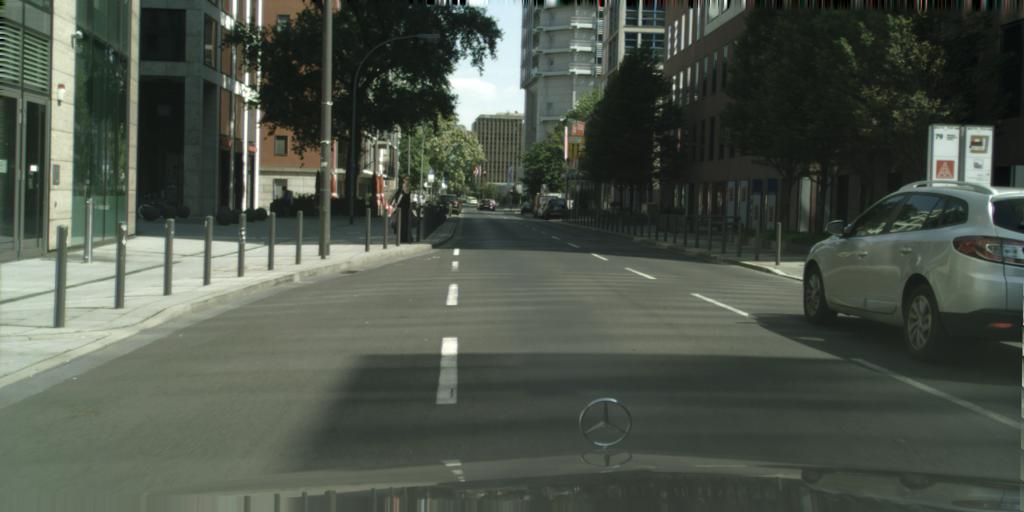} &
    \includegraphics[width=0.33\linewidth]{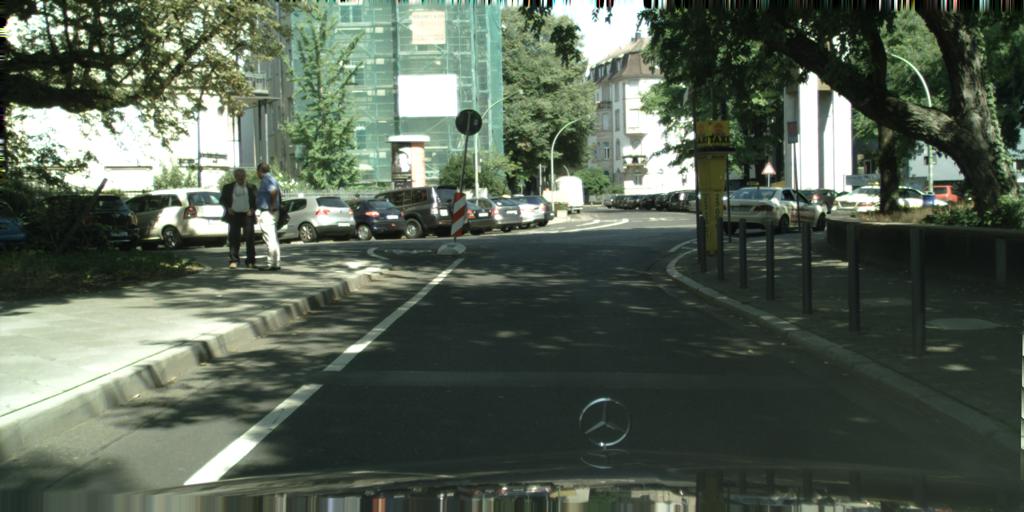} \\
\begin{overpic}[width=0.33\linewidth]{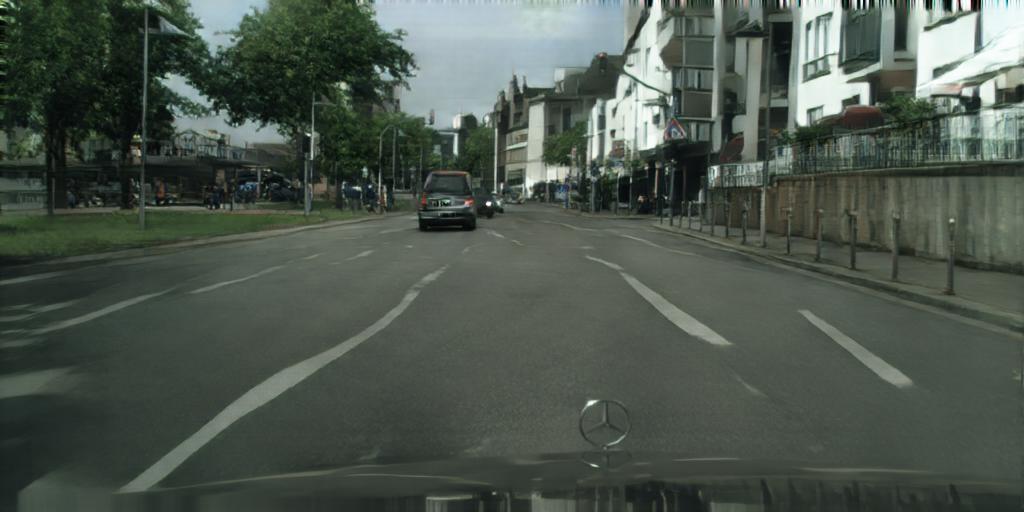}
\put(1,1){\setlength{\fboxsep}{0pt}\setlength{\fboxrule}{0.2pt}\fbox{\includegraphics[width=0.082500\linewidth]{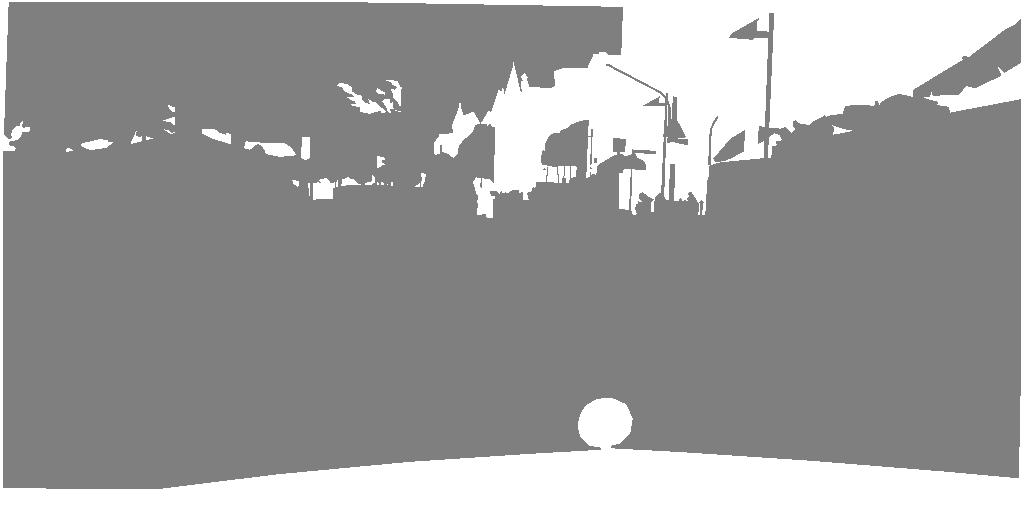}}}
\end{overpic} &
\begin{overpic}[width=0.33\linewidth]{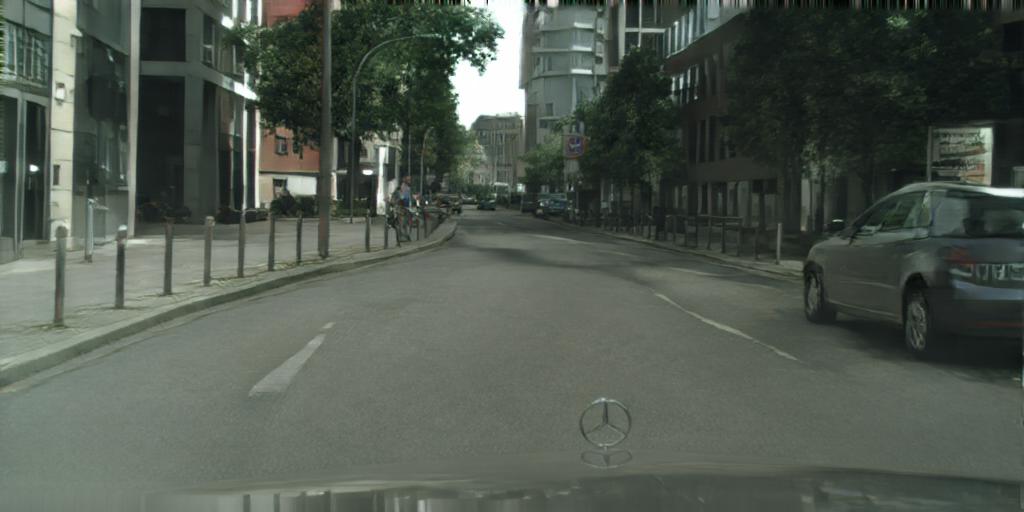}
\put(1,1){\setlength{\fboxsep}{0pt}\setlength{\fboxrule}{0.2pt}\fbox{\includegraphics[width=0.082500\linewidth]{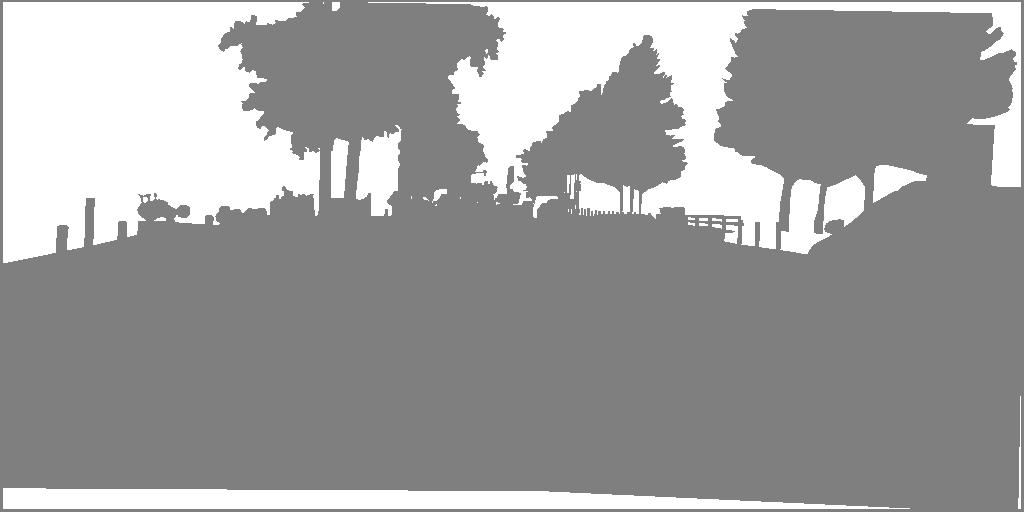}}}
\end{overpic} &
\begin{overpic}[width=0.33\linewidth]{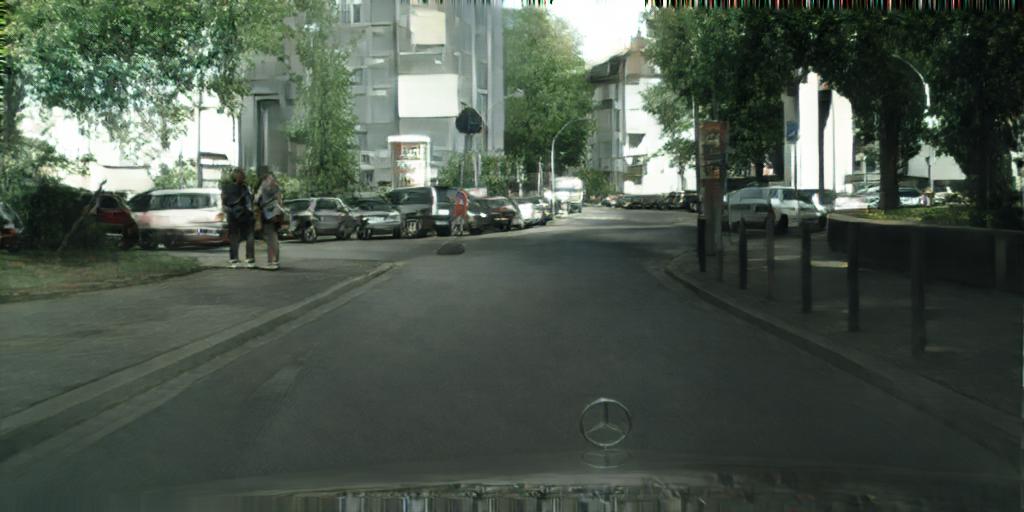}
\put(1,1){\setlength{\fboxsep}{0pt}\setlength{\fboxrule}{0.2pt}\fbox{\includegraphics[width=0.082500\linewidth]{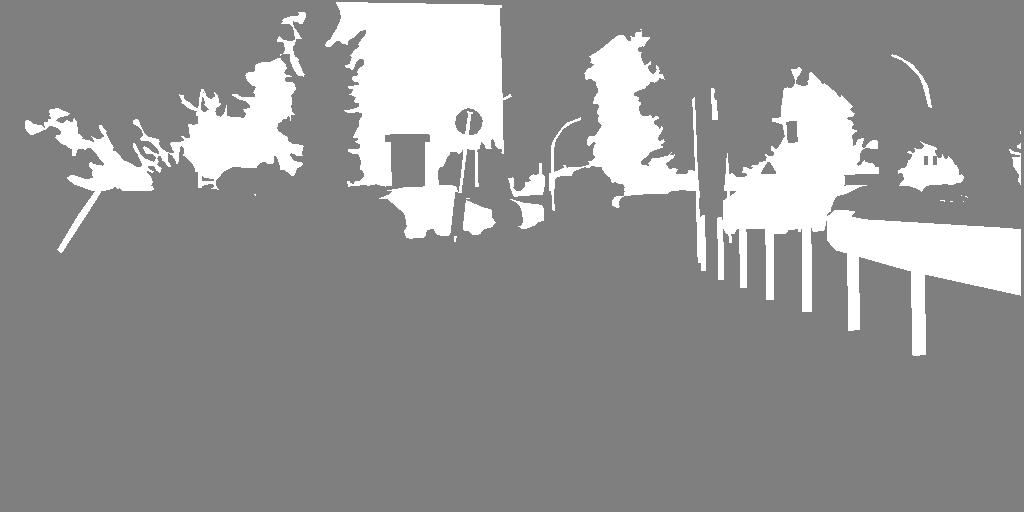}}}
\end{overpic} \vspace{-2.6ex} \\
$0.019$ bpp &  $0.021$ bpp & $0.013$ bpp    \\
%
\end{tabular}